\documentclass[11pt]{article}

\usepackage{acl}

\usepackage{times}
\usepackage{latexsym}

\usepackage[T1]{fontenc}

\usepackage[utf8]{inputenc}

\usepackage{microtype}

\usepackage{inconsolata}

\usepackage[utf8]{inputenc}
\usepackage[T1]{fontenc}
\usepackage{hyperref}
\usepackage{url}
\usepackage{amsfonts}
\usepackage{amsmath, amssymb}
\usepackage{nicefrac}
\usepackage{microtype}
\usepackage[table]{xcolor}
\usepackage{graphicx}
\usepackage{multirow}
\usepackage{listings}
\usepackage{courier}
\usepackage{longtable}
\usepackage{makecell}
\usepackage{array}
\usepackage{booktabs}
\usepackage{arydshln}      
\usepackage[ruled,vlined,linesnumbered]{algorithm2e}
\usepackage[most]{tcolorbox}
\usepackage{pifont}
\usepackage{caption}
\usepackage{tabularx}
\usepackage{fontawesome5}
\usepackage{gradient-text}

\definecolor{cNeg1}{HTML}{FAECE7}
\definecolor{cNeg2}{HTML}{F6C7BA}
\definecolor{cNeg3}{HTML}{EE9B8C}

\definecolor{cZero}{HTML}{F1EFE8}
\definecolor{cPos1}{HTML}{E1F5EE}
\definecolor{cPos2}{HTML}{9FE1CB}
\definecolor{cPos3}{HTML}{5DCAA5}

\definecolor{myblue}{HTML}{8BC0E7}

\definecolor{lightbluebg}{HTML}{E8F1FB}
\definecolor{blueframe}{HTML}{2C5AA0}
\definecolor{lightgreenbg}{HTML}{E8F5EC}
\definecolor{greenframe}{HTML}{2E7D32}
\definecolor{lightorangebg}{HTML}{FBEFE3}
\definecolor{orangeframe}{HTML}{B85C00}
\definecolor{lightyellowbg}{HTML}{FFF8DC}
\definecolor{yellowframe}{HTML}{B8860B}

\newcommand{\cmark}{\textcolor{green!50!black}{\checkmark}}   
\newcommand{\xmark}{\textcolor{red}{\ding{55}}}

\definecolor{BlueBG}{rgb}{0,0.46,0.71}

\newtcolorbox[auto counter, number within=section]{promptbox}[2][]{%
  enhanced, breakable,
  colback=lightbluebg!40!white, colframe=blueframe,
  boxrule=0.6pt, arc=2pt,
  title={\bfseries Prompt~\thetcbcounter: #2}, fonttitle=\bfseries,
  left=4pt, right=4pt, top=4pt, bottom=4pt,
  #1,
}
\newtcolorbox[auto counter, number within=section]{pfbox}[2][]{%
  enhanced, breakable,
  colback=lightorangebg!40!white, colframe=orangeframe,
  boxrule=0.6pt, arc=2pt,
  title={\bfseries PF~\thetcbcounter: #2}, fonttitle=\bfseries,
  left=4pt, right=4pt, top=4pt, bottom=4pt,
  #1,
}
\newtcolorbox[auto counter, number within=section]{skillbox}[2][]{%
  enhanced, breakable,
  colback=lightgreenbg!40!white, colframe=greenframe,
  boxrule=0.6pt, arc=2pt,
  title={\bfseries Skill~\thetcbcounter: #2}, fonttitle=\bfseries,
  left=4pt, right=4pt, top=4pt, bottom=4pt,
  #1,
}
\newtcolorbox{failbox}[1][]{%
  enhanced, breakable,
  colback=cNeg1!50!white, colframe=cNeg3,
  boxrule=0.6pt, arc=2pt,
  title={\bfseries Baseline trajectory (incorrect)}, fonttitle=\bfseries,
  left=4pt, right=4pt, top=4pt, bottom=4pt,
  #1,
}
\newtcolorbox{winbox}[1][]{%
  enhanced, breakable,
  colback=cPos1!60!white, colframe=cPos3,
  boxrule=0.6pt, arc=2pt,
  title={\bfseries Our trajectory (correct)}, fonttitle=\bfseries,
  left=4pt, right=4pt, top=4pt, bottom=4pt,
  #1,
}
\newtcolorbox{pfeventbox}[1][]{%
  enhanced, breakable,
  colback=lightyellowbg!75!white, colframe=yellowframe,
  boxrule=0.5pt, arc=1.5pt,
  left=3pt, right=3pt, top=2pt, bottom=2pt,
  fontupper=\scriptsize, sharp corners=northwest,
  #1,
}

\newcommand{\ours}{\textsc{HASP}\xspace}    

\title{Harnessing LLM Agents with Skill Programs}

\author{%
 \textbf{Hongjun Liu}$^{1}$,
 \textbf{Yifei Ming}$^{2}$, 
 \textbf{Shafiq Joty}$^{2}$,
 \textbf{Chen Zhao}$^{1}$
   \vspace{.5em} 
  \\
  $^1$New York University, 
  $^2$Salesforce AI Research, \\[4pt]
\href{https://holayan.github.io/HASP_Blog/} 
{\faGlobe\xspace\nolinkurl{ProjectBlog}}
\quad
\href{https://holayan.github.io/HASP_Blog/clips/hasp_promo.mp4}
{\faPlayCircle\xspace\nolinkurl{DemoVideo}}
\quad
\href{https://github.com/HolaYan/ExecutableSkills-HASP}
{\faPlayCircle\xspace\nolinkurl{CodeBase}}
}

\begin{document}
\maketitle

\begin{abstract}
Equipping LLM agents with reusable skills derived from past experience has become a popular and successful approach for tackling complex and long-horizon tasks. However, such lessons are often encoded as textual guidance that remains largely advisory, lacking explicit mechanisms for when and how to intervene in the agent loop. To bridge the gap, we introduce \ours (\textbf{H}arnessing LLM \textbf{A}gents with \textbf{S}kill \textbf{P}rograms), a framework that upgrades skills into executable Program Functions (PFs). Rather than offering passive advice, PFs act as executable guardrails that activate on failure-prone states and modify the next action or inject corrective context. \ours is highly modular: it can be applied at inference time for direct agent-loop intervention, during post-training to provide structured supervision, or for self-improvement by evolving validated, teacher-reviewed PFs. Empirically, \ours drives substantial gains compared to both training-free and training-based methods on web-search, math reasoning, and coding tasks. For example, on web-search reasoning, inference-time PFs alone improve the average performance by 25\% compared to (multi-loop) ReAct Agent, while post-training and controlled evolution achieve a 30.4\% gain over Search-R1. To provide deeper insights into \ours, our mechanism analysis reveals how PFs trigger and intervene, how skills are internalized, and the requirement for stable skill library evolution. Our code is released on \href{https://github.com/HolaYan/ExecutableSkills-HASP}{\gradientRGB{HASP}{214,93,72}{86,81,155}}.
\begingroup
\renewcommand{\thefootnote}{}
\makeatletter
\renewcommand{\@makefntext}[1]{\noindent #1}
\makeatother
\footnotetext{Correspondence: 
\href{mailto:lh3862@nyu.edu}{lh3862@nyu.edu}}
\endgroup

\end{abstract}

\section{Introduction}

Recent advances in large language models (LLMs) enable increasingly capable agents~\citep{ schick2023toolformerlanguagemodelsteach, wu2024autogen,li2025intheflowagenticoptimizationeffective, li2025swarmsysdecentralizedswarminspiredagents} that plan, interact with environments, and solve complex tasks effectively~\citep{jin2025search,jiang2025verltool, liu2025suceareasoningintensiveretrievaladversarial}. However, as task distributions shift and feedback accumulates across episodes, many agent failures recur in recognizable forms. Multi-step agents may terminate before verification, commit to brittle intermediate conclusions, or repeat unproductive actions~\citep{cemri2026why}. A central challenge, therefore, is to enable agents to recognize recurring failure patterns, abstract them into reusable knowledge or skills, and adapt future behavior accordingly~\citep{shinn2023reflexionlanguageagentsverbal,lu2026skill0incontextagenticreinforcement}.

\begin{figure}[!t]
    \centering
    \includegraphics[width=\linewidth]{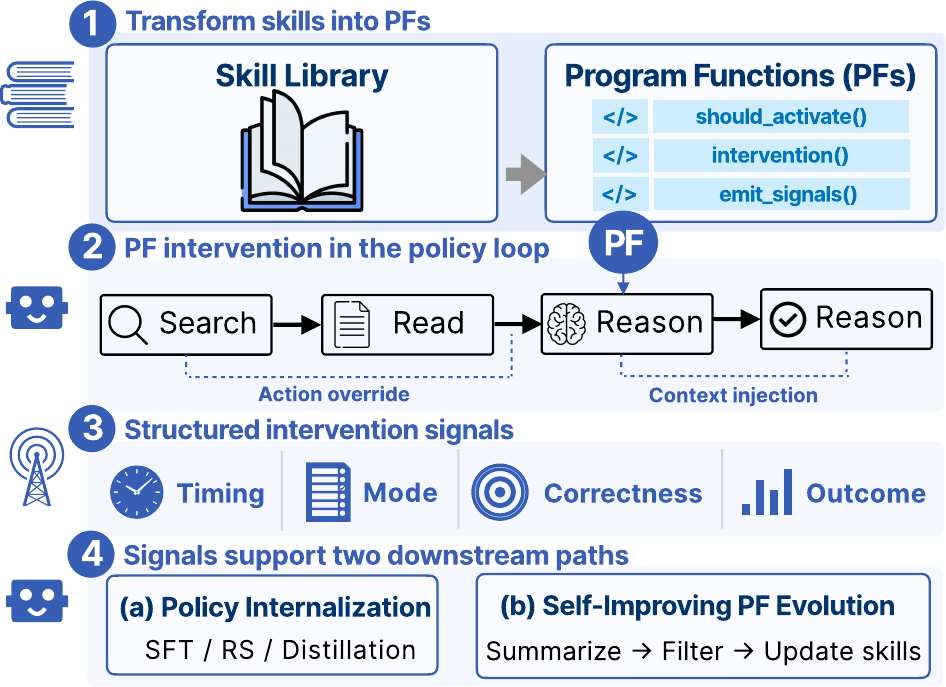}
    \vspace{-7mm}
    \caption{
    Overview of \ours. Unlike prompted skills, \ours represents skills as Program Functions (PFs) that activate on runtime states, intervene through replacing the proposed action or injecting corrective context, and support training and validated skill evolution.  
    }
    \vspace{-5mm}
\label{fig:figure1}
\end{figure}
A natural response is to reuse past experience as \emph{skills}, as behavioral knowledge abstracted from prior agent interactions. Existing agent systems already do so, but mostly in \emph{textual} form~\citep{shinn2023reflexionlanguageagentsverbal,lu2026skill0incontextagenticreinforcement,zhao2024expelllmagentsexperiential}: they are injected into prompts, retrieved as advice, or used indirectly to shape rewards during post-training. This makes them flexible, but also largely \emph{advisory}. A textual skill can express what the agent should do in principle, but not precisely \emph{when} it should activate inside the policy loop, \emph{how} it should alter the next decision, and it is often ignored by the model in practice. As a result, there remains a gap between reusable experience expressed in language and reusable experience that can reliably and explicitly control agent behavior.

To bridge this gap, we present \textbf{\ours}~(\textbf{H}arnessing LLM \textbf{A}gents with \textbf{S}kill \textbf{P}rograms), a framework that reframes skills into executable \emph{Program Functions} (PFs). As shown in Figure~\ref{fig:figure1}, PF is a reusable state--action intervention function: given the current agent state and a candidate next action, it decides whether intervention is needed and, if so, explicitly revises the proposed action or injects corrective context. For example, when the agent prematurely proposes \texttt{FINAL}, a PF can replace it with \texttt{READ}; when the error lies in the reasoning itself, it can instead inject corrective context that redirects the next step. In this way, a skill is no longer a passive guideline the model may choose to follow; it becomes an executable object that can be triggered on demand and can intervene directly in the agent loop. \ours operates as an external \emph{agent harness}, a control layer around the base agent: at each step, the base policy proposes an action, the harness retrieves relevant PFs, evaluates their activation predicates, executes valid interventions, and feeds the revised action or injected context back into the loop. 
A concrete PF follows the two-module template: \texttt{should\_activate()} specifies when the skill fires, while \texttt{intervene()} returns an action modification, context injection, or abstention. 

We initialize PFs from failure cases in the training pool, instantiate them with explicit activation and intervention interfaces, and admit them into the active library after syntax and mock-execution validation. 
The same state-action intervention interface makes \ours modular. At inference time, \ours can be plugged into existing agent loop to revise actions or inject corrective context without model updates. When post-training is available, each PF execution provides a structured record containing the original action, the PF repair, the activated skill, and the observed effect. 
\ours scores these events with structured PF-derived criteria, and uses the resulting PF-corrected traces to train the student base policy via SFT, rejection sampling, or on-policy distillation.
Finally, when in-training skill evolution is enabled, \ours revisits failures under the current checkpoint, summarizes recurring failure-repair patterns into candidate PFs, filters them through executable validation and teacher review, and updates the external skill library, thereby closing the loop between execution, learning, and library growth.

We evaluate \ours on web-search reasoning, mathematical reasoning, and coding. 
Even in the inference-time PF intervention, where PFs trigger based on agent's state, \ours achieves large gains over competitive baselines, underscoring the inherent value of PF design: on web-search reasoning, average accuracy rises to 51.0\%. Adding auxiliary teacher for PF selection further improves inference-time performance to 56.2\%. Beyond inference-time intervention, PF-derived supervision can be internalized without full reinforcement learning: under fixed skill library, rejection sampling reaches 59.3\%, and on-policy distillation reaches 62.5\% on web-search reasoning. Controlled evolution further improves the external skill library when paired with stable selection, with closed-loop rejection sampling reaching 60.3\% on web-search reasoning and improving mathematical reasoning to 45.4\%. On coding, teacher-augmented PF intervention reaches 68.7\% average pass@1, while PF-based training improves it to 69.9\%. 
Mechanism analysis further studies how PFs trigger and intervene, how skills are internalized, and the requirement for stable library evolution.


Our contributions are threefold:
(1) \textbf{Skills as executable state-action intervention functions.}
We transform reusable agent experience into executable Program Functions that can be triggered on demand and intervene explicitly in the agent loop, moving beyond passive textual skills.
(2) \textbf{\ours: a highly-modular agent harness framework.}
We propose \ours, an agent harness that supports controllable PF triggering and intervention, effective across inference-only, post-training, and self-improving paradigms.
(3) \textbf{Strong empirical performance}. \ours achieves substantial gains over competitive baselines across a wide range of tasks.

\section{Harnessing LLM Agents with Skill Programs}
\label{sec:method}

\begin{figure*}[!t]
    \centering
    \includegraphics[width=\linewidth]{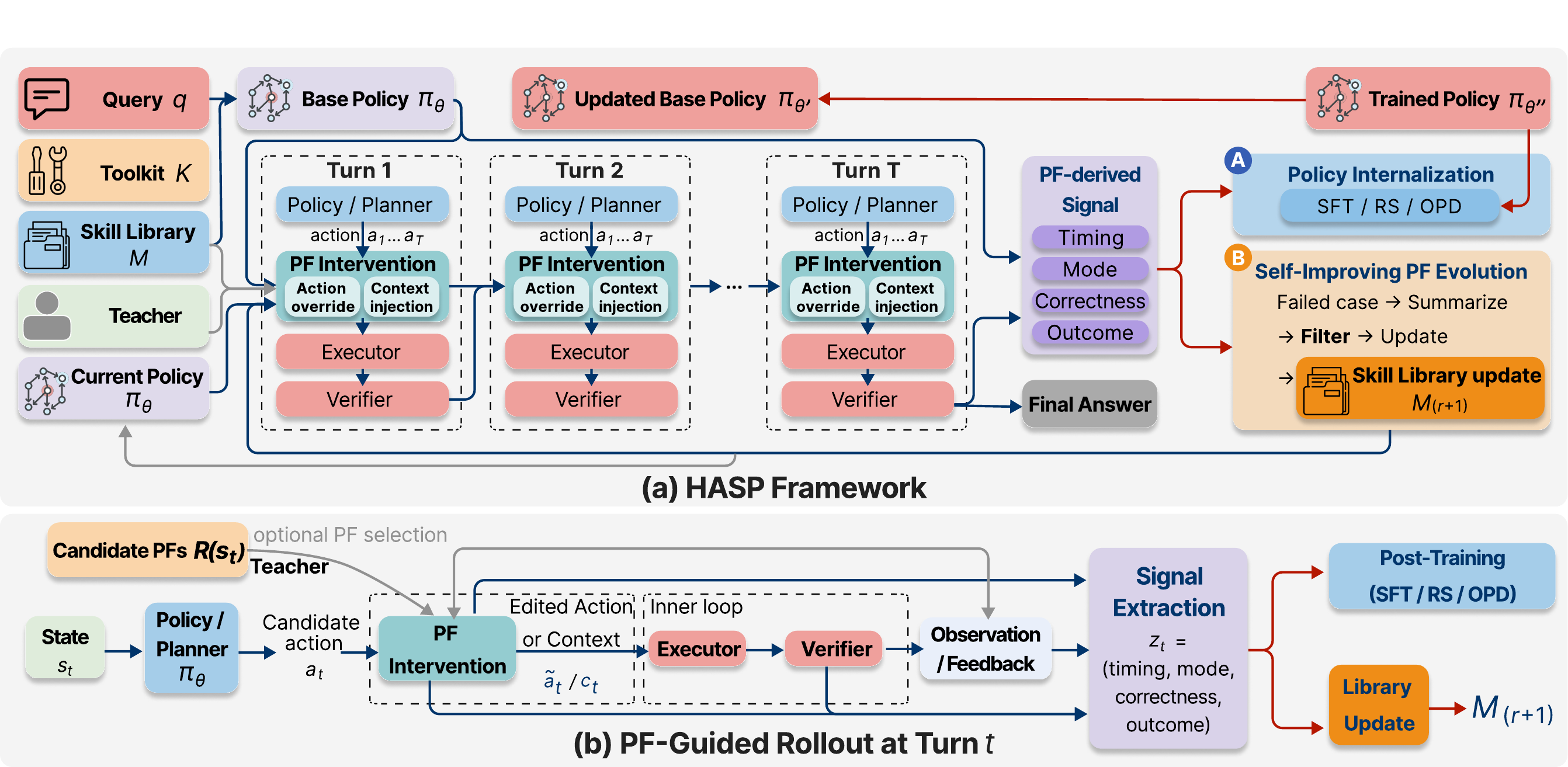}
    \vspace{-7mm}
    \caption{
    Overview of \ours.
    (a) At inference time, retrieved Program Functions (PFs) guide multi-turn agent rollouts by modifying actions or injecting context, while emitted signals support policy internalization and PF evolution.
    (b) A PF-guided turn converts the policy proposal, intervention, execution result, and feedback into signals (timing, mode, correctness, and outcome) for post-training and skill library update. }
    \vspace{-4mm}
\label{fig:pipeline}
\end{figure*}

We present \textbf{\ours}, a framework that turns agent experience into executable state--action intervention functions through \emph{Program Functions} (PFs). PFs can be inserted into an agent's reasoning process and triggered while the agent is solving a task, allowing them to correct intermediate decisions.
Each PF execution leaves an intervention record of what the agent planned to do, how the PF changed it, and what happened afterward. These records can be used for post-training, while repeated failure patterns are turned into new candidate PFs and added to the skill library after validation. Figure~\ref{fig:pipeline} provides an overview of the framework and how PFs intervene in the reasoning process.

\subsection{Inference-Time Agent Harness with Program Functions}

We consider an agent that reasons step by step and uses external tools while solving a task. Its policy is denoted by $\pi_\theta$.
Given input $x$, the agent produces a sequence of steps $\tau=(s_0,a_0,o_1,\ldots,s_T,a_T)$, where $s_t$ is agent's current state, $a_t$ is the next action it proposes, and $o_{t+1}$ is the result returned by the environment or external tools. At inference time, \ours wraps this base policy with an external \emph{harness}, a control layer that retrieves relevant PFs from the skill library and lets them intervene before the next action is executed, enabling direct correction and improvement of intermediate decisions. The agent has access to an external toolkit $K$ for tool execution and environment interaction.

\noindent\textbf{Program Functions and Skill Library.}
Each skill in \ours is represented as a \emph{Program Function} (PF), an executable module that decides whether to intervene under the current state and proposed action, and if so, how to repair the next decision. As shown in Figure~\ref{fig:figure1}, each executable skill has two parts: \texttt{should\_activate}, which decides whether the skill should fire, and \texttt{intervene}, which returns the repair. Compared with natural-language reminders, PFs make skills explicit and executable: instead of merely stating a principle such as ``avoid repeated searches,'' a PF specifies both when that principle applies and how the next decision should change.
We maintain an external skill library $\mathcal{M}$. To initialize this library, we collect recovered failure cases from the training pool and summarize recurring failure--repair patterns into reusable candidate PFs (\emph{e.g.,} entity confusion). Each candidate PF must specify both an activation condition and an intervention behavior, and is admitted into the library only after syntax, interface, and mock-execution validation. This ensures that the skill library contains executable and reusable interventions.

\noindent\textbf{PF-guided intervention in the agent loop.}
At step $t$, the base policy first proposes an action $a_t^{\mathrm{orig}} \sim \pi_\theta(\cdot \mid s_t)$. The harness then retrieves candidate PFs $\mathcal{R}(s_t)\subseteq\mathcal{M}$ and evaluates their activation functions on the current state and proposed action. In the PF-only setting, PFs are triggered solely by these activation functions. The harness then applies the activated PFs through an intervention operator \texttt{Intervene}, producing $(\tilde a_t, c_t, \kappa_t)=\texttt{Intervene}(s_t,a_t^{\mathrm{orig}},\mathcal{R}(s_t))$, where $\tilde a_t$ denotes the revised action after PF intervention, $c_t$ is optional corrective context injected into subsequent reasoning, and $\kappa_t$ records the fired PFs and intervention mode. If no PF activates, then $\tilde a_t=a_t^{\mathrm{orig}}$; otherwise, $\tilde a_t$ may be a modified or redirected action returned by the activated PFs.

PF intervention can operate in two main modes. First, a PF may directly modify the next action itself, yielding a repaired executable action $\tilde a_t$; for example, it may rewrite an over-constrained search query, redirect retrieval toward a more informative intermediate entity. Second, a PF may inject corrective context $c_t$ back into the reasoning process, such as a warning about similar entities. This design separates choosing the next action from correcting it: the policy proposes what to do next, while PFs determine whether the proposed action should be executed as is, revised into a better action, or augmented with additional context. The pair $(a_t^{\mathrm{orig}},\tilde a_t)$ therefore records both what the policy would have done and how the harness repaired or redirected it, turning skill use into supervision over intermediate decisions rather than only final-answer feedback.

\noindent\textbf{Auxiliary teacher for PF selection.}
\ours supports a minimal PF-only intervention setting, where candidate PFs are triggered solely by their activation functions. When available, an auxiliary teacher can further help select which PF to apply when more than one PF could reasonably be applied, improving intervention precision in ambiguous cases.

\subsection{HASP for Post-Training: Internalizing PF-Guided Interventions}

Beyond improving inference-time behavior, PF-guided intervention also provides supervision during post-training. Each PF activation produces record $e_t=(s_t,a_t^{\mathrm{orig}},\tilde a_t,c_t,\kappa_t,\Delta_t)$, which contains the triggering state, the action proposed by original policy, the corrected action, any injected context, metadata, and the resulting feedback. These events connect inference-time intervention to post-training: they capture when the intervention occurred, how it was applied, and whether it helped. \ours scores each intervention record using four signals
$\mathbf{z}_t=(t_t,m_t,q_t,o_t)$, corresponding to intervention timing, mode, correctness, and outcome, and aggregates them into a single per-step score $A_t$.
Each signal is a normalized weighted sum of sub-signals that evaluates a distinct aspect of the intervention: whether a PF should have fired at this step, whether it intervened at the appropriate point in the ReAct cycle, whether the resulting repair was valid and well-formed, and whether the intervention improved the final episode outcome after accounting for its cost.
Importantly, undesirable behaviors, such as firing on a safe step, failing to intervene on a risky one, or modifying an action that subsequently fails, receive negative weights. Thus, $A_t$ is a signed measure of intervention quality rather than a reward for intervention itself (Appendix~\ref{app:signals}).
We further compute trajectory-level PF score as $A(\tau)=T^{-1}\sum_{t=1}^{T}A_t$. Rather than training only on final answers, \ours uses PF-corrected actions and trajectories, scored by these PF-derived signals, to update the student base policy. 
The goal is to train the base policy toward PF-corrected actions and trajectories.
Our main trained variant, \textbf{\ours-Evolve + RS}, combines evolving skill library with PF-guided rejection sampling; SFT and OPD are evaluated as alternative training paths.

\noindent\textbf{Main recipe: PF-guided rejection sampling.}
Given sampled trajectories $\{\tau_i\}_{i=1}^{N}$, \ours scores each trajectory using both final task success and PF-guided intervention quality, with $\mathrm{Score}(\tau_i)=\beta_1\mathrm{TaskSuccess}(\tau_i)+\beta_2A(\tau_i)$. Only top-scoring trajectories are retained for training. Unlike standard rejection sampling, which only filters by final correctness, \ours also prefers trajectories whose intermediate decisions better match PF-guided correction: interventions should occur at appropriate states, use suitable modes, and lead to valid repairs with positive downstream effects. This recipe is well suited to the evolving skill library because it filters out noisy or harmful trajectories.

\noindent\textbf{Variant 1: supervised fine-tuning on corrected actions.}
As a simpler baseline, each intervention record provides a corrected target
$s_t\mapsto\tilde a_t$, weighted by intervention quality.
We retain records with $A_t\ge\tau$ (we use $\tau=0.3$) and optimize
$
\mathcal{L}_{\mathrm{SFT}}
=
-\sum_t w_t\log\pi_\theta(\tilde a_t\mid s_t),
\qquad
w_t=A_t,$
optionally keeping only the top-$k$ scoring intervention steps within each episode.
Because $A_t$ is signed, thresholding removes interventions that fire on safe states or produce unsuccessful repairs rather than training on them with a small positive weight.
This directly trains the student on PF-corrected local actions, such as rewriting over-constrained search queries, enforcing evidence verification before finalization, or using injected context to improve the next decision.

\noindent\textbf{Variant 2: on-policy distillation (OPD).}
To test whether the scoring scheme also helps on the student's own states, OPD rolls out current policy, keeps PFs active on failure-prone steps, and trains the student on corrected behavior with $\mathcal{L}_{\mathrm{OPD}}=-\sum_t \hat w_t\log\pi_\theta(\tilde a_t\mid s_t)$, where $\hat w_t$ reflects intervention quality under current policy. Unlike the SFT weight $w_t=A_t$, $\hat w_t$ does not depend on the four intervention-quality signals; instead, it emphasizes states where the student's current behavior diverges from the reference trajectory. OPD aligns training with student's own states, but can be less stable when the skill library also changes.

\subsection{HASP for Self-Improving Skill Library}

The same PF interface also supports controlled growth of the skill library. After fixed training intervals, \ours revisits remaining failures under the current checkpoint and proposes candidate PFs $\mathcal{C}_r$ from recurring failure--repair patterns. Each candidate must specify both an activation condition and an intervention behavior so that it can be inserted into the same rollout harness. To prevent library pollution, candidates pass two gates. The first is executable and binary: a candidate is discarded unless its code parses, exposes \texttt{should\_activate} and \texttt{intervene} with the right signatures, runs without raising on sampled step contexts, and returns a well-formed \texttt{Intervention}. Survivors are then scored on five dimensions, concept, trigger, intervention, executability, and validation utility, weighted $(0.25,0.20,0.20,0.20,0.15)$, into $Q_{\mathrm{rev}}(c)\in[0,1]$, and admitted when $Q_{\mathrm{rev}}(c)\ge\eta$. We use $\eta=0.60$ for a candidate that refines an existing failure category and $\eta=0.75$ for one that opens a new category, since a new category enlarges the space the selector must rank over and a weak one costs every later question; candidates in $[0.42,0.60)$ are returned for revision rather than dropped. A candidate whose executability dimension falls below $0.30$ is rejected outright regardless of its total, on the grounds that a skill that cannot run cannot be repaired by a better trigger.
This filtering step removes overly specific, redundant, or noisy PFs that would make retrieval less accurate and weaken future interventions. Accepted PFs update the external skill library $\mathcal{M}_r\rightarrow\mathcal{M}_{r+1}$, while post-training updates the base policy $\pi_\theta\rightarrow\pi_{\theta'}$.

\section{Experiment}
\label{sec:exp}

\begin{table*}[!ht]
\centering
\tiny
\renewcommand{\arraystretch}{1.08}
\resizebox{\textwidth}{!}{
\begin{tabular}{llccccccccccc}
\toprule
\multirow{2}{*}{\textbf{Method}} 
& \multirow{2}{*}{\textbf{Size}}
& \multicolumn{5}{c}{\textbf{Web-search reasoning}}
& \multicolumn{5}{c}{\textbf{Mathematical reasoning}} \\
\cmidrule(lr){3-7} \cmidrule(lr){8-13}
&
& \textbf{HotpotQA} & \textbf{2Wiki} & \textbf{MuSiQue}
& \textbf{Avg.} & $\boldsymbol{\Delta}$ \textbf{to Ref.}
& \textbf{AIME24} & \textbf{AMC23} & \textbf{GameOf24}
& \textbf{Avg.} & $\boldsymbol{\Delta}$ \textbf{to Ref.} \\
\midrule

\multicolumn{13}{l}{\textbf{\textit{Training-free / inference-time methods}} \quad Ref. = \textbf{\ours-Intervention (Infer.)}} \\
\hdashline

 GPT-4o~\citep{hurst2024gpt}
& $\sim$200B
& 54.0 & 49.5 & 24.0
& 42.5 & \cellcolor{cPos2}$\uparrow 13.7$
& 13.3 & 60.0 & 32.0
& 35.1 & \cellcolor{cPos1}$\uparrow 3.7$ \\

 GPT-4o-mini~\citep{hurst2024gpt}
& $\sim$8B
& 41.0 & 35.6 & 15.0
& 30.5 & \cellcolor{cPos3}$\uparrow 25.7$
& 13.3 & 57.5 & 16.0
& 28.9 & \cellcolor{cPos2}$\uparrow 9.9$ \\

 Qwen2.5-7B-Instruct~\citep{qwen2025qwen25technicalreport}
& 7B-Inst
& 21.0 & 23.0 & 6.0
& 16.7 & \cellcolor{cPos3}$\uparrow 39.5$
& 6.7 & 47.5 & 33.0
& 29.1 & \cellcolor{cPos2}$\uparrow 9.7$ \\

 RA-Agent (multi-loop)
& 7B-Inst
& 39.5 & 34.0 & 20.0
& 31.2 & \cellcolor{cPos3}$\uparrow 25.0$
& 6.7 & 50.0 & 46.0
& 34.2 & \cellcolor{cPos1}$\uparrow 4.6$ \\

Prompt-Only Skills
& 7B-Inst
& 28.0 & 26.0 & 7.5 & 20.5 & \cellcolor{cPos3}$\uparrow 35.7$ 
& 10.0 & 47.5 & 41.0 & 32.8 & \cellcolor{cPos1}$\uparrow 6.0$
\\

 Iter-RetGen~\citep{shao2023enhancing}
& 7B-Inst
& 22.0 & 25.9 & 6.6
& 18.2 & \cellcolor{cPos3}$\uparrow 38.0$
& -- & -- & --
& -- & -- \\

 AutoGen~\citep{wu2024autogen}
& 7B-Inst
& 50.0 & 44.0 & 15.9
& 36.6 & \cellcolor{cPos3}$\uparrow 19.6$
& 13.3 & 57.5 & 24.0
& 31.6 & \cellcolor{cPos2}$\uparrow 7.2$ \\

 TIR~\citep{xue2025simpletir}
& 7B-Inst
& -- & -- & --
& -- & --
& 10.0 & 50.0 & 33.3
& 31.1 & \cellcolor{cPos2}$\uparrow 7.7$ \\

\cdashline{1-13}[5pt/5pt]

\textbf{\ours-Intervention (PF-only)}
& 7B-Inst & 59.0 & 67.0 & 27.0 & 51.0 & \cellcolor{cPos1}$\uparrow 5.2$
& 6.7 & 55.0 & 46.0 & 35.9 & \cellcolor{cPos1}$\uparrow 2.9$
\\

 \textbf{\ours-Intervention (w. Teacher)}
& 7B-Inst
& \textbf{64.5} & \textbf{70.0} & \textbf{34.0}
& \textbf{56.2} & --
& \textbf{10.0} & \textbf{56.5} & \textbf{50.0}
& \textbf{38.8} & -- \\

\midrule
\multicolumn{13}{l}{\textbf{\textit{Training-based methods}} \quad Ref. = \textbf{\ours-Evolve + RS}} \\
\hdashline

 SFT (vanilla)
& 7B-Inst
& 22.0 & 25.9 & 6.6
& 18.2 & \cellcolor{cPos3}$\uparrow 42.1$
& 6.7 & 47.5 & 33.0
& 29.1 & \cellcolor{cPos3}$\uparrow 16.3$ \\

 SKILL0~\citep{lu2026skill0incontextagenticreinforcement}
& 7B-Inst
& 40.0 & 38.3 & 16.4
& 31.6 & \cellcolor{cPos3}$\uparrow 28.7$
& -- & -- & --
& -- & -- \\

 Search-R1~\citep{jin2025search}
& 7B-Inst
& 37.0 & 38.2 & 14.6
& 29.9 & \cellcolor{cPos3}$\uparrow 30.4$
& -- & -- & --
& -- & -- \\

 ZeroSearch~\citep{sun2025zerosearch}
& 7B-Base
& 34.6 & 35.2 & 18.0
& 29.3 & \cellcolor{cPos3}$\uparrow 31.0$
& -- & -- & --
& -- & -- \\

 ReSearch~\citep{chen2025research}
& 7B-Base
& 43.5 & 47.6 & 22.3
& 37.8 & \cellcolor{cPos3}$\uparrow 22.5$
& -- & -- & --
& -- & -- \\

 StepSearch~\citep{wang2025stepsearch}
& 7B-Base
& 38.6 & 36.6 & 22.6
& 32.6 & \cellcolor{cPos3}$\uparrow 27.7$
& -- & -- & --
& -- & -- \\

 VerlTool~\citep{jiang2025verltool}
& 7B-Base
& 44.8 & 45.3 & 19.3
& 36.5 & \cellcolor{cPos3}$\uparrow 23.8$
& -- & -- & --
& -- & -- \\

 SimpleRL-reason~\citep{zeng2025simplerl}
& 7B-Base
& -- & -- & --
& -- & --
& 16.7 & 60.0 & 33.0
& 36.6 & \cellcolor{cPos2}$\uparrow 8.8$ \\

 Open-Reasoner-Zero~\citep{hu2025open}
& 7B-Base
& -- & -- & --
& -- & --
& 16.7 & 54.9 & 32.0
& 34.5 & \cellcolor{cPos2}$\uparrow 10.9$ \\

 General-Reasoner~\citep{ma2025general}
& 7B-Base
& -- & -- & --
& -- & --
& 13.3 & 55.0 & 33.0
& 33.8 & \cellcolor{cPos2}$\uparrow 11.6$ \\

 Luffy~\citep{yan2025learning}
& 7B-Inst
& -- & -- & --
& -- & --
& 30.7 & 44.8 & 33.0
& 36.2 & \cellcolor{cPos2}$\uparrow 9.2$ \\

 ToRL~\citep{li2025torl}
& 7B-Inst
& -- & -- & --
& -- & --
& 20.0 & 60.0 & 31.0
& 37.0 & \cellcolor{cPos2}$\uparrow 8.4$ \\

 AgentFlow (w/ Flow-GRPO)~\citep{li2025intheflowagenticoptimizationeffective}
& 7B-Inst
& 57.0 & 77.2 & 25.3
& 53.2 & \cellcolor{cPos1}$\uparrow 7.1$
& 40.0 & 61.5 & 53.0
& 51.5 & \cellcolor{cNeg1}$\downarrow 6.1$ \\

\cdashline{1-13}[5pt/5pt]

 \textbf{\ours-Evolve + RS}
& 7B-Inst
& \textbf{69.0} & \textbf{74.0} & \textbf{38.0}
& \textbf{60.3} & --
& \textbf{16.7} & \textbf{57.5} & \textbf{62.0}
& \textbf{45.4} & -- \\

\bottomrule
\end{tabular}}
{\raggedright\tiny
\textit{\textbf{Note:}} For each domain, we report per-dataset accuracy, domain average, and $\Delta$ to the reference method in the same block.
In the training-free block, the reference is \ours-Intervention (Infer.); in the training-based block, the reference is \ours-Evolve + RS.
Methods without results on a given domain are marked as ``--''.\par}
\vspace{-2mm}
\caption{
Main results on web-search and mathematical reasoning under training-free and training-based settings.
}
\vspace{-2mm}
\label{tab:main_search_math_combined}
\end{table*}

\begin{table*}[!ht]
\centering
\tiny
\renewcommand{\arraystretch}{1.08}
\resizebox{1.0\textwidth}{!}{
\begin{tabular}{lcccccccccc}
\toprule
\multirow{2}{*}{\textbf{Method}}
& \multirow{2}{*}{\textbf{Size}}
& \multirow{2}{*}{\textbf{Metric}}
& \multicolumn{2}{c}{\textbf{HumanEval}}
& \multicolumn{2}{c}{\textbf{MBPP}}
& \multicolumn{2}{c}{\textbf{BigCodeBench}}
& \multirow{2}{*}{\textbf{Avg.}}
& \multirow{2}{*}{$\boldsymbol{\Delta}$ \textbf{to Ref.}} \\
\cmidrule(lr){4-5} \cmidrule(lr){6-7} \cmidrule(lr){8-9}
& & & \textbf{Base} & \textbf{Plus} & \textbf{Base} & \textbf{Plus} & \textbf{Full} & \textbf{Hard} & & \\
\midrule

\multicolumn{11}{l}{\textbf{\textit{Training-free / inference-time methods}} \quad Ref. = \textbf{\ours-Intervention (Infer.)}} \\
\hdashline

GPT-4o~\citep{hurst2024gpt}
& $\sim$200B & pass@1
& 90.0 & 81.0 & 91.0 & 71.0 & 41.0 & 9.0
& 63.8 & \cellcolor{cPos1}$\uparrow 4.8$ \\

GPT-4o-mini~\citep{hurst2024gpt}
& $\sim$8B & pass@1
& 87.0 & 80.0 & 88.0 & 67.0 & 32.0 & 12.0
& 61.0 & \cellcolor{cPos1}$\uparrow 7.7$ \\

Qwen2.5-7B-Instruct~\citep{qwen2025qwen25technicalreport}
& 7B-Inst & pass@1
& 81.7 & 73.2 & 79.4 & 67.7 & 45.6 & 16.9
& 60.8 & \cellcolor{cPos1}$\uparrow 7.9$ \\

RA-Agent (multi-loop)
& 7B-Inst & pass@1
& 76.0 & 62.0 & 82.0 & 62.0 & 31.0 & 14.0
& 54.5 & \cellcolor{cPos2}$\uparrow 14.2$ \\

Prompt-Only Skills
& 7B-Inst & pass@1
& 86.0 & 75.0 & 86.0 & 66.0 & 40.0 & 14.0 & 61.2 & \cellcolor{cPos1}$\uparrow 7.5$
\\

\cdashline{1-11}[5pt/5pt]
\addlinespace[2pt]

\textbf{\ours-Intervention (PF-only)}
& 7B-Inst & pass@1 & 84.5 & 82.0 & 88.0 & 65.0 & 42.0 & 19.0 & 63.4 & \cellcolor{cPos1}$\uparrow 5.3$

\\

\textbf{\ours-Intervention (w. Teacher)}
& 7B-Inst & pass@1
& 88.0 & 85.0 & 91.0 & 73.0 & 49.0 & 26.0
& \cellcolor{cZero}\textbf{68.7} & -- \\

\midrule

\multicolumn{11}{l}{\textbf{\textit{Training-based methods}} \quad Ref. = \textbf{\ours-Evolve + RS}} \\
\hdashline

SFT (vanilla)
& 7B-Inst & pass@1
& 82.0 & 77.0 & 76.0 & 63.0 & 34.0 & 13.0
& 57.5 & \cellcolor{cPos2}$\uparrow 12.4$ \\

AceCoderRM~\citep{zeng2025acecoderacingcoderrl}
& 7B-Inst & pass@1
& 83.5 & 77.4 & 83.1 & 71.2 & 46.8 & 16.9
& 63.2 & \cellcolor{cPos1}$\uparrow 6.7$ \\

AceCoderRule~\citep{zeng2025acecoderacingcoderrl}
& 7B-Inst & pass@1
& 84.1 & 77.4 & 80.2 & 68.3 & 46.8 & 15.5
& 62.1 & \cellcolor{cPos1}$\uparrow 7.8$ \\

GRPO (Code)~\citep{fan2025posteriorgrporewardingreasoningprocesses}
& 7B-Inst & pass@1
& 85.9 & 81.1 & 86.7 & 75.1 & 52.0 & 29.7
& 68.4 & \cellcolor{cPos1}$\uparrow 1.5$ \\

P-GRPO (Code+RM)~\citep{fan2025posteriorgrporewardingreasoningprocesses}
& 7B-Inst & pass@1
& 86.6 & 81.1 & 87.0 & 76.2 & 54.0 & 33.8
& 69.8 & \cellcolor{cPos1}$\uparrow 0.1$ \\

KodCode-RL Step 128~\citep{xu2025kodcodediversechallengingverifiable}
& 7B-Inst & pass@1
& 90.2 & 83.5 & 81.0 & 70.9 & 47.0 & 19.6
& 65.4 & \cellcolor{cPos1}$\uparrow 4.5$ \\

KodCode-RL Step 256~\citep{xu2025kodcodediversechallengingverifiable}
& 7B-Inst & pass@1
& 91.5 & 86.0 & 82.8 & 72.8 & 48.2 & 19.6
& 66.8 & \cellcolor{cPos1}$\uparrow 3.1$ \\

\cdashline{1-11}[5pt/5pt]
\addlinespace[2pt]

\textbf{\ours-Evolve + RS}
& 7B-Inst & pass@1
& 91.0 & 84.5 & 92.0 & 74.0 & 50.0 & 28.0
& \cellcolor{cZero}\textbf{69.9} & -- \\

\bottomrule
\end{tabular}}
{\raggedright\tiny
\textit{\textbf{Note:}} For each block, $\Delta$ is computed against the reference method in that block:
\ours-Intervention (Infer.) for training-free methods and \ours-Evolve + RS for training-based methods.
All of our methods use Qwen2.5-7B-Instruct as the backbone unless otherwise specified.\par}
\vspace{-2mm}
\caption{
Main results on coding under training-free and training-based settings.}
\vspace{-4mm}
\label{tab:code_combined}
\end{table*}

We evaluate \ours along three axes: whether PFs improve inference-time decisions, whether PF-corrected traces help post-training, and whether remaining failures can be turned into new PFs through filtered evolution.

\noindent\textbf{Tasks and metrics.}
We report accuracy for web-search and mathematical reasoning, and pass@1 for coding. For web-search reasoning, we use HotpotQA~\citep{yang2018hotpotqadatasetdiverseexplainable}, 2Wiki~\citep{xanh2020_2wikimultihop}, and MuSiQue~\citep{trivedi2022musiquemultihopquestionssinglehop}. For mathematical reasoning, we use AIME24~\citep{aime24}, AMC23~\citep{AMC23}, and GameOf24~\citep{24game}, with answers judged by GPT-4o~\citep{hurst2024gpt}. We evaluate Coding on HumanEval~(\textsc{Base}, \textsc{Plus})~\cite{chen2021evaluating}, MBPP~(\textsc{Base}, \textsc{Plus})~\cite{austin2021program}, and BigCodeBench~(\textsc{Full}, \textsc{Hard})~\cite{zhuo2024bigcodebench}.\footnote{Test-set and training-pool for web-search and math follow AgentFlow verbatim; for coding we adopt the same training-data protocol. Details are reported in Appendix~\ref{app:datasets}.}

\noindent\textbf{Backbone and agent setup.}
Unless otherwise specified, all methods use
\texttt{Qwen/Qwen2.5-7B-Instruct}~\citep{qwen2025qwen25technicalreport}
as the base policy and share the same multi-step ReAct agent setup described in
Section~\ref{sec:method}.
For inference-time intervention, we evaluate two variants:
\texttt{\ours-Intervention (PF-only)}, which retrieves, activates, and executes PFs
without a teacher, and
\texttt{\ours-Intervention (w. Teacher)}, where the teacher is used only to select
among candidate PFs, while the selected PF performs the actual intervention.
The initial skill library is derived from recurring recovered failure--repair
patterns in the training pool; full PF families and examples are provided in
Appendix~\ref{app:exp_details}.

\begin{table*}[!ht]
\centering
\tiny
\renewcommand{\arraystretch}{1.06}
\vspace{-1mm}
\resizebox{\linewidth}{!}{
\begin{tabular}{lccccccccccccc}
\toprule
\textbf{Setting}
& \textbf{PF}
& \textbf{Teacher}
& \textbf{T}
& \textbf{M}
& \textbf{C}
& \textbf{O}
& \textbf{Exec}
& \textbf{Teach}
& \textbf{HotpotQA}
& \textbf{2Wiki}
& \textbf{MuSiQue}
& \textbf{Avg.}
& $\boldsymbol{\Delta}$ \\
\midrule

\multicolumn{14}{l}{\textbf{\textit{Inference-time component ablation}}} \\
Full \ours-Intervention (Infer.)
  & \checkmark & \checkmark & \checkmark & \checkmark & \checkmark & \checkmark & N/A & N/A
  & \cellcolor{cZero}\textbf{64.5}
  & \cellcolor{cZero}\textbf{70.0}
  & \cellcolor{cZero}\textbf{34.0}
  & \cellcolor{cZero}\textbf{56.2}
  & -- \\
RA-Agent (multi-loop)
  & -- & -- & \checkmark & \checkmark & \checkmark & \checkmark & N/A & N/A
  & \cellcolor{cNeg3}39.5
  & \cellcolor{cNeg3}34.0
  & \cellcolor{cNeg2}20.0
  & \cellcolor{cNeg3}31.2
  & \cellcolor{cNeg3}$\downarrow 25.0$ \\

PF only
  & \checkmark & -- & \checkmark & \checkmark & \checkmark & \checkmark & N/A & N/A
  & \cellcolor{cNeg2}59.0
  & \cellcolor{cNeg1}67.0
  & \cellcolor{cNeg2}27.0
  & \cellcolor{cNeg2}51.0
  & \cellcolor{cNeg2}$\downarrow 5.2$ \\

Teacher only
  & -- & \checkmark & \checkmark & \checkmark & \checkmark & \checkmark & N/A & N/A
  & \cellcolor{cNeg2}59.0
  & \cellcolor{cNeg1}69.0
  & \cellcolor{cNeg2}24.0
  & \cellcolor{cNeg2}50.7
  & \cellcolor{cNeg2}$\downarrow 5.5$ \\

\midrule
\multicolumn{14}{l}{\textbf{\textit{Signal ablation (on E5)}}} \\
All four signals
  & \checkmark & \checkmark
  & \checkmark & \checkmark & \checkmark & \checkmark
  & \checkmark & \checkmark
  & \cellcolor{cZero}\textbf{69.0}
  & \cellcolor{cZero}\textbf{74.0}
  & \cellcolor{cZero}\textbf{38.0}
  & \cellcolor{cZero}\textbf{60.3}
  & -- \\
w/o Timing
  & \checkmark & \checkmark
  & -- & \checkmark & \checkmark & \checkmark
  & \checkmark & \checkmark
  & \cellcolor{cNeg1}68.0
  & \cellcolor{cNeg2}60.0
  & \cellcolor{cNeg2}29.5
  & \cellcolor{cNeg2}52.5
  & \cellcolor{cNeg2}$\downarrow 7.8$ \\

w/o mode
  & \checkmark & \checkmark
  & \checkmark & -- & \checkmark & \checkmark
  & \checkmark & \checkmark
  & \cellcolor{cNeg2}55.0
  & \cellcolor{cNeg3}52.0
  & \cellcolor{cNeg2}27.5
  & \cellcolor{cNeg3}44.8
  & \cellcolor{cNeg3}$\downarrow 15.5$ \\

w/o Correctness
  & \checkmark & \checkmark
  & \checkmark & \checkmark & -- & \checkmark
  & \checkmark & \checkmark
  & \cellcolor{cNeg2}62.0
  & \cellcolor{cNeg3}53.0
  & \cellcolor{cNeg2}29.5
  & \cellcolor{cNeg2}48.2
  & \cellcolor{cNeg2}$\downarrow 12.1$ \\

w/o Outcome
  & \checkmark & \checkmark
  & \checkmark & \checkmark & \checkmark & --
  & \checkmark & \checkmark
  & \cellcolor{cNeg2}55.5
  & \cellcolor{cNeg3}59.0
  & \cellcolor{cNeg2}28.0
  & \cellcolor{cNeg2}47.5
  & \cellcolor{cNeg2}$\downarrow 12.8$ \\

\midrule
\multicolumn{14}{l}{\textbf{\textit{Filtering ablation (on E5)}}} \\
Evolution, full filtering
  & \checkmark & \checkmark
  & \checkmark & \checkmark & \checkmark & \checkmark
  & \checkmark & \checkmark
  & \cellcolor{cZero}\textbf{69.0}
  & \cellcolor{cZero}\textbf{74.0}
  & \cellcolor{cZero}\textbf{38.0}
  & \cellcolor{cZero}\textbf{60.3}
  & -- \\
No evolution
  & \checkmark & \checkmark
  & \checkmark & \checkmark & \checkmark & \checkmark
  & -- & --
  & \cellcolor{cPos1}69.5
  & \cellcolor{cNeg1}72.5
  & \cellcolor{cNeg1}36.0
  & \cellcolor{cNeg1}59.3
  & \cellcolor{cNeg1}$\downarrow 1.0$ \\

Evolution, no filtering
  & \checkmark & \checkmark
  & \checkmark & \checkmark & \checkmark & \checkmark
  & -- & --
  & \cellcolor{cNeg3}35.5
  & \cellcolor{cNeg3}58.0
  & \cellcolor{cNeg3}15.5
  & \cellcolor{cNeg3}36.3
  & \cellcolor{cNeg3}$\downarrow 24.0$ \\

Evolution, exec-only
  & \checkmark & \checkmark
  & \checkmark & \checkmark & \checkmark & \checkmark
  & \checkmark & --
  & \cellcolor{cNeg2}61.0
  & \cellcolor{cNeg2}61.0
  & \cellcolor{cNeg2}24.5
  & \cellcolor{cNeg2}48.8
  & \cellcolor{cNeg2}$\downarrow 11.5$ \\

Evolution, teacher-only
  & \checkmark & \checkmark
  & \checkmark & \checkmark & \checkmark & \checkmark
  & -- & \checkmark
  & \cellcolor{cNeg2}59.0
  & \cellcolor{cNeg3}57.0
  & \cellcolor{cNeg2}25.5
  & \cellcolor{cNeg2}47.2
  & \cellcolor{cNeg2}$\downarrow 13.1$ \\
\bottomrule
\end{tabular}}
{\raggedright\tiny
\textit{\textbf{Note:}} T/M/C/O denote timing, mode, correctness, and outcome; Exec/Teach denote executable validation and teacher review. \par}
\vspace{-2mm}
\caption{
Ablations on web-search reasoning. $\Delta$ is computed relative to the first row of each block. }
\vspace{-4mm}
\label{tab:ablation_all}
\end{table*}

\begin{table}[!t]
\centering
\tiny
\renewcommand{\arraystretch}{1.08}
\setlength{\tabcolsep}{3.0pt}

\resizebox{\columnwidth}{!}{%
\begin{tabular}{@{}l c c c c c c@{}}
\toprule
\multicolumn{7}{c}{\textbf{Web-search reasoning}} \\
\midrule
\textbf{Method} & \textbf{Size}
& \textbf{HotpotQA} & \textbf{2Wiki} & \textbf{MuSiQue}
& \textbf{Avg.} & $\boldsymbol{\Delta}$ \textbf{vs PFs} \\
\midrule

\textbf{\ours-Intervention (Infer.)}
& 7B-Inst
& 64.5 & 70.0 & 34.0
& 56.2 & -- \\

\hdashline

\textbf{\ours + SFT}
& 7B-Inst
& 66.0 & 69.0 & 35.5
& 56.8 & \cellcolor{cPos1}$+0.6$ \\

\textbf{\ours + RS}
& 7B-Inst
& 69.5 & 72.5 & 36.0
& 59.3 & \cellcolor{cPos2}$+3.1$ \\

\textbf{\ours + OPD}
& 7B-Inst
& 64.5 & 69.0 & 54.0
& \cellcolor{cPos3}62.5 & \cellcolor{cPos3}\textbf{$+6.3$} \\

\textbf{\ours-Evolve + SFT}
& 7B-Inst
& 68.5 & 70.0 & 37.0
& 58.5 & \cellcolor{cPos2}$+2.3$ \\

\textbf{\ours-Evolve + RS}
& 7B-Inst
& 69.0 & 74.0 & 38.0
& 60.3 & \cellcolor{cPos3}$+4.1$ \\

\textbf{\ours-Evolve + OPD}
& 7B-Inst
& 65.0 & 69.5 & 35.5
& 56.7 & \cellcolor{cPos1}$+0.5$ \\

\midrule
\multicolumn{7}{c}{\textbf{Mathematical reasoning}} \\
\midrule
\textbf{Method} & \textbf{Size}
& \textbf{AIME24} & \textbf{AMC23} & \textbf{GameOf24}
& \textbf{Avg.} & $\boldsymbol{\Delta}$ \textbf{vs PFs} \\
\midrule

\textbf{\ours-Intervention (Infer.)}
& 7B-Inst
& 10.0 & 56.5 & 50.0
& 38.8 & -- \\

\hdashline

\textbf{\ours + SFT}
& 7B-Inst
& 13.3 & 58.5 & 51.0
& 40.9 & \cellcolor{cPos2}$+2.1$ \\

\textbf{\ours + RS}
& 7B-Inst
& 16.7 & 57.5 & 54.0
& 42.7 & \cellcolor{cPos2}$+3.9$ \\

\textbf{\ours + OPD}
& 7B-Inst
& 16.7 & 57.5 & 53.0
& 42.4 & \cellcolor{cPos2}$+3.6$ \\

\textbf{\ours-Evolve + SFT}
& 7B-Inst
& 13.3 & 59.0 & 52.0
& 41.4 & \cellcolor{cPos2}$+2.6$ \\

\textbf{\ours-Evolve + RS}
& 7B-Inst
& 16.7 & 57.5 & 62.0
& \cellcolor{cPos3}45.4 & \cellcolor{cPos3}$+6.6$ \\

\textbf{\ours-Evolve + OPD}
& 7B-Inst
& 16.7 & 60.0 & 55.0
& 43.9 & \cellcolor{cPos3}$+5.1$ \\

\bottomrule
\end{tabular}%
}

{\raggedright\tiny
\textit{\textbf{Note:}} For each domain, we report per-dataset accuracy, domain average, and improvement over the inference-time PF system, \ours-Intervention (Infer.).
For both reasoning domains, $\Delta$ is computed against the corresponding average of \ours-Intervention (Infer.). \par}

\vspace{-3mm}
\caption{
Ablation of \ours training and evolution strategies.
We report results on six \ours variants: fixed-library SFT/RS/OPD and their evolved counterparts.
}
\vspace{-5mm}
\label{tab:six_training}
\end{table}

\noindent\textbf{Training setup and baselines.}
Our main trained variant is {\ours-Evolve + RS}, combining closed-loop PF evolution
with PF-conditioned rejection sampling.
For ablation, we evaluate six post-training settings:
fixed-library \ours + SFT/RS/OPD and closed-loop
\ours-Evolve + SFT/RS/OPD.
SFT and RS train on PF-corrected trajectories without requiring teacher-generated
targets, whereas in OPD, GPT-4o provides reference trajectories while PF
interventions remain active.
When skill evolution is enabled, teacher review is additionally used to filter
candidate PFs before admission to the library.
Across all post-training variants of \ours, we use a LoRA-based training setup;
implementation details are provided in Appendix~\ref{app:exp_details}.
Web-search and mathematical baselines follow
AgentFlow~\citep{li2025intheflowagenticoptimizationeffective}, while coding
baselines are taken from reported results or evaluated under our coding split.

\subsection{Main Results: PF Intervention, Selection, and Internalization}

We present the main results in four stages: whether executable PFs alone improve inference-time decisions, whether auxiliary teacher selection further improves PF dispatch, whether PF-derived events can be internalized through post-training, and whether closed-loop PF evolution further improves the external skill library.

\begin{table*}[!ht]
\centering
\scriptsize
\setlength{\tabcolsep}{4pt}
\renewcommand{\arraystretch}{1.15}
\resizebox{\textwidth}{!}{
\begin{tabular}{l l c l c}
\toprule
\textbf{Recovered failure family} & \textbf{Representative PF} & \textbf{Recovered proportion} & \textbf{Typical pattern} & \textbf{Avg. $\Delta$EM / $\Delta$Acc.} \\
\midrule
Reasoning drift / unsupported finalization
& \texttt{reasoning\_hallucination}
& $31.7\%$
& \texttt{FINAL} $\rightarrow$ \texttt{SEARCH(question)}
& $+13.3$ / $+9.3$ \\

Premature finalization
& \texttt{premature\_final}
& $17.8\%^{\dagger}$
& early \texttt{FINAL} $\rightarrow$ \texttt{SEARCH}
& $+11.1$ / $+4.4$ \\

Insufficient evidence before final
& \texttt{no\_read\_before\_final}
& $17.8\%^{\dagger}$
& no-read \texttt{FINAL} $\rightarrow$ \texttt{SEARCH(verify)}
& $+0.4$ / $+0.8$ \\

Wrong entity focus
& \texttt{wrong\_entity\_focus}
& $24.1\%$
& wrong entity $\rightarrow$ \texttt{SEARCH(entity)}
& $+6.7$ / $+5.0$ \\

Harmful PF override
& \texttt{pf\_override\_harmful}
& $8.7\%$
& override $\rightarrow$ fallback \texttt{SEARCH}
& $-0.8$ / $-0.8$ \\

\midrule
\multicolumn{5}{c}{\textbf{Generated skill review scores (scale 0--1)}} \\
\midrule
\textbf{Concept} & \textbf{Trigger} & \textbf{Intervention} & \textbf{Executability} & \textbf{Validation / Avg.} \\
0.83 & 0.75 & 0.83 & 0.95 & 0.80 / 0.83 \\
\bottomrule
\multicolumn{5}{l}{\footnotesize $^{\dagger}$ joint proportion of the \emph{exploration\_control} cluster, which contains both PFs.} \\
\end{tabular}}
\vspace{-3mm}
\caption{
Recovered failure structure and utility of representative evolved PF families on the web-search benchmark.
The lower block reports average review scores of generated skills. 
}
\vspace{-2mm}
\label{tab:failure-evolution-merged}
\end{table*}
\begin{figure*}[!t]
\centering
\includegraphics[width=0.95\linewidth]{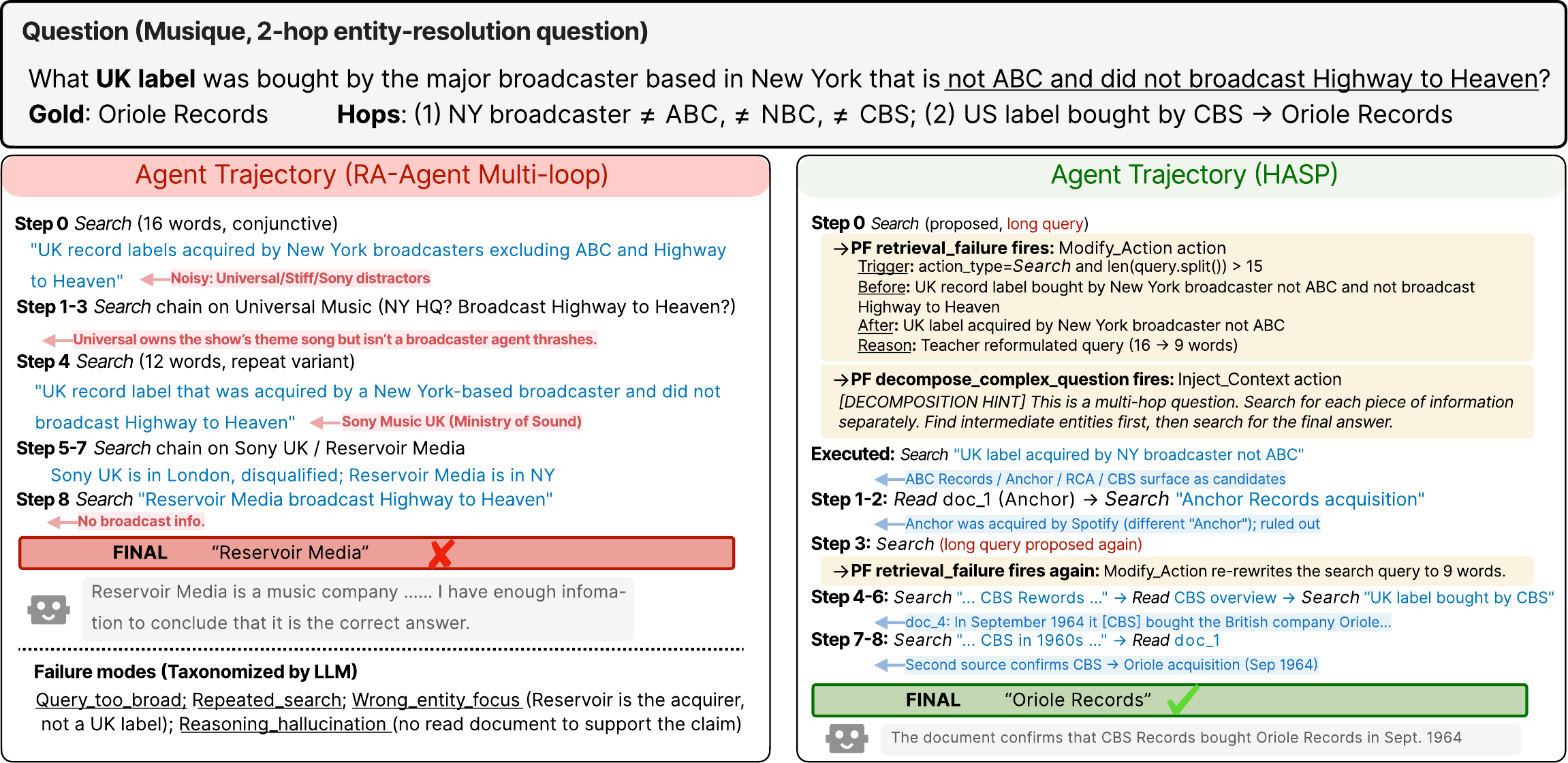}
\vspace{-2mm}
\caption{
Case study on a MuSiQue two-hop entity-resolution question. 
}
\vspace{-4mm}
\label{fig:case_study}
\end{figure*}

\noindent\textbf{Inference-time PF intervention outperforms strong baselines.}
The upper block of Table~\ref{tab:main_search_math_combined} and Table~\ref{tab:code_combined} compare PF-only intervention and PF intervention with auxiliary teacher selection. PF-only already outperforms the multi-loop ReAct agent and Prompt-Only Skills, showing that directly changing actions or adding corrective context is more effective than injecting skill text alone. On web-search reasoning, PF-only improves average accuracy to 51.0\%, compared with 20.5\% for Prompt-Only Skills. On coding, PF-only reaches 63.4\% average pass@1. On mathematical reasoning, PF-only reaches 35.9\%, compared with 32.8\% for Prompt-Only Skills. Adding auxiliary teacher selection further improves performance to 56.2\% on web-search reasoning, 38.8\% on mathematical reasoning, and 68.7\% average pass@1 on coding. 

\noindent\textbf{PF-derived scores provide effective supervision for post-training.}
Table~\ref{tab:six_training} evaluates whether PF-corrected traces improve the trained student under the same PF-guided pipeline. Under a fixed skill library, all three post-training variants improve over inference-time intervention alone. On web-search reasoning, performance increases to 56.8\%, 59.3\%, and 62.5\% for SFT, RS, and OPD, respectively, while on mathematical reasoning, the corresponding increases are from 38.8\% to 40.9\%, 42.7\%, and 42.4\%. Training dynamics shown in Figure~\ref{fig:training_dynamics} of App~\ref{app:training_dynamics} 
are consistent with this pattern: SFT is stable but gives smaller gains, RS offers the best overall balance, and OPD reaches the strongest fixed-library web-search result. Relative to prior training-based baselines, PF-guided post-training is already competitive with strong training-based baselines, especially on web-search reasoning.

\noindent\textbf{Closed-loop PF evolution further improves the external skill library.}
Table~\ref{tab:six_training} evaluates closed-loop evolution, where failure cases are summarized into candidate PFs and filtered before updating the external library. Our main trained variant, \ours-Evolve + RS, further improves over fixed-library RS to 60.3\% on web-search reasoning and 45.4\% on mathematical reasoning, and reaches 69.9\% average pass@1 on coding. In contrast, \ours-Evolve + OPD drops to 56.7\% on web-search reasoning, showing that updating the model and the skill library together can be less stable. Training dynamics are consistent with this pattern: fixed-library runs are smoother, while closed-loop evolution introduces more variation, which works better with RS than with OPD. Overall, \ours-Evolve + RS gives the strongest overall results among our open-weight settings, while staying competitive with GRPO and KodCode-RL on coding.

\noindent\textbf{PF intervention improves efficiency alongside accuracy.}
Executable PFs improve performance without increasing inference overhead. On web-search reasoning, \ours-Intervention (PF-only) reduces average token usage from 23.6K to 20.0K, latency from 11.8s to 10.0s, and estimated cost from \$0.0495 to \$0.0418 per query relative to the ReAct agent, while improving accuracy from 31.2\% to 51.0\%. Teacher-assisted \ours remains similarly selective, requiring only 1.1 teacher calls per query compared with 3.2 for Teacher-only, while achieving higher accuracy (56.2\% vs.\ 50.7\%). Thus, the gains arise from targeted intervention at failure-prone states rather than longer trajectories or persistent teacher supervision.

\subsection{Mechanism Analysis: How PFs Intervene and Why They Work}

PFs intervene through two main mechanisms: directly revising the next action and injecting corrective context into the reasoning process. Across all fired PF events on web-search reasoning, 65.1\% are action-level interventions and 34.9\% are context-level interventions, showing that \ours frequently changes the next executable decision itself rather than only adding textual guidance.

\noindent\textbf{Intervention patterns are concentrated in a few high-value skills.}
Trigger statistics over all PF activations in multi-step ReAct rollouts show that PF use is highly concentrated. The most frequently activated skills are \textit{decompose\_complex\_question}, \textit{insufficient\_exploration}, and \textit{answer\_completeness}, and MuSiQue accounts for far more activations than HotpotQA or 2Wiki, indicating that harder multi-hop settings rely more heavily on PF intervention. The intervention patterns are also highly regular: most action-level PFs delay premature \texttt{FINAL} decisions by redirecting them into additional \texttt{SEARCH} steps, while most context-level PFs provide decomposition or completeness guidance.

\noindent\textbf{Structured signals and strict filtering are both necessary.}
Table~\ref{tab:ablation_all} shows that PF effectiveness depends on more than final-answer reward alone. Full \ours-Intervention reaches 56.2\%, while removing timing, mode, correctness, and outcome causes drops of 7.8, 15.5, 12.1, and 12.8 points. The same table also shows that strict filtering is essential for stable skill-library evolution: full filtering reaches 60.3\%, whereas unfiltered evolution drops sharply to 36.3\%; using only executable filtering or only teacher-side filtering yields intermediate results of 48.8\% and 47.2\%. 

\noindent\textbf{Case-level evidence and failure structure support the PF design.}
Figure~\ref{fig:case_study} shows how PFs recover a failure-prone MuSiQue trajectory by rewriting an over-constrained query and injecting a decomposition hint. Table~\ref{tab:failure-evolution-merged} shows that recovered cases concentrate in a small number of recurring failure types, and that the most useful evolved PFs align closely with these high-frequency recoverable patterns, whereas overly specific or weakly reusable PFs contribute little or can even be harmful. We also study skills that cannot be converted into executable PFs and are therefore added to the skill library only as textual guidance. Their gains are limited and mixed, indicating that simply enlarging the library is not enough; the main benefit comes from skills that can be executed as PFs. Details are reported in Appendix~\ref{app:prompt_equivalent}.

\noindent\textbf{Skill internalization is strongest for behavior-correcting PFs.}
PFs that correct the model's own behavior, such as not reading enough or writing overly long queries, often become less necessary after training, whereas PFs tied to the question itself usually remain active. Quantitatively, \textit{multi\_hop\_reasoning\_failure} and \textit{retrieval\_failure} become silent on 100\% of their previously triggered cases, while \textit{decompose\_complex\_question} is rarely internalized. More broadly, only 8--38\% of the accuracy gain can be directly attributed to cases where those PFs no longer need to fire; the larger effect comes from broader behavior changes, including more SEARCH and READ actions. This suggests that post-training mainly internalizes behavior-correcting PFs and a stronger overall reasoning style, while input-dependent PFs continue to provide useful online guidance.

\section{Related Work}

Post-training has been widely used to improve search, reasoning, tool use, and coding agents. Search-oriented methods such as Search-R1~\citep{jin2025search}, ReSearch~\citep{chen2025research}, and VerlTool~\citep{jiang2025verltool} train models to interact more effectively with search or tool environments, while RL-based reasoning and coding methods further improve policy quality through reward-driven objectives~\citep{zeng2025simplerl,hu2025open,ma2025general,li2025torl,fan2025posteriorgrporewardingreasoningprocesses, liu2025medmmvcontrollablemultimodalmultiagent}. More broadly, LLM agents solve complex tasks by interleaving reasoning and acting~\citep{yao2023reactsynergizingreasoningacting}, using external tools~\citep{schick2023toolformerlanguagemodelsteach}, and extending these loops through richer agentic workflows~\citep{wu2024autogen,li2025intheflowagenticoptimizationeffective, hou2026biokdselectivephysiologytovideoknowledge}. Another prominent direction improves agent behavior by reusing past experience or skills. Methods such as Reflexion~\citep{shinn2023reflexionlanguageagentsverbal}, ExpeL~\citep{zhao2024expelllmagentsexperiential}, and Voyager~\citep{wang2023voyageropenendedembodiedagent} store verbal lessons or memories, while more recent self-improving systems such as MemSkill~\citep{zhang2026memskilllearningevolvingmemory}, SkillRL~\citep{xia2026skillrlevolvingagentsrecursive}, and SAGE~\citep{wang2026reinforcementlearningselfimprovingagent} study memory or skill evolution. Most prior methods reuse experience as prompt text, memory, or training signals. In contrast, \ours represents skills as executable state-action intervention functions that can be triggered inside the agent loop, while also supporting post-training and controlled skill evolution. Table~\ref{tab:related_work_comparison} compares \ours with representative prior work.

\begin{table}[!t]
\centering
\footnotesize
\setlength{\tabcolsep}{1.2pt}
\renewcommand{\arraystretch}{0.98}
\resizebox{\columnwidth}{!}{%
\begin{tabular}{%
@{}>{\raggedright\arraybackslash}m{0.50\columnwidth}
>{\centering\arraybackslash}m{0.26\columnwidth}
*{4}{>{\centering\arraybackslash}m{0.17\columnwidth}}@{}}

\rowcolor{myblue!40}
\textbf{Method}
& \begin{tabular}[c]{@{}c@{}}\textbf{Skill / Memory}\\\textbf{Form}\end{tabular}
& \begin{tabular}[c]{@{}c@{}}\textbf{Runtime}\\\textbf{Control}\end{tabular}
& \begin{tabular}[c]{@{}c@{}}\textbf{Learning}\\\textbf{Signal}\end{tabular}
& \begin{tabular}[c]{@{}c@{}}\textbf{Policy}\\\textbf{Training}\end{tabular}
& \begin{tabular}[c]{@{}c@{}}\textbf{Skill}\\\textbf{Evolution}\end{tabular} \\
\Xhline{1pt}

\rowcolor{gray!10}
\multicolumn{6}{l}{\textit{Skill-based / self-improving}} \\
ExpeL~\citep{zhao2024expelllmagentsexperiential}
  & \cmark & \(\circ\) & \(\circ\) & \xmark & \xmark \\
MemSkill~\citep{zhang2026memskilllearningevolvingmemory}
  & \cmark & \(\circ\) & \(\circ\) & \xmark & \cmark \\
SKILL0~\citep{lu2026skill0incontextagenticreinforcement}
  & \cmark & \xmark & \cmark & \cmark & \xmark \\
SkillRL~\citep{xia2026skillrlevolvingagentsrecursive}
  & \cmark & \(\circ\) & \cmark & \cmark & \cmark \\
SAGE~\citep{wang2026reinforcementlearningselfimprovingagent}
  & \cmark & \(\circ\) & \cmark & \cmark & \cmark \\

\midrule
\rowcolor{gray!10}
\multicolumn{6}{l}{\textit{General RL-based}} \\
AgentFlow~\citep{li2025intheflowagenticoptimizationeffective}
  & \(\circ\) & \(\circ\) & \cmark & \cmark & \(\circ\) \\
Search-R1/ReSearch~\citep{jin2025search,chen2025research}
  & \xmark & \(\circ\) & \cmark & \cmark & \xmark \\
P-GRPO~\citep{fan2025posteriorgrporewardingreasoningprocesses}
  & \xmark & \xmark & \cmark & \cmark & \xmark \\

\midrule
\rowcolor{myblue!15}
\textbf{\ours (Ours)}
  & \cmark & \cmark & \cmark & \cmark & \cmark \\
\bottomrule
\end{tabular}%
}

\vspace{-2mm}
\caption{Qualitative comparison of \ours with representative prior methods.
\cmark: explicit support; \(\circ\): partial or indirect support; \xmark: not directly supported.}
\label{tab:related_work_comparison}
\vspace{-4mm}
\end{table}

\section{Conclusion}

We presented \textbf{\ours}, a framework that represents skills as reusable state-action intervention functions. Instantiated as Program Functions (PFs), skills become executable modular that can directly intervene on the agent's next decision, provide supervision for post-training, and grow through validated updates to an external skill library. This shared interface allows PFs to \emph{act} during inference, \emph{teach} during post-training, and \emph{evolve} from recurring residual failures. Across web-search reasoning, mathematical reasoning, and coding tasks, \ours improves inference-time behavior, supports selective internalization without full reinforcement learning, and benefits from filtered skill evolution. Overall, our results suggest a complementary path for improving agents: instead of relying only on larger-scale policy optimization, skills can be made executable, explicitly triggered inside the agent loop, and gradually absorbed into both the policy and the external skill library.

\section*{Limitations}
\label{sec:limit}

\ours has several limitations. First, while the core PF-only setting does not require a teacher, some stronger variants use external teachers for PF selection, teacher review, or on-policy distillation, which increases cost and may introduce teacher-specific bias. Second, PFs mainly help the model use and absorb existing strategies rather than discover fundamentally new ones; as suggested by harder benchmarks such as AIME24, tasks requiring new strategy discovery may benefit more from RL-style optimization. Third, our evaluation focuses on benchmark-style search, math, and coding tasks with relatively clear verification signals, leaving transfer to open-ended, long-horizon, interactive, or weakly verifiable environments unresolved. Fourth, the current PF interface supports only a limited set of intervention forms, mainly action modification and context injection. This improves executability and auditability, but also limits how much PFs can intervene. Finally, self-improving evolution may still degrade over time if generated PFs become too narrow, repetitive, or teacher-biased, and the current PF scoring interface may need to be adapted in domains where local repairs do not reliably translate into better final outcomes. Compute footprint and broader-impact analysis are provided in Appendix~\ref{app:compute}.

\bibliography{ref}

\appendix
\newpage

\appendix


\section{Additional Details of \ours Framework}
\label{app:pf_framework}

\subsection{PF Interface and Intervention Types}

Each Program Function (PF) is a typed Python object exposing two methods:
\begin{itemize}
    \item \texttt{should\_activate(step\_context, action\_type, arg) $\rightarrow$ bool}
    \item \texttt{intervene(step\_context, action\_type, arg, teacher=None) $\rightarrow$ Intervention}
\end{itemize}
\texttt{should\_activate} decides whether the skill is relevant at the current decision point; \texttt{intervene} returns a structured \texttt{Intervention} object whose \texttt{type} field is one of:
\begin{itemize}
    \item \texttt{MODIFY\_ACTION}: rewrite or refine the next action (action type and/or argument);
    \item \texttt{INJECT\_CONTEXT}: append auxiliary text that the policy will see in the next observation;
    \item \texttt{NOOP}: abstain from changing the current decision (still emits an audit record).
\end{itemize}

PFs are by default deterministic and run on every step. A PF that sets \texttt{needs\_teacher = True} additionally receives a teacher-model handle in \texttt{intervene}; such PFs must still degrade gracefully when \texttt{teacher} is \texttt{None}. The runtime enforces a per-PF rate limit (\texttt{\_MAX\_MODIFY\_FIRES = 2} for \texttt{MODIFY\_ACTION}) and per-skill caps (e.g.\ \texttt{RetrievalFailurePF} fires at most three times per episode) to prevent oscillation. The same dispatcher serves the web-search, math, and code domains without modification, so cross-domain transfer is mechanical.

\subsection{Phase Instructions and Handlers}

Beyond step-level activation, each PF may expose two extra surface areas. \emph{Phase instructions}, declared in the YAML frontmatter of \texttt{SKILL.md}, attach prompt-level reminders to specific reasoning phases (\texttt{pre\_final}, \texttt{post\_search}, \texttt{post\_read}); the runtime renders them only when the phase conditions match. \emph{Handlers}, registered in \texttt{skill\_handlers.py}, are FINAL-time verifiers that can flip a FINAL action back to SEARCH (capped at one such override per episode). The handler family is enabled only in web-search; math and code use single-step rollouts where there is no follow-up step for an override to land on.

\subsection{An Example PF: Retrieval Failure}
\label{app:pf_example}

Skill~\ref{box:skill_md_retrieval} shows the full \texttt{SKILL.md} for the \texttt{retrieval\_failure} skill, and PF~\ref{box:pf_retrieval} shows the corresponding PF class. The skill is one of three always-on web-search PFs, alongside \texttt{format\_extraction\_error} and \texttt{relevant\_content\_extractor}, because the rest of the library depends on the agent producing a well-formed search action and a clean observation.

\begin{skillbox}[label={box:skill_md_retrieval}]{Adaptive Search Recovery (\texttt{retrieval\_failure})}
\begin{lstlisting}
---
skill_id: retrieval_failure
name: Adaptive Search Recovery
version: 2
priority: 0.8
error_category: retrieval_failure
applicable_modes: [all]
applicable_phases: [think, search]
system_summary: >
  When searches fail, reformulate queries with synonyms or
  decompose into sub-queries.
phases:
  post_search:
    conditions: [search_empty]
    priority_boost: 0.3
    action: reformulate_search
  pre_final:
    conditions: [no_read_yet]
    priority_boost: 0.2
    action: force_read_best_doc
---

# Adaptive Search Recovery

When initial searches fail, adapt by reformulating queries, using
synonyms, or decomposing complex questions into sub-queries.

## Detection Triggers
- Search returns no results or only irrelevant results
- Search query is too vague or too specific
- Multi-hop question requires intermediate entities not yet found

## Phase: post_search
No useful results. Try: (1) synonyms, (2) decompose multi-hop,
(3) broaden query. Do NOT proceed to FINAL without a relevant document.

## Phase: pre_final
STOP: You have not READ any document yet. Go back and READ at least
one relevant source before answering.
\end{lstlisting}
\end{skillbox}

\begin{pfbox}[label={box:pf_retrieval}]{\texttt{RetrievalFailurePF}}
\begin{lstlisting}[language=Python]
@register_pf("retrieval_failure")
class RetrievalFailurePF(ProgramFunction):
    needs_teacher = True
    _MAX_FIRES = 3

    def should_activate(self, ctx, action_type, arg):
        if action_type != "SEARCH":
            return False
        if len(arg.split()) <= 15:
            return False
        return ctx.get("_retrieval_failure_fires", 0) < self._MAX_FIRES

    def intervene(self, ctx, action_type, arg, model=None):
        ctx["_retrieval_failure_fires"] = ctx.get("_retrieval_failure_fires", 0) + 1
        if model is not None:
            clean = model.generate(
                messages=[{"role": "user", "content": (
                    f"A search agent is trying to answer this question:\n"
                    f"Question: {ctx.get('question','')}\n\n"
                    f"The agent generated this overly long search query:\n"
                    f"\"{arg[:500]}\"\n\n"
                    f"Extract a concise, effective search query "
                    f"(5-12 words) that captures the key search intent. "
                    f"Reply with ONLY the query.")}],
                max_tokens=50, temperature=0.0).strip().strip('"')
            if 1 < len(clean.split()) <= 20:
                return Intervention(
                    type=InterventionType.MODIFY_ACTION,
                    new_action_type="SEARCH",
                    new_action_arg=clean,
                    reason="Reformulated long query")
        # Deterministic fallback: extract quoted span / first short line
        return Intervention(
            type=InterventionType.MODIFY_ACTION,
            new_action_type="SEARCH",
            new_action_arg=self._smart_shorten(arg, ctx.get("question","")),
            reason="Smart-shortened query")
\end{lstlisting}
\end{pfbox}

A math-domain example, \texttt{verification\_missing}, is shown in Skill~\ref{box:skill_md_verify}; it is the highest-priority skill in the math library (priority $0.90$, always active on \texttt{pre\_final}).

\begin{skillbox}[label={box:skill_md_verify}]{Solution Verification (\texttt{verification\_missing}, math)}
\begin{lstlisting}
---
skill_id: verification_missing
name: Solution Verification
version: 1
priority: 0.90
error_category: verification_missing
applicable_modes: [all]
applicable_phases: [answer]
system_summary: >
  Before finalizing, plug the candidate answer back into the
  original problem to verify it satisfies all constraints.
phases:
  pre_final:
    conditions: []
    priority_boost: 0.5
    action: verify_solution
---

# Solution Verification

Substitute the candidate answer back into the original equation /
problem statement and confirm every constraint is satisfied.

## Phase: pre_final
Substitute your candidate answer back into each constraint of the
original problem. Does it satisfy every one? If not, recompute.

## Examples
**Scenario:** Solve sqrt(x+6) = x. Candidate: x in {-2, 3}.
**Wrong:** FINAL({-2, 3}) without checking.
**Correct:** Check x=-2: sqrt(4)=2 != -2 (extraneous).
            Check x=3:  sqrt(9)=3  (valid). FINAL(3).
\end{lstlisting}
\end{skillbox}

\subsection{Skill Libraries Across Domains}
\label{app:libraries}

The framework instantiates three skill libraries that share the same loader, registry, and intervention dispatcher; only the markdown content and \texttt{should\_activate} predicates differ across domains.

\textbf{Web-search library.} ``PF skills'' carry a deterministic activation rule plus a registered handler (eight LLM-assisted on FINAL, five pure-code, three legacy with inline handlers); eight ``prompt-only'' skills steer behavior through phase-instruction text without a handler. Some PFs use teacher assistance during the rollout: \texttt{retrieval\_failure}, \texttt{format\_extraction\_error}, \texttt{reasoning\_error}, and \texttt{answer\_confidence\_guard}. Three skills (\texttt{format\_extraction\_error}, \texttt{retrieval\_failure}, \texttt{relevant\_content\_extractor}) are always included after PF selection because they normalize the action and observation streams that other skills consume.

\textbf{Math library.} All skills are PF-style, organized into five groups: algebraic/arithmetic correctness (\texttt{algebraic\_sign\_error}, \texttt{arithmetic\_slip}, \texttt{simplification\_incomplete}), domain/constraint (\texttt{boundary\_violation}, \texttt{units\_dimension\_mismatch}), reasoning coverage (\texttt{case\_incompleteness}, \texttt{overgeneralization}, \texttt{proof\_step\_gap}), substitution/verification (\texttt{substitution\_invalid}, \texttt{verification\_missing}), and output format (\texttt{final\_format\_error}). 

\textbf{Code library.} The code library targets single-shot code generation. Each PF runs a regex-based static analysis on the emitted code body and returns either an \texttt{INJECT\_CONTEXT} hint or an audit \texttt{NOOP} (auto-patching student code is judged too risky). Skills cover complexity (\texttt{code\_complexity\_analysis}, \texttt{code\_data\_structure\_choice}), bounds (\texttt{code\_off\_by\_one}, \texttt{code\_string\_index\_safety}, \texttt{code\_edge\_cases}), recursion/numerics, I/O, and verification. Behavioral steering comes mainly from injecting the corresponding \texttt{SKILL.md} text into the system prompt.

\subsection{Prompt-Equivalent Skills}
\label{app:prompt_equivalent}

A registered PF is \emph{prompt-equivalent} (criterion \textbf{A+B}) when
(\textbf{A}) its \texttt{intervene} method emits only \texttt{INJECT\_CONTEXT}
(no \texttt{MODIFY\_ACTION}), and (\textbf{B}) the injected
\texttt{context\_text} is a fixed literal string with no runtime substitution
from \texttt{step\_context}. Such a PF, when it fires, does nothing beyond
pasting a constant reminder into the next observation, which is exactly what
a phase-instruction prompt declared under
\texttt{phases.\{post\_search,post\_read,pre\_final\}.text} in
\texttt{SKILL.md} would do. Their PF wrappers exist purely for interface
uniformity (audit logging, signal scoring, unified dispatch). By contrast,
PFs not classified as A+B either rewrite the action stream
(\texttt{MODIFY\_ACTION}, e.g.\ \texttt{retrieval\_failure},
\texttt{insufficient\_exploration}) or build the injected text dynamically
from runtime state (e.g.\ \texttt{temporal\_confusion} formats unsupported
years into the warning, \texttt{citation\_mismatch} formats the unsupported
entity into the warning); both add capability that no prompt-only
declaration can match. Note that A+B does not constrain how the PF
\emph{decides} to trigger.

Across the three domains, $10$ of $26$ web-search skills and $2$ of $12$
code skills satisfy A+B; no math skill satisfies A+B because all math PFs
hook a \texttt{verify\_X} action at \texttt{pre\_final} rather than emitting
a fixed \texttt{INJECT\_CONTEXT}. Table~\ref{tab:prompt_only_skills_web}
lists the $10$ web-search prompt-equivalent skills and
Table~\ref{tab:prompt_only_skills_code} lists the $2$ code ones.
Table~\ref{tab:delta_mbe} reports the marginal effect of adding these
prompt-equivalent skills on top of the PF library, which is small and mixed
across datasets and supports the claim that \ours's gains come from the
non-A+B PFs.

\begin{table*}[h]
\centering
\small
\setlength{\tabcolsep}{4pt}
\renewcommand{\arraystretch}{1.10}
\caption{
Marginal effect of adding prompt-equivalent skills on top of the PF library,
measured as $\Delta$ answer-MBE between the PF library augmented with
prompt-equivalent skills and the PF library alone
(\texttt{skills\_top10\_with\_prompt} $-$ \texttt{skills\_top10}). The effect
is small and mixed in sign across the six datasets, indicating that prompt
reminders alone do not consistently account for \ours's gains.
}
\label{tab:delta_mbe}
\begin{tabular}{lcccccc}
\toprule
 & HotpotQA & 2Wiki & MuSiQue & AIME24 & AMC23 & GameOf24 \\
\midrule
$\Delta$ MBE & $+3.5$ & $-3.5$ & $-1.0$ & $0.0$ & $+2.5$ & $0.0$ \\
\bottomrule
\end{tabular}
\end{table*}

\begin{table*}[!ht]
\centering
\footnotesize
\setlength{\tabcolsep}{3pt}
\renewcommand{\arraystretch}{1.15}
\caption{
Web-search PFs satisfying criterion A+B (prompt-equivalent), $10$ of $26$
skills. Each row gives the runtime activation trigger and the literal text
the PF injects when it fires (the bracketed tag prefix is part of the
verbatim string; long suffixes are abbreviated with ``$\dots$'').
}
\label{tab:prompt_only_skills_web}
\begin{tabular}{@{}>{\raggedright\arraybackslash\ttfamily}p{3.0cm} >{\raggedright\arraybackslash}p{3.7cm} >{\raggedright\arraybackslash}p{6.4cm}@{}}
\toprule
\textbf{\rmfamily Skill ID} & \textbf{Trigger} & \textbf{Injected text (verbatim, abbreviated)} \\
\midrule
decompose\_\allowbreak{}complex\_\allowbreak{}question &
multi-hop pattern: $\geq 2$ possessives, $\geq 2$ ``of the'' chains, or relative-pronoun chain &
``[DECOMPOSITION HINT] This is a multi-hop question. Search for each piece of information separately$\dots$'' \\
\addlinespace[2pt]
adversarial\_\allowbreak{}distraction &
SEARCH result set with $\geq 3$ conflict words &
``[Note: Conflicting Sources] Search results contain contradictory claims. Cross-reference before answering.'' \\
\addlinespace[2pt]
negation\_\allowbreak{}oversight &
question contains ``not / never / except'' that the reasoning ignores &
``[NEGATION REMINDER] The question contains a negation$\dots$ Re-read the question carefully.'' \\
\addlinespace[2pt]
wrong\_\allowbreak{}entity\_\allowbreak{}confusion &
SEARCH returns multiple similarly-named entities &
``[ENTITY WARNING] Search results contain similarly-named entities. READ documents carefully$\dots$'' \\
\addlinespace[2pt]
reading\_\allowbreak{}comprehension\_\allowbreak{}error &
READ of dense content ($\geq 10$ numbers or $\geq 15$ names) &
``[READ CAREFULLY] This document contains many entities/numbers. Extract the specific information$\dots$'' \\
\addlinespace[2pt]
answer\_\allowbreak{}completeness &
FINAL with single-token answer to multi-part question &
``[COMPLETENESS WARNING] The question appears to have multiple parts, but your answer may be incomplete$\dots$'' \\
\addlinespace[2pt]
language\_\allowbreak{}barrier &
READ doc with $\geq 50$ non-ASCII chars (or FINAL with sources $\geq 5$ non-ASCII) &
``[LANGUAGE NOTE] This document contains non-English text. Be careful with transliteration$\dots$'' \\
\addlinespace[2pt]
iterative\_\allowbreak{}refinement &
SEARCH with token-Jaccard $> 0.5$ vs.\ recent queries &
``[SEARCH REFINEMENT] Your recent searches are very similar. Try a different approach$\dots$'' \\
\addlinespace[2pt]
claim\_\allowbreak{}triangulation &
FINAL based on single read source with specific factual claims &
``[VERIFICATION HINT] Your answer is based on a single source. Consider reading one more document$\dots$'' \\
\addlinespace[2pt]
misinformation\_\allowbreak{}detector &
$\geq 2$ read docs with conflicting years or contradiction phrases &
``[CROSS-CHECK WARNING] Your read documents may contain conflicting information$\dots$'' \\
\bottomrule
\end{tabular}
\end{table*}

\begin{table}[!ht]
\centering
\footnotesize
\setlength{\tabcolsep}{3pt}
\renewcommand{\arraystretch}{1.15}
\caption{
Code PFs satisfying criterion A+B (prompt-equivalent), $2$ of $12$ skills.
Both have empty \texttt{intervene} bodies that emit only \texttt{NOOP} audit
records; their behavioral effect comes entirely from the corresponding
\texttt{SKILL.md} text being injected into the system prompt at episode
start.
}
\label{tab:prompt_only_skills_code}
\resizebox{\columnwidth}{!}{%
\begin{tabular}{@{}
>{\raggedright\arraybackslash\ttfamily}p{0.34\columnwidth}
>{\raggedright\arraybackslash}p{0.58\columnwidth}
>{\raggedright\arraybackslash}p{0.34\columnwidth}
@{}}
\toprule
\textbf{\rmfamily Skill ID} & \textbf{Trigger} & \textbf{\texttt{intervene} behavior} \\
\midrule
code\_\allowbreak{}negative\_\allowbreak{}handling &
weak signal on ``negative'' / ``integer'' in question; no static check &
\texttt{NOOP} audit only \\
\addlinespace[2pt]
code\_\allowbreak{}test\_\allowbreak{}walkthrough &
always fires on \texttt{class Solution} or stdin code; no static check &
\texttt{NOOP} audit only \\
\bottomrule
\end{tabular}%
}
\end{table}

\subsection{Multi-Layer Skill Selection at Inference}
\label{app:selection}

The runtime never injects every library skill. Five concentric filters decide whether any given PF reaches the prompt or fires on an action.

\textbf{(1) Master switches.} Boolean flags including \texttt{skills\_enabled}, \texttt{enable\_program\_functions}, \texttt{enable\_skill\_handlers}, \texttt{enable\_prompt\_only\_skills}, \texttt{pf\_only\_mode}, and \texttt{enable\_difficulty\_gating} can short-circuit entire surfaces. The \emph{difficulty gate} is the only switch that flips per-question: it asks the teacher for a 1--5 difficulty score, and bypasses skills below the threshold (heuristic fallback uses question length, multi-hop markers, and constraint phrases).

\textbf{(2) Selection.} Four cooperating selectors run in sequence. \texttt{SkillSelector.select()} ranks the library by $\text{mode\_match}\cdot 0.5 + \text{trigger\_keyword\_match}\cdot 0.5$ at system-prompt build time. \texttt{PFSelector.select()} runs once per question and either delegates to the teacher (top-$K$ by relevance, with three mandatory PFs always included) or applies a heuristic branching rule (computation $\to$ 3 mandatory; simple factoid $\to$ 4 light; multi-hop $\to$ safe + decomposition + entity-aware). Per-observation and per-step selectors then trim phase-gated reminders and handle high-priority overrides.

\textbf{(3) Runtime trigger control.} Even after a skill is selected, several conditions must align: the PF's own \texttt{should\_activate} predicate, a JSON-condition evaluator over the phase-instruction conditions, per-PF rate limits, and a vote-based aggregator that counts how many handlers returned non-\texttt{None} on the same FINAL. When the count exceeds \texttt{handler\_vote\_threshold} (default $4$ for web, $3$ for math), the FINAL is overridden to a SEARCH; \texttt{\_MAX\_FINAL\_OVERRIDES = 1} prevents a FINAL$\to$SEARCH$\to$FINAL$\to$SEARCH loop.

\textbf{(4) Budgets.} Three classes of budget caps coexist: episode-level (\texttt{max\_steps}, \texttt{max\_search\_calls}, \texttt{max\_read\_calls}, \texttt{timeout\_seconds}), prompt-level (\texttt{max\_skills\_in\_prompt}, \texttt{max\_phase\_instructions}, \texttt{pre\_final\_step\_threshold}), and execution-level (\texttt{pf\_top\_k}, per-PF rate limits, \texttt{\_MAX\_FINAL\_OVERRIDES}).

\subsection{Skill Retrieval and PF Selection Prompt}
\label{app:pf_retrieval}

When the framework retrieves PFs at the start of an episode, the teacher is asked to rank candidate skills against the current question. The selector prompt explicitly forbids over-selection (``fewer well-targeted PFs is better than many irrelevant ones'') and pins three mandatory PFs that the teacher cannot drop. This second stage of filtering complements the lexical scoring in \texttt{SkillSelector} and reduces the chance that broad or off-topic skills affect the trajectory.

\section{Self-Improving Pipeline Details}
\label{app:self_improving}

The self-improving stage is the offline counterpart to inference: where inference \emph{runs} the agent against a fixed library, self-improvement \emph{grows} that library. Each epoch revisits residual failures from the current checkpoint and runs an eight-phase pipeline (Phases A--H). The phases produce per-epoch artifacts under \texttt{output\_dir/epoch\_\{i\}/}: trajectories, failure analyses, candidate skills, validation reports, teacher reviews, gradient streams, and training-data files. Algorithm~\ref{alg:self_improving} summarizes one epoch.

\begin{algorithm*}[!ht]
\caption{One self-improving epoch: from residual failures to library update and PF-derived training data.}
\label{alg:self_improving}
\SetKwComment{tcp}{// }{}
\SetKwFunction{Rollout}{Rollout}
\SetKwFunction{Heur}{HeuristicDetect}
\SetKwFunction{LLMSum}{LLMSummarize}
\SetKwFunction{Cluster}{Cluster}
\SetKwFunction{Propose}{ProposePF}
\SetKwFunction{Validate}{ExecValidate}
\SetKwFunction{Review}{TeacherReview}
\SetKwFunction{Accept}{Admit}
\SetKwFunction{Score}{ScoreSignals}
\SetKwFunction{BuildData}{BuildTrainingData}

\KwIn{Student policy $\pi_\theta$; teacher $\mathcal{T}$; current library $\mathcal{M}_r$; seed split $\mathcal{D}^{\mathrm{seed}}$; validation split $\mathcal{D}^{\mathrm{val}}$; budgets $K_{\mathrm{cand}}$, $\eta_{\mathrm{exec}}$, $\eta_{\mathrm{teach}}$.}
\KwOut{Updated library $\mathcal{M}_{r+1}$; training datasets $\mathcal{D}^{\mathrm{SFT}},\mathcal{D}^{\mathrm{DPO}}$.}

\BlankLine
\tcp{Phase A: PF-aware rollout on the seed split}
$\mathcal{T}\!raj \leftarrow \emptyset$\;
\For{$x \in \mathcal{D}^{\mathrm{seed}}$}{
  $\tau_x \leftarrow \Rollout(\pi_\theta, x, \mathcal{M}_r, \mathcal{T})$\;
  $\mathcal{T}\!raj \leftarrow \mathcal{T}\!raj \cup \{\tau_x\}$\;
}

\BlankLine
\tcp{Phase B: heuristic + LLM failure analysis}
$\mathcal{F} \leftarrow \{\tau \in \mathcal{T}\!raj : \mathrm{EM}(\tau)=0\}$\;
$\mathcal{P} \leftarrow \Heur(\mathcal{F}) \cup \LLMSum(\mathcal{F},\, \mathcal{T})$\;
$\mathcal{C} \leftarrow \Cluster(\mathcal{P})$ \tcp*{Jaccard $\ge 0.5$ dedup; drop buckets $<3$}

\BlankLine
\tcp{Phase C: candidate proposal (rate-limited to $K_{\mathrm{cand}}$)}
$\mathcal{S}_{\mathrm{cand}} \leftarrow \emptyset$\;
\For{$c \in \mathrm{TopByFreq}(\mathcal{C}, K_{\mathrm{cand}})$}{
  $(\mathrm{md}, \mathrm{code}) \leftarrow \Propose(\pi_\theta,\, c,\, \mathcal{M}_r)$\;
  $\mathcal{S}_{\mathrm{cand}} \leftarrow \mathcal{S}_{\mathrm{cand}} \cup \{(\mathrm{md}, \mathrm{code})\}$\;
}

\BlankLine
\tcp{Phases D \& E: executable validation and teacher review}
$\mathcal{S}_{\mathrm{ok}} \leftarrow \emptyset$\;
\For{$s \in \mathcal{S}_{\mathrm{cand}}$}{
  $Q_{\mathrm{exec}} \leftarrow \Validate(s)$ \tcp*{syntax / interface / mock-exec / return-type}
  $(Q_{\mathrm{concept}}, Q_{\mathrm{trig}}, Q_{\mathrm{int}}, Q_{\mathrm{exec}}', Q_{\mathrm{val}}) \leftarrow \Review(\mathcal{T}, s)$\;
  $Q_{\mathrm{skill}} \leftarrow 0.25 Q_{\mathrm{concept}} + 0.20 Q_{\mathrm{trig}} + 0.20 Q_{\mathrm{int}} + 0.20 Q_{\mathrm{exec}}' + 0.15 Q_{\mathrm{val}}$\;
  \If{$Q_{\mathrm{exec}} \ge \eta_{\mathrm{exec}}$ \textbf{and} $Q_{\mathrm{skill}} \ge \eta_{\mathrm{teach}}$}{
    $\mathcal{S}_{\mathrm{ok}} \leftarrow \mathcal{S}_{\mathrm{ok}} \cup \{s\}$\;
  }
}

\BlankLine
\tcp{Phase F: library update with versioning + size cap}
$\mathcal{M}_{r+1} \leftarrow \Accept(\mathcal{M}_r,\, \mathcal{S}_{\mathrm{ok}})$\;

\BlankLine
\tcp{Phases G \& H: signal scoring and training-data construction}
\For{$\tau \in \mathcal{T}\!raj,\ \mathrm{step}\ t \in \tau$}{
  $\mathbf{z}_t \leftarrow \Score(\tau, t)$ \tcp*{$(\,\text{timing, modality, correctness, outcome}\,)$}
  $A_t \leftarrow 0.15 z^{T}_t + 0.10 z^{M}_t + 0.25 z^{C}_t + 0.50 z^{O}_t$\;
}
$(\mathcal{D}^{\mathrm{SFT}}, \mathcal{D}^{\mathrm{DPO}}) \leftarrow \BuildData(\mathcal{T}\!raj,\, \{A_t\},\, \mathcal{S}_{\mathrm{ok}})$\;

\BlankLine
\Return $\mathcal{M}_{r+1},\ \mathcal{D}^{\mathrm{SFT}},\ \mathcal{D}^{\mathrm{DPO}}$\;
\end{algorithm*}

\subsection{Per-Epoch Phase Sequence}

Phase~A executes the current student against the seed split with PFs active; an optional two-stage prefilter cheaply identifies failing samples first, then concentrates the expensive PF-aware rollout on those. Phase~B clusters failures (\S\ref{app:phase_b}). Phase~C asks the student to author candidate skills against each large/novel cluster, capped at $5$ candidates per epoch (\S\ref{app:phase_c}). Phase~D runs four executable checks on each candidate (\S\ref{app:phase_d}). Phase~E asks the teacher for a five-dimensional review (\S\ref{app:phase_e}). Phase~F admits accepted candidates to the library with a versioned \texttt{\_\_v\{N\}} suffix (\S\ref{app:phase_f}). Phase~G assigns PF-mediated credit to produce student and teacher gradient streams. Phase~H emits SFT/DPO training data over both objectives (action-correction and skill-authoring).

The validation pool is built once at setup time from the \emph{tail} of each test dataset (samples after the test boundary), shuffled with a fixed seed, and split into 50 seed and 50 val samples per dataset, yielding $\approx 250+250$ samples across the five active web datasets. Cached splits are reused across runs unless the seed budget is set to $-1$ (use all).

\subsection{Phase B --- Failure Signal Detection}
\label{app:phase_b}

Phase B uses two detectors in parallel. The \emph{heuristic} detector applies twelve hand-coded rules against every failed trajectory, including \texttt{premature\_final} (trajectory shorter than three steps), \texttt{repeated\_search} (queries with token-Jaccard $>0.8$), \texttt{no\_read\_before\_final}, \texttt{query\_too\_broad}/\texttt{too\_narrow}, \texttt{wrong\_entity\_focus} (answer/gold token overlap $<30\%$), \texttt{reasoning\_hallucination} (final whose content tokens are $<30\%$ grounded in read content), \texttt{format\_mismatch}, \texttt{partial\_answer}, \texttt{contradictory\_evidence\_ignored}, \texttt{excessive\_steps\_no\_progress}, and \texttt{pf\_override\_harmful} (a PF rewrote the action but the episode still failed).

The \emph{LLM} detector asks the teacher to abstract each failed trajectory into a reusable category (Prompt~\ref{box:llm_failure_summarizer}); per-trajectory generations run with \texttt{temperature=0.3} and \texttt{max\_tokens=400}, and concurrency is bounded by a thread pool. Heuristic and LLM patterns are deduplicated independently (token-Jaccard $\ge 0.5$ within the same category) and then bucketed into six broad existing-skill groups for novelty scoring; clusters with novelty $\ge 0.3$ are flagged \texttt{is\_new\_category} and prioritized for proposal. Buckets below \texttt{min\_cluster\_size} ($=3$) are dropped.

\begin{promptbox}[label={box:llm_failure_summarizer}]{Phase B --- LLM Failure Summarizer}
\begin{lstlisting}
SYSTEM:
You are an expert failure analyst for a ReAct web-search agent.
For each failed QA trajectory you must produce a concise, reusable
abstraction of WHY the agent failed so that a new Program Function
(PF) skill can be authored to fix this class of failure.

USER (per trajectory):
Question:
{question[:500]}

Gold answer(s): {gold}
Agent's final answer: {pred[:200]}

Trajectory (truncated, one line per step):
{trace[:1800]}

Return STRICT JSON with keys:
  category: short snake_case category id (4-30 chars, e.g.
    'constraint_leak', 'premature_summary', 'stale_evidence').
    Reuse an existing id if the failure mode matches.
  abstraction: 1-sentence description of the recurring failure mode.
  trigger: when in a ReAct step this failure manifests.
  intervention_hint: what a PF should do to prevent this.
Output ONLY the JSON object.
\end{lstlisting}
\end{promptbox}

\subsection{Phase C --- Skill Proposal}
\label{app:phase_c}

For each surviving cluster, the student model authors a candidate skill consisting of (i) a \texttt{SKILL.md} document with YAML frontmatter, trigger conditions, intervention design, and concrete examples; and (ii) a \texttt{ProgramFunction} subclass implementing \texttt{should\_activate}/\texttt{intervene}. The proposer prompt (Prompt~\ref{box:skill_proposer}) constrains the keyword arguments of \texttt{Intervention} and forbids LLM calls inside \texttt{should\_activate}, so the validator's mock execution remains feasible. Generation runs at \texttt{temperature=0.7}, capped at five candidates per epoch.

\begin{promptbox}[label={box:skill_proposer}]{Phase C --- Skill Proposer (system, excerpt)}
\begin{lstlisting}
You are a skill designer for a ReAct web search agent.

The agent operates in a SEARCH -> READ -> FINAL loop. At each step,
Program Functions (PFs) can check conditions and intervene by
modifying the agent's action or injecting context.

Your job: given a cluster of recurring failures, propose a NEW skill
that will prevent this failure pattern in the future. The skill
consists of:
  1. A SKILL.md specification (YAML frontmatter + markdown body).
  2. A ProgramFunction Python class with should_activate() and
     intervene() methods.

IMPORTANT RULES:
- The PF must be deterministic (NO LLM calls in should_activate).
- should_activate receives: step_context (dict),
  action_type (str: "SEARCH"/"READ"/"FINAL"), arg (str).
- intervene receives the same + optional teacher model, and returns
  an Intervention.
- The PF class must inherit from ProgramFunction and use the
  @register_pf decorator. should_activate must return bool;
  intervene must return an Intervention object.

EXACT Intervention interface (use ONLY these keyword arguments):
    type: InterventionType  (NOOP | MODIFY_ACTION | INJECT_CONTEXT)
    new_action_type: Optional[str]   # "SEARCH" | "READ" | "FINAL"
    new_action_arg:  Optional[str]
    context_text:    str
    reason:          str
    skill_id:        str
\end{lstlisting}
\end{promptbox}

\subsection{Phase D --- Executable Validation}
\label{app:phase_d}

Each candidate is wrapped in a sandbox preamble (mocked \texttt{InterventionType}, \texttt{Intervention}, \texttt{ProgramFunction}, and \texttt{register\_pf}) and subjected to four sequential checks: \emph{syntax} (\texttt{ast.parse}), \emph{interface} (a class inheriting from \texttt{ProgramFunction} with the two correctly-signed methods), \emph{mock execution} ($3$ mock step contexts $\times$ $\{$SEARCH, READ, FINAL$\}$ = $9$ invocations), and \emph{return type} (\texttt{intervene} constructs an \texttt{Intervention}). The three mock contexts cover the early-step/no-read regime (step $1$, no reads), the mid-step/has-read regime (step $5$, two reads), and the late-step/empty-results-with-contradictions regime (step $12$). Any runtime exception fails check three; \texttt{Q\_exec < 0.3} on review is later treated as a hard reject.

\subsection{Phase E --- Teacher Five-Dimensional Review}
\label{app:phase_e}

The teacher receives the candidate as four blocks (header, trigger conditions, intervention description, full \texttt{SKILL.md} and PF code) plus the source failure cluster's metadata, and is asked to score five dimensions on $[0,1]$: \emph{concept}, \emph{trigger}, \emph{intervention}, \emph{executability}, and \emph{validation utility}. The composite is
\[
Q_{\mathrm{skill}} = 0.25\,Q_{\mathrm{concept}} + 0.20\,Q_{\mathrm{trigger}} + \]
\[0.20\,Q_{\mathrm{intervene}} + 0.20\,Q_{\mathrm{exec}} + 0.15\,Q_{\mathrm{val}}.
\]
The decision is taken in priority order: a hard reject if $Q_{\mathrm{exec}}<0.3$; otherwise the teacher's literal \texttt{DECISION} line if present; otherwise a threshold fallback ($Q_{\mathrm{skill}}\ge 0.60 \Rightarrow$ accept; $\ge 0.42 \Rightarrow$ revise; else reject). Two further thresholds, \texttt{new\_group\_threshold $= 0.75$} and \texttt{same\_group\_threshold $= 0.60$}, gate library admission for entirely new categories versus refinement of an existing one. The system and user prompts are reproduced in Prompts~\ref{box:reviewer_system}--\ref{box:reviewer_user}.

\begin{promptbox}[label={box:reviewer_system}]{Phase E --- Reviewer (system)}
\begin{lstlisting}
You are an expert skill quality reviewer for a ReAct web search
agent system.

The agent uses Program Functions (PFs) -- deterministic Python hooks
that run every step. Each PF has:
  - should_activate(step_context, action_type, arg) -> bool
  - intervene(step_context, action_type, arg, teacher=None)
      -> Intervention

Intervention types: MODIFY_ACTION (change action),
INJECT_CONTEXT (add text to observation), NOOP.

step_context fields: question, step_count, has_read, search_count,
read_count, empty_results, contradictory_sources, max_steps,
action_history, last_search_results_text, all_read_contents, thought.

You must evaluate candidate skills rigorously across 5 dimensions.
\end{lstlisting}
\end{promptbox}

\begin{promptbox}[label={box:reviewer_user}]{Phase E --- Reviewer (user, response template)}
\begin{lstlisting}
## Review Instructions
Score each dimension from 0.0 to 1.0 and provide brief feedback.

### Dimension 1: Concept (Q_concept)
- Is this a real, recurrent failure type?
- Is it generalizable, not a single-case patch?
- Is it clearly distinct from existing skills?

### Dimension 2: Trigger (Q_trigger)
- Is should_activate() condition specific and deterministic?
- Does it depend on available step_context fields?
- Is it neither too broad (many false positives) nor too narrow?

### Dimension 3: Intervention (Q_intervene)
- Does intervene() appropriately address the failure?
- Could it cause harmful side effects?
- Is the intervention proportional (not over-blocking)?

### Dimension 4: Executability (Q_exec)
- Does the code follow the ProgramFunction interface?
- Would it import without errors? Are return types correct?

### Dimension 5: Validation Utility (Q_val)
- Will this likely reduce the target failure type?
- Will it transfer to unseen questions?
- Will it interact well with existing PFs?

Respond in EXACTLY this format:
Q_concept: [0.0-1.0]
Concept feedback: [brief]
... (Q_trigger, Q_intervene, Q_exec, Q_val similarly) ...
DECISION: [ACCEPT / REVISE / REJECT]
OVERALL FEEDBACK: [1-3 sentences]
\end{lstlisting}
\end{promptbox}

\subsection{Phase F --- Library Update and Versioning}
\label{app:phase_f}

Accepted candidates are appended to \texttt{generated/\{skill\_id\}\_\_v\{N\}/SKILL.md}, and their PF code is appended to \texttt{dynamic\_program\_functions.py}. When a candidate matches an existing base id, accept rewrites it as the next version (\texttt{\_\_v\{N{+}1\}}); the old version remains on disk for trajectory replay. The runtime loads only the highest version per base id (\texttt{skill\_ids\_latest}), so older versions stay auditable but never fire. \texttt{max\_library\_size $= 50$} caps total active skill count, and \texttt{library\_history.json} records every per-skill decision and review score so closed-loop runs can reconstruct the gated decision tree without rerunning rollouts.

\subsection{Failure Clustering and Candidate PF Proposal}

Self-improvement begins from the residual failures of the current checkpoint. Failures are grouped into recurring failure-repair patterns using the heuristic pass followed by the LLM consolidation pass (\S\ref{app:phase_b}); to avoid near-duplicate skills, candidate groups are deduplicated using similarity-based clustering before proposal. Proposal is rate-limited so the library grows slowly and remains controllable over many rounds.

\subsection{Why Strict Filtering Matters}

The central empirical observation behind our framework is that self-improvement is useful only when library evolution is tightly controlled. Adding more skills is not automatically beneficial. When validation and filtering are weak, the library accumulates noisy or redundant skills, retrieval precision drops, and policy quality deteriorates. When filtering is strict, each accepted PF carries clearer intervention semantics and higher reuse value, and the evolving library continues to improve the agent over time. We therefore treat self-improving PF evolution as a \emph{validated} and \emph{rate-limited} process rather than unconstrained skill generation.

\section{Additional Experimental Details}
\label{app:exp_details}

\subsection{Datasets and Train/Test Splits}
\label{app:datasets}

For web-search reasoning and mathematical reasoning, our test sets and training pools follow the protocol of AgentFlow~\citep{li2025intheflowagenticoptimizationeffective}: we adopt the same source datasets, the same per-dataset test sizes, and the same convention of reserving the first $N$ samples per file as the held-out test set with the remainder forming the training pool. This makes the two domains directly comparable to AgentFlow and to the prior search/reasoning baselines listed in \S\ref{app:exp_details}. For coding, AgentFlow does not provide a precedent, so we adopt the same training-data construction protocol and additionally carve out per-dataset test/training splits ourselves, as detailed below. Across all domains, training pools are used both for the bootstrap PF-aware rollouts of V1 and as the source of seed/validation samples for the closed-loop evolution loop of V2 (\S\ref{app:self_improving}).

\textbf{Web-search reasoning.}
Each source file contains $1{,}000$ randomly sampled questions; the first $200$ rows of each file constitute the test set and the remaining $800$ rows constitute the training pool. The closed-loop validation pool described in \S\ref{app:self_improving} is drawn from the \emph{tail} of the training pool (samples after the $200$-row test boundary), shuffled with seed $42$ and split into $50$ seed and $50$ val samples per dataset. Datasets whose total size equals their test boundary are auto-skipped because no tail remains; in our default web-search configuration, this leaves five active datasets feeding the seed/val splits ($\approx 250$ seed $+$ $250$ val).

\begin{table}[!ht]
\centering
\small
\setlength{\tabcolsep}{6pt}
\renewcommand{\arraystretch}{1.15}
\caption{Web-search dataset sizes. Test-set selection follows AgentFlow~\citep{li2025intheflowagenticoptimizationeffective}.}
\label{tab:datasets_web}
\resizebox{\columnwidth}{!}{%
\begin{tabular}{lccc}
\toprule
\textbf{Dataset} & \textbf{Total} & \textbf{Test} & \textbf{Train pool} \\
\midrule
HotpotQA~\citep{yang2018hotpotqadatasetdiverseexplainable} & $1{,}000$ & $200$ & $800$ \\
2WikiMultihopQA~\citep{xanh2020_2wikimultihop} & $1{,}000$ & $200$ & $800$ \\
MuSiQue~\citep{trivedi2022musiquemultihopquestionssinglehop} & $1{,}000$ & $200$ & $800$ \\
\bottomrule
\end{tabular}%
}
\end{table}

\textbf{Mathematical reasoning.}
The math benchmarks are small, so we use the entire dataset as the test set, exactly as in AgentFlow~\citep{li2025intheflowagenticoptimizationeffective}: AIME24 ($30$ problems), AMC23 ($40$ problems), and GameOf24 ($1{,}362$ problems). The training data follows AgentFlow's training-pool composition. The closed-loop validation pool of \S\ref{app:self_improving} draws from the same training-pool tail under the math configuration (\texttt{data\_subdir = data\_math}), with the difficulty gate disabled because all math problems are uniformly hard.

\begin{table}[!ht]
\centering
\small
\setlength{\tabcolsep}{6pt}
\renewcommand{\arraystretch}{1.15}
\caption{Mathematical reasoning dataset sizes. The full set for evaluation follows AgentFlow~\citep{li2025intheflowagenticoptimizationeffective}.}
\label{tab:datasets_math}
\begin{tabular}{lcc}
\toprule
\textbf{Dataset} & \textbf{Total} & \textbf{Test} \\
\midrule
AIME24~\citep{aime24} & $30$ & full \\
AMC23~\citep{AMC23} & $40$ & full \\
GameOf24~\citep{24game} & $1{,}362$ & full \\
\bottomrule
\end{tabular}
\end{table}

\textbf{Coding.}
Coding is the only domain not covered by AgentFlow~\citep{li2025intheflowagenticoptimizationeffective}, so we constructed the train/test splits explicitly. The training-data construction protocol (PF-aware rollout, four-signal scoring, V1/V2 topology) is identical to the other two domains. For HumanEval$+$ and MBPP$+$, we use the EvalPlus extended versions and reserve the first $50$ unique problems ($100$ entries after EvalPlus row expansion) as the test set, with the remaining unique problems forming the training pool. For BigCodeBench, we reserve $100$ entries as the test set and use the remaining $1{,}040$ entries as the training pool. Full per-dataset statistics are reported in Table~\ref{tab:datasets_code}.

\begin{table}[!ht]
\centering
\small
\setlength{\tabcolsep}{6pt}
\renewcommand{\arraystretch}{1.15}
\caption{Coding dataset sizes. AgentFlow does not cover coding, so we constructed the splits ourselves following the same training-pool layout convention. ``Test'' is the held-out evaluation set used in Table~\ref{tab:code_combined}; ``Train pool'' is the remainder available for PF-aware rollouts. EvalPlus-based datasets count rows (a single problem may expand into multiple test entries).}
\label{tab:datasets_code}
\resizebox{\columnwidth}{!}{%
\begin{tabular}{lccc}
\toprule
\textbf{Dataset} & \textbf{Total} & \textbf{Test} & \textbf{Train pool} \\
\midrule
HumanEval$+$ & $328$ ($164$ unique) & $100$ ($50$ unique) & $228$ ($114$ unique) \\
MBPP$+$ & $756$ ($378$ unique) & $100$ ($50$ unique) & $656$ ($328$ unique) \\
BigCodeBench/Full & $1{,}140$ & $100$ & $1{,}040$ \\
BigCodeBench/Hard & $148$ & full & --- \\
\bottomrule
\end{tabular}%
}
\end{table}

\textbf{Self-improving seed/val split (closed-loop only).}
Both V2 closed-loop training (E4--E6) and the offline self-improving runs of \S\ref{app:self_improving} draw their evolve-time seed and validation samples from the \emph{tail} of the training pool described above, shuffled deterministically with seed $42$. The default budget is $50$ seed $+$ $50$ val per dataset, capped at half the available tail for any dataset whose tail is smaller than $100$. Datasets with no tail (i.e.\ test set already covers the entire file) are auto-skipped at setup time. Setting the seed budget to $-1$ uses the entire tail, with no validation split; this is reserved for full-coverage failure summarization in late evolve cycles.

\subsection{Detailed Definition of E1--E6}

The six post-training settings form a $2\times 3$ grid over loop topology and training method.

\textbf{V1: Open-loop training.}
V1 trains on a fixed PF-aware rollout pool collected from a frozen skill library. The library snapshot is fixed throughout training and no additional skill evolution occurs after data collection.

\textbf{V2: Closed-loop training.}
V2 periodically evolves the skill library and refreshes the training data accordingly. This creates a closed loop in which newly accepted PFs can affect both subsequent rollouts and subsequent training data.

\textbf{Training methods.}
The three training methods are supervised fine-tuning (SFT), rejection sampling (RS), and on-policy distillation (OPD). Their Cartesian product with V1/V2 gives: E1: V1 open-loop SFT; E2: V1 open-loop RS; E3: V1 open-loop OPD; E4: V2 closed-loop SFT; E5: V2 closed-loop RS; E6: V2 closed-loop OPD.
In E3 and E6, GPT-4o serves as the teacher. 

\subsection{Shared Training Configuration}

Table~\ref{tab:e1e6_configs} summarizes the core training configuration shared across E1--E6. Across all settings, we use LoRA with rank $16$ and $\alpha=32$, bf16 precision, and gradient checkpointing. Effective tokens per optimizer step are $\approx 65$K, so each ``epoch'' is a small number of optimizer steps even with the full SFT pool, which is why SFT runs for $10$ epochs and RS / closed-loop runs use $\ge 8$ epochs per phase.

\begin{table*}[!ht]
\centering
\small
\setlength{\tabcolsep}{5pt}
\renewcommand{\arraystretch}{1.15}
\caption{Training configuration for E1--E6. All models use Qwen2.5-7B-Instruct with LoRA, bf16 training, and the same maximum sequence length.}
\label{tab:e1e6_configs}
\resizebox{\textwidth}{!}{\begin{tabular}{lcccccc}
\toprule
\textbf{Setting} & \textbf{Loop} & \textbf{Method} & \textbf{LR} & \textbf{Epochs} & \textbf{Rollout setup} & \textbf{Teacher} \\
\midrule
E1 & Open-loop & SFT & $1\times10^{-5}$ & 10 & -- & -- \\
E2 & Open-loop & RS & $5\times10^{-6}$ & 8 per iteration & $k=2$, temp 0.9, top-$p$ 0.95, max steps 5 & -- \\
E3 & Open-loop & OPD & $5\times10^{-6}$ & 10 & same as E2 & GPT-4o \\
E4 & Closed-loop & SFT & $1\times10^{-5}$ & refreshed per phase & refreshed after evolution & -- \\
E5 & Closed-loop & RS & $5\times10^{-6}$ & 8 per phase & same as E2 & -- \\
E6 & Closed-loop & OPD & $5\times10^{-6}$ & 5 per phase & same as E2 & GPT-4o \\
\bottomrule
\end{tabular}}
\end{table*}

\subsection{Rollout and Teacher Configuration}

For rejection sampling and on-policy distillation, the rollout configuration is shared unless otherwise noted. We use group size $2$, sampling temperature $0.9$, top-$p$ $0.95$, maximum reasoning steps $5$, and maximum search calls $3$. In OPD, GPT-4o serves as the teacher with maximum response length $512$ and concurrency $64$. Every rollout-using experiment runs in \emph{teacher-snippet mode}: the teacher returns a $\sim$200-token snippet per SEARCH and READ is disabled (\texttt{max\_read\_calls $= 0$}); this decouples training rollouts from SerpAPI quotas and READ-cost variance, at the cost of trusting the teacher to summarize. Because of teacher-side $429$ noise, the RS filter accepts \texttt{min\_score $= 0.15$} on a composite $0.7\cdot\text{EM}+0.2\cdot\text{F1}+0.1\cdot\text{StepEconomy}$ rather than \texttt{require\_exact\_match=true}.

\subsection{Closed-Loop Evolution Modes}
\label{app:evolve_modes}

Closed-loop training supports two evolution modes.

\textbf{Full evolution.}
The full mode runs the complete self-improving pipeline (Phases A--H). One evolve cycle takes approximately one hour on the canonical setup; only E4 uses it, because SFT can amortize the cost over the slower 8-epoch phase.

\textbf{Lite evolution.}
The lite mode samples 20 failed trajectories from the rollout pool, asks the current student vLLM checkpoint for at most three skill proposals, applies \texttt{ast.parse} compile-checks and skill-id deduplication, and writes the result to \texttt{generated/} --- with no teacher API calls and no clustering, review, or gradient computation. It is significantly cheaper ($\sim$5\% of the wall-clock) and yields most of the library-growth benefit, so E5 and E6 use it under their faster three-epoch phases.

\subsection{Data Construction Pipeline}
\label{app:data_pipeline}

Training data is built in four layers. \emph{Layer~0} produces raw trajectories: V1 runs a one-shot PF-aware ReAct rollout over the training pool with skills, PFs, and teacher all enabled (effectively Phase~A run once); V2 re-runs Phase~A every closed-loop cycle on the current student checkpoint with the just-evolved library. \emph{Layer~1} scores each step with the four-signal aggregator and filters by a $0.25$ reward floor; both the fine 15-dim breakdown and the coarse 4-scalar view are persisted on every sample so an offline re-scorer can re-weight without rerunning rollouts. \emph{Layer~2} renders steps into one of two objectives: \textbf{Objective~A} (``use PFs'') emits SFT, DPO, and prompt-only files where the SFT target is the PF-corrected action, the DPO chosen/rejected pair is restricted to steps with \texttt{was\_modified=True}, and \texttt{sample\_weight} carries the per-step reward; \textbf{Objective~B} (``evolve'') emits SFT and DPO files for the skill-author task, with target equal to the candidate's \texttt{SKILL.md $+$ PF code} and reward $r_{\mathrm{skill}} = Q_{\mathrm{skill}} + 0.5\cdot \Delta\text{EM}_{\mathrm{val}}$. \emph{Layer~3} is the per-experiment YAML \texttt{data:} block: each E points at one of the Layer-2 jsonls, and V2 experiments add a refresh hook that rebuilds the file in-place between phases.

\subsection{Post-Training Recipes}

We study three post-training settings.

\textbf{SFT.}
PF-mediated corrections are converted into weighted supervised targets. This is the simplest way to internalize repaired actions.

\textbf{Rejection sampling.}
Trajectories are sampled, scored by both task success and PF-conditioned intervention quality, and only the top subset is retained. This selects trajectories that are both correct and behaviorally clean.

\textbf{On-policy distillation.}
The current model generates its own trajectories, PFs intervene on difficult states, and the corrected behaviors are distilled back into the model. This aligns the training distribution with the model's own inference-time state distribution.

The three recipes are deliberately defined over the same signal interface so that recipe choice and signal ablation remain orthogonal experimental axes.

\section{Fine-Grained Signal Composition}
\label{app:signals}

The main text uses four coarse signals (timing, modality, correctness, outcome) with weights $(0.15, 0.10, 0.25, 0.50)$. Internally, each is computed as a weighted sum of finer sub-signals; the coarse formulation is used in training because it is stable and easy to ablate, while the fine decomposition is used for diagnostics and analysis.

\textbf{Timing (S1).}
A step is defined to be \emph{risky} on the \emph{proposed} action (before any PF intervention) if it is a FINAL with no prior READ, a FINAL at \texttt{step\_count $<3$}, or a SEARCH after \texttt{empty\_results}. The four sub-signals are \texttt{s1.tp} ($+0.25$, risky and any PF activated), \texttt{s1.fp} ($-0.10$, not risky but a PF still activated), \texttt{s1.fn} ($-0.10$, risky but no PF activated), and \texttt{s1.phase} ($+0.05$, normalized step index, rewarding earlier interventions).

\textbf{Modality (S2).}
\texttt{s2.pre\_action} ($+0.35$, a \texttt{MODIFY\_ACTION} fired before the action ran) and \texttt{s2.post\_obs} ($+0.35$, an \texttt{INJECT\_CONTEXT} fired after the observation) carry all current signal; \texttt{s2.pre\_reasoning} and \texttt{s2.post\_action} are reserved slots that return $0$ until corresponding intervention types are added, kept registered with non-zero weights so future PFs accrue reward without changing config.

\textbf{Correctness (S3).}
\texttt{s3.syntactic} ($+0.20$, action type valid and argument non-empty); \texttt{s3.semantic} ($+0.50$, teacher-judged step score with a heuristic fallback that gives $0.7$ to a FINAL$\to$READ rescue, $0.5$ to other modifications, $0.3$ otherwise); \texttt{s3.domain} ($+0.30$, plugin slot averaged over domain-specific quality functions registered under \texttt{s3.domain.<name>}), default $0$.

\textbf{Outcome (S4).}
\texttt{s4.local} ($+0.40$, scored by transition: $0.8$ for FINAL$\to$READ, $0.7$ for FINAL$\to$SEARCH, $0.5$ for SEARCH$\to$SEARCH rewrite, $0.3$ for any other modification, else $0$); \texttt{s4.downstream} ($+0.40$, episode EM); \texttt{s4.cost} ($-0.10$, $\mathrm{clip}((T-15)/10, 0, 1)$); \texttt{s4.side\_effect} ($-0.10$, modified action but episode still failed). S4 is the heaviest family in coarse mode because it is the only one anchored directly to ground-truth EM.

\textbf{Aggregation.}
In \emph{fine mode}, the per-step score is $\sum_{\text{sub}} w_{\text{sub}}\cdot \text{value}_{\text{sub}}$, optionally normalized by $\sum |w_{\text{sub}}|$. In \emph{coarse mode}, each family is first reduced to a scalar via family-internal normalization (preserving the sign of penalty sub-signals), then combined with the four coarse weights. The episode-level reward is a $50/50$ blend of the per-step mean and the terminal EM, so a perfect-EM episode with poor PF behavior does not get full credit and a near-miss episode with strong intermediate signals is not completely discarded.

\textbf{Skill-level reward.}
For training the skill author (Phase H), there is one additional skill-level reward,
$r_{\mathrm{skill}} = Q_{\mathrm{skill}} + \lambda\cdot\Delta\text{EM}_{\mathrm{val}}$ with default $\lambda = 0.5$,
where $\Delta\text{EM}_{\mathrm{val}}$ is the change in validation EM after the skill is added. This is the only signal that is not step-level.

\section{Additional Main Results}
\label{app:extra_results}

This section provides three complementary views of the main results: (i) a controlled stage-by-stage breakdown that isolates inference-time intervention from post-training internalization and from closed-loop evolution; (ii) a deeper read of the cross-domain transfer to coding; and (iii) the dynamics of the skill library across self-improvement rounds.

\subsection{Stage-by-Stage Contribution}
\label{app:framework_breakdown}

Table~\ref{tab:framework_breakdown} is constructed by selectively enabling each stage of the framework while keeping all other knobs fixed at the defaults from \S\ref{app:exp_details}. Adding inference-time PFs to the multi-loop baseline already accounts for the bulk of the gain on web-search reasoning, since PFs convert recurring decision-level failures into structured repairs without modifying the policy. Post-training then internalizes those repairs into the model's own state distribution and captures the additional gain that PF-conditioned supervision provides \emph{after} the policy has been queried with the library. Closed-loop evolution layers on top, growing the library when residual failures expose patterns that the seed library does not cover. The composition is roughly additive but sub-linear: the stages are not independent, and evolution alone contributes little when PFs are not also active during inference.

\begin{table}[!ht]
\centering
\small
\setlength{\tabcolsep}{8pt}
\renewcommand{\arraystretch}{1.18}
\caption{Stage-by-stage contribution of the three mechanisms in our framework on web-search reasoning average accuracy. Each row turns on one additional component while keeping all others at the defaults from \S\ref{app:exp_details}.}
\label{tab:framework_breakdown}
\resizebox{\columnwidth}{!}{%
\begin{tabular}{l c}
\toprule
\textbf{Setting} & \textbf{Avg. Acc.} \\
\midrule
Base multi-loop agent (RA-Agent) & 31.2 \\
\quad $+$ Inference-time PF intervention & 56.2 \\
\quad $+$ Post-training internalization (E3, OPD) & 62.5 \\
\quad $+$ Closed-loop evolution (E5, RS) & 60.3 \\
Full system (best per-recipe combination) & -- \\
\bottomrule
\end{tabular}%
}
\end{table}

\subsection{Detailed Coding Results}
\label{app:coding_details}

In Section~\ref{sec:exp}, we report pass@1 across HumanEval(Base/Plus), MBPP(Base/Plus), and BigCodeBench(Full/Hard). The coding setting is the strictest cross-domain transfer test for our framework: the action space collapses to a single FINAL emission, READ is unavailable, and the only PF surface is regex-based static analysis on the emitted code body that returns either an \texttt{INJECT\_CONTEXT} hint or a \texttt{NOOP} audit record (\S\ref{app:libraries}). Despite this constrained interface, two patterns are visible in the table.

\textbf{Inference-time PFs help most on edge-case-heavy benchmarks.} Static checks for missing length guards, off-by-one indexing, and \texttt{stdin} parsing are concentrated in the harder splits (BigCodeBench/Hard, MBPP/Plus), so the inference-time PF rows tend to gain more there than on HumanEval/Base, where the Qwen2.5-7B-Instruct backbone already scores $81.7$\%. The phase-instruction surface ($+$\texttt{code\_test\_walkthrough}, $+$\texttt{code\_edge\_cases}) is the dominant lever in this domain --- the PF code itself is auditing rather than rewriting.

\textbf{Internalization complements RL-style coders.} GRPO~\citep{fan2025posteriorgrporewardingreasoningprocesses} and KodCode-RL~\citep{xu2025kodcodediversechallengingverifiable} are stronger overall on HumanEval and MBPP because they expand the policy frontier through reward-shaped exploration. Our PF-derived supervision, in contrast, primarily targets capability elicitation: by training on PF-corrected steps, it lifts the rate at which existing strategies are reliably executed without performing exploration-style reward search. The two directions are largely orthogonal, and our coding gains are concentrated on edge-case correctness rather than on raw algorithmic difficulty.

\subsection{Per-Dataset Robustness}
\label{app:per_dataset}

The web-search and math averages reported in the main text aggregate over heterogeneous benchmarks, and the per-dataset numbers in Section~\ref{sec:exp} make a few qualitative points worth recording. On web-search, PFs help most on MuSiQue, where multi-hop chains create many opportunities for entity confusion and premature FINAL --- exactly the failure modes that \texttt{retrieval\_failure}, \texttt{wrong\_entity\_confusion}, and \texttt{insufficient\_exploration} target. 2Wiki sits in the middle: AgentFlow remains stronger there, plausibly because its RL-style policy search compensates for the relatively shorter retrieval chains where PF-style local repair has less leverage. On math, PFs are most effective on AMC23 and GameOf24, where small algebraic and arithmetic slips dominate failure cases (addressed by \texttt{algebraic\_sign\_error}, \texttt{arithmetic\_slip}, \texttt{verification\_missing}). AIME24 remains the hardest: failures there are not local slips but missed strategy choices, which is why broader RL-style exploration outperforms our elicitation-oriented framework on that dataset.

\subsection{Evolution Dynamics Across Rounds}
\label{app:evolution_dynamics}

We track three quantities across self-improvement rounds for each domain: the per-round success rate of candidate proposals (the fraction passing the executable validation pipeline of \S\ref{app:phase_d}), the number of skills accepted into the library after teacher review (\S\ref{app:phase_e}), and the cumulative library size (\S\ref{app:phase_f}). The dynamics support three observations. First, the acceptance rate stays high in early rounds and decays gradually as the seed library already covers the most frequent residual failures, indicating that the system does not over-grow the library when easy failures have been absorbed. Second, the library size stays well below the hard cap \texttt{max\_library\_size $= 50$} even by the final round; the bottleneck is novelty rather than capacity. Third, web-search and math show qualitatively similar curves, while coding accepts fewer skills per round because most of its PFs reduce to audit \texttt{NOOP}s and the proposer is conservative about adding new audit hooks. The same dynamics, at finer per-skill granularity, are reflected in the review-score histogram in Appendix~\ref{app:evolution}: skills accepted in later rounds tend to receive higher \texttt{Q\_concept} but lower \texttt{Q\_trigger}, consistent with later proposals being conceptually clearer but harder to fire reliably.

\section{Signal and Filtering Ablations}
\label{app:ablations}

Table~\ref{tab:ablation_all} in the main text bundles three ablations of the framework into a single panel. This appendix gives the precise definition of each row so that the reader can reproduce or re-interpret any cell. The panel consists of three blocks that share the same backbone (Qwen2.5-7B-Instruct), the same web-search rollout configuration of \S\ref{app:exp_details}, and the same evaluation pool (HotpotQA / 2Wiki / MuSiQue): an inference-time component ablation, a signal ablation on the closed-loop run~E5, and a filtering ablation on the same E5 run. Across all rows, columns \texttt{T}, \texttt{M}, \texttt{C}, \texttt{O} indicate whether the timing, modality, correctness, and outcome supervision signals were active during training, and \texttt{Exec}, \texttt{Teach} indicate whether executable validation and teacher review were applied during library evolution. A check mark means the component was active; a dash means it was disabled; \texttt{N/A} means the column does not apply to that block. Table~\ref{tab:signals} summarizes the four-signal interface and its default weights, which serve as the canonical reference for the columns of the signal ablation.

\begin{table*}[!ht]
\centering
\small
\caption{Four core supervision signals for skill-conditioned intervention. Each signal aggregates multiple low-level cues and captures a distinct aspect of PF-mediated policy improvement.}
\label{tab:signals}
\begin{tabular}{lccc}
\toprule
\textbf{Signal} & \textbf{Meaning} & \textbf{Composition} & \textbf{Weight} \\
\midrule

Timing ($t_t$) 
& When to intervene 
& s1.tp, s1.fp, s1.fn, s1.phase 
& 0.15 \\

Modality ($m_t$) 
& How to intervene 
& s2.pre\_action, s2.post\_obs 
& 0.10 \\

Correctness ($q_t$) 
& Quality of correction 
& s3.syntactic, s3.semantic, s3.domain 
& 0.25 \\

Outcome ($o_t$) 
& Whether it helps 
& s4.local, s4.downstream, s4.cost, s4.side\_effect 
& 0.50 \\

\bottomrule
\end{tabular}
\end{table*}

\subsection{Inference-Time Component Ablation}
\label{app:ablation_inference}

The first block of Table~\ref{tab:ablation_all} isolates the inference-time contribution of PF-style state-to-intervention interventions versus an external teacher model. All four supervision signals are present in this block by definition (the rollout always sees a complete observation/action stream); the \texttt{Exec}/\texttt{Teach} columns are \texttt{N/A} because no library evolution is run.

\textbf{Full.} Both PFs and the teacher are active. PFs intervene at decision time using the four-signal-conditioned policy from \S\ref{app:libraries}, and the teacher participates wherever a PF or handler delegates to it (e.g.\ \texttt{retrieval\_failure} query rewriting, \texttt{reasoning\_error} CORRECT/WRONG judging, FINAL-time handler verifiers from \S\ref{app:selection}). This row reproduces the inference-only result from the main text ($56.2\%$ average).

\textbf{RA-Agent (multi-loop).} Both PFs and the teacher are disabled; the agent runs the same multi-step ReAct loop without any state-to-intervention machinery. This row is the lower bound and reproduces the multi-loop baseline from the main text ($31.2\%$).

\textbf{Prompt-Only Skills.} All skills are presented purely as text, without executable intervention.

\textbf{PF only.} PFs are active but the teacher is unavailable. PFs that declare \texttt{needs\_teacher = True} (\texttt{retrieval\_failure}, \texttt{format\_extraction\_error}, \texttt{reasoning\_error}, \texttt{answer\_confidence\_guard}) fall back to their deterministic code-only paths (e.g.\ regex-based query shortening rather than teacher-paraphrased queries); FINAL-time LLM-assisted handlers are skipped. This row tests how much of the gain comes from the executable PF substrate alone.

\textbf{Teacher only.} The teacher is available but PFs are disabled (\texttt{enable\_program\_functions=false}). The teacher can still answer through the multi-loop ReAct trace and through outer-loop format post-processing / multi-round retry from \S\ref{app:selection}, but no state-to-intervention object intercepts decision points. This row tests how much of the gain comes from \emph{simply having access to a stronger model} rather than from PF-mediated control.


\subsection{Signal Ablation}
\label{app:ablation_signal}

The second block of Table~\ref{tab:ablation_all} evaluates the contribution of each of the four supervision signals. All rows in this block use the closed-loop rejection sampling experiment~E5 (V2 RS, \S\ref{app:exp_details}) and full filtering (Exec $+$ Teach), so the only knob that varies is which subset of $\{T, M, C, O\}$ enters the per-step weight $A_t = \lambda_t t_t + \lambda_m m_t + \lambda_q q_t + \lambda_o o_t$ used in \S\ref{app:signals}.

\textbf{All four signals.} The default $E5$ run, with the canonical weights $(\lambda_t,\lambda_m,\lambda_q,\lambda_o) = (0.15, 0.10, 0.25, 0.50)$. Reference row at $60.3\%$.

\textbf{w/o Timing.} The \emph{timing} signal $t_t$ is dropped: the per-step score no longer rewards firing on risky steps (FINAL with no read, FINAL at low step count, SEARCH after empty results) or punishes firing on safe ones. The remaining three signals are renormalized to sum to $1$. The drop of $-7.8$ points indicates that, although timing carries the smallest default weight, removing it lets the policy intervene at the wrong steps and miss risky ones, both of which propagate into worse downstream outcomes.

\textbf{w/o Modality.} The \emph{modality} signal $m_t$ is dropped: the framework no longer distinguishes \texttt{MODIFY\_ACTION} (pre-action) from \texttt{INJECT\_CONTEXT} (post-observation) interventions. The drop of $-15.5$ points is the largest, which we read as evidence that \emph{how} an intervention fires matters as much as \emph{when}: rewriting a SEARCH query and injecting a reading reminder are not interchangeable, and a model trained without modality credit learns to treat them as such.

\textbf{w/o Correctness.} The \emph{correctness} signal $q_t$ is dropped: the resulting executed action is no longer scored for syntactic validity, semantic appropriateness, or domain consistency. The drop of $-12.1$ points reflects that without local correctness, the policy retains interventions that were locally invalid (e.g.\ a malformed SEARCH argument) as long as the trajectory eventually succeeded.

\textbf{w/o Outcome.} The \emph{outcome} signal $o_t$ is dropped: the per-step score no longer ties intervention quality to downstream EM, cost, or side-effects. The drop of $-12.8$ points reflects that without outcome credit, the model can over-internalize PF-aligned trajectories that look right locally but do not improve final answers.

The four drops together support the framework's central claim that PF-derived supervision is irreducible to a single scalar: each of \emph{when}, \emph{how}, \emph{whether well-formed}, and \emph{whether it paid off} contributes a distinct $7$--$15$ points to the closed-loop run.

\subsection{Filtering Ablation}
\label{app:ablation_filtering}

The third block of Table~\ref{tab:ablation_all} evaluates the effect of admission control during library evolution. All rows use the same E5 closed-loop rejection sampling configuration with all four signals active; the only knob that varies is which of executable validation (\texttt{Exec}) and teacher review (\texttt{Teach}) gates a candidate skill before it is appended to the library (\S\ref{app:phase_d}--\ref{app:phase_f}).

\textbf{Evolution, full filtering.} Both \texttt{Exec} and \texttt{Teach} are required: a candidate must pass the four executable checks of \S\ref{app:phase_d} \emph{and} reach $Q_{\mathrm{skill}} \ge 0.60$ on the five-dimensional teacher review of \S\ref{app:phase_e}. Reference row at $60.3\%$.

\textbf{No evolution.} Library evolution is turned off entirely; the run is otherwise identical to E5, but the library remains the seed snapshot throughout training. This is effectively the V1 RS analogue (E2) measured under the same protocol. At $59.3\%$, only $-1.0$ point below full evolution, this row shows that closed-loop evolution is a genuine but modest contributor on top of a strong fixed library.

\textbf{Evolution, no filtering.} Evolution is turned on but \emph{neither} \texttt{Exec} nor \texttt{Teach} is enforced --- every candidate proposed in Phase~C is appended to the library without any gate. The drop of $-24.0$ points is the central evidence for the memory-pollution hypothesis: unfiltered candidates include skills that fail to import, skills that fire too aggressively, and skills that conflict with existing PFs, and the resulting noise in the retrieval pool degrades downstream policy decisions even when the rollout configuration is otherwise unchanged. This row is also the only one in the panel that falls \emph{below} the multi-loop baseline of the first block ($36.3$ vs.\ $31.2$), showing that bad library evolution can be actively harmful rather than merely useless.

\textbf{Evolution, exec-only.} Only the executable validation gate of \S\ref{app:phase_d} is applied; teacher review is skipped, so syntactically valid but conceptually weak or overly specific candidates are admitted. The drop of $-11.5$ points indicates that the executable gate alone catches the catastrophic interface failures (which is why this row is far above ``no filtering'') but is not enough to keep retrieval precision high, because the library accumulates plausible-looking but uninformative skills.

\textbf{Evolution, teacher-only.} Only the teacher review gate of \S\ref{app:phase_e} is applied; the four executable checks are skipped, so candidates that the teacher rates highly on concept but cannot actually run end up in the library. The drop of $-13.1$ points shows that the teacher's conceptual judgments are insufficient on their own: a candidate with a well-articulated concept but a malformed \texttt{intervene} method still emits runtime errors that disable the surrounding skill at inference, and the resulting interface inconsistency erodes the gains from healthy skills.

Together, the four non-default rows trace a clean ordering: full filtering ($60.3$) $>$ no evolution ($59.3$) $>$ exec-only ($48.8$) $>$ teacher-only ($47.2$) $>$ no filtering ($36.3$). Both gates are necessary: executable validation prevents broken or ill-typed skills, teacher review filters conceptually weak ones, and removing either gate falls back to a regime where the noise added by evolution outweighs the benefit of new skills. The pattern justifies our policy of treating self-improving PF evolution as a \emph{validated and rate-limited} process rather than unconstrained skill generation.

\begin{figure*}[!t]
\centering
\includegraphics[width=0.95\textwidth]{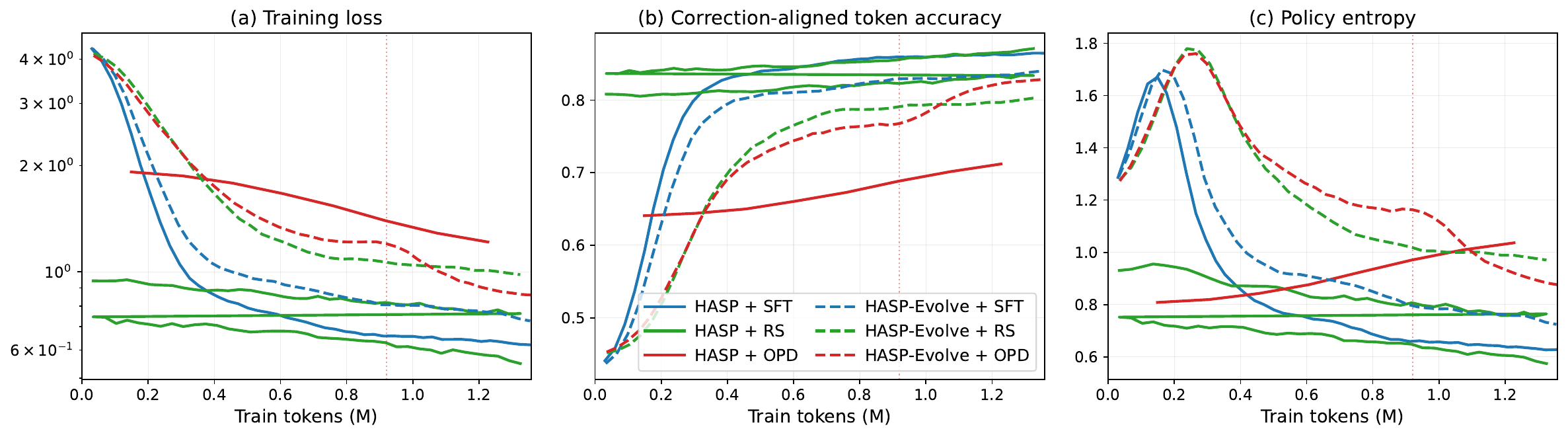}
\vspace{-2mm}
\caption{
Training dynamics for the six post-training settings over processed tokens.
We report training loss, correction-aligned token accuracy, and policy entropy.
Colors indicate training recipe: SFT (blue), RS (green), and OPD (red); solid and dashed lines denote fixed and evolving PF libraries.
}
\vspace{-2mm}
\label{fig:training_dynamics}
\end{figure*}

\begin{figure*}[!ht]
\centering
\includegraphics[width=0.8\textwidth]{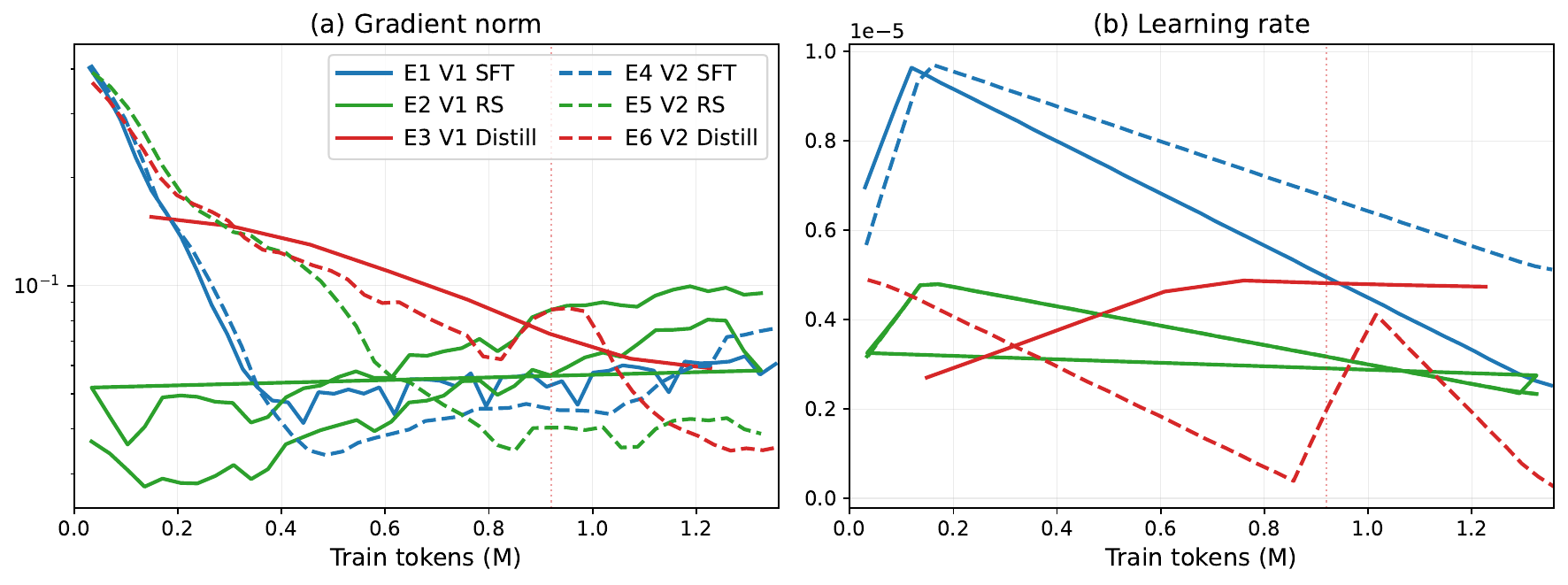}
\vspace{-4mm}
\caption{
Optimization diagnostics for the six post-training settings.
\textbf{Left:} gradient norm as a function of processed training tokens.
\textbf{Right:} learning rate as a function of processed training tokens.
These curves help distinguish genuine differences in training dynamics from artifacts of optimization instability or scheduler choice, and provide additional context for the phase transitions observed in the main training plots.
}
\vspace{-4mm}
\label{fig:opt_diagnostics}
\end{figure*}

\begin{figure*}[!ht]
\centering
\includegraphics[width=0.95\textwidth]{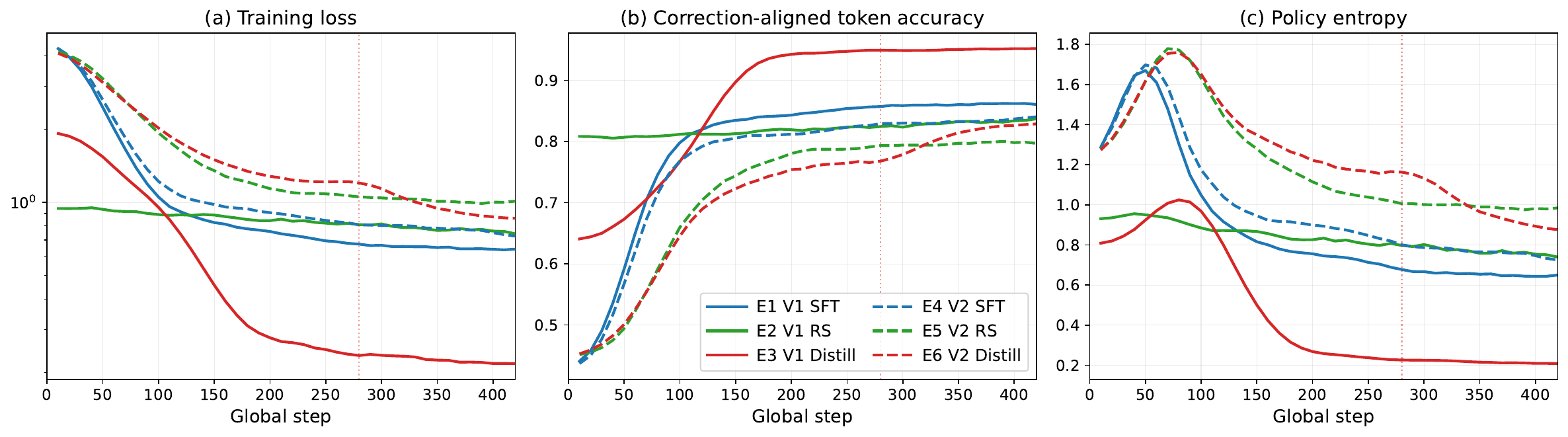}
\vspace{-4mm}
\caption{
Training dynamics of all six post-training settings plotted against global step.
From left to right, we report training loss, correction-aligned token accuracy, and policy entropy.
This figure complements Figure~\ref{fig:training_dynamics} by showing that the qualitative differences across recipes remain visible under a step-based view rather than only under token normalization.
}
\vspace{-4mm}
\label{fig:train_global_step_all}
\end{figure*}

\begin{figure*}[!ht]
\centering
\includegraphics[width=0.95\textwidth]{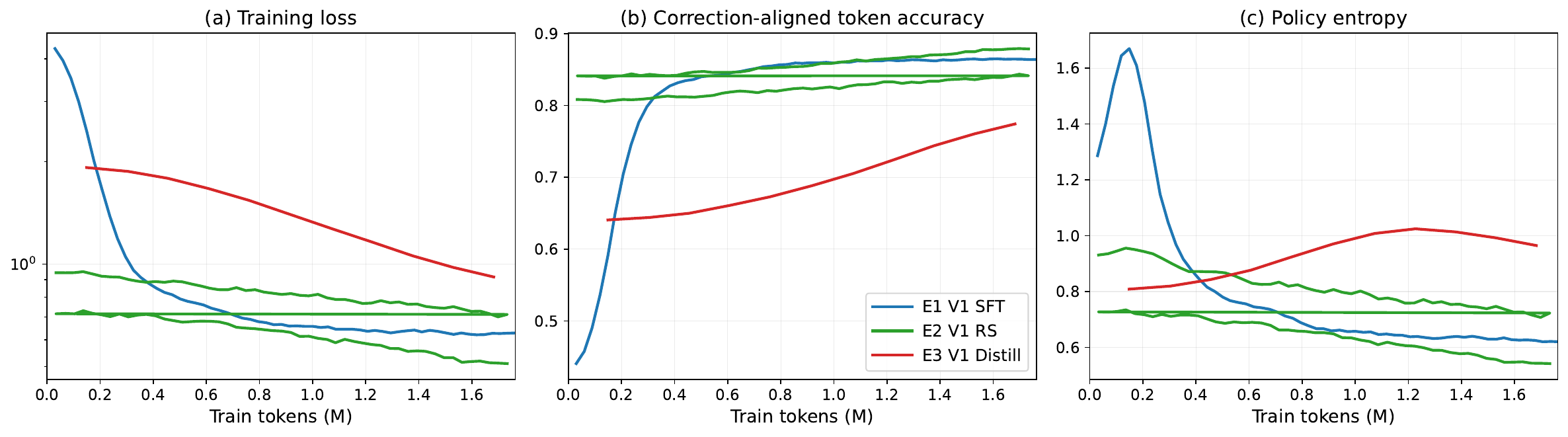}
\vspace{-4mm}
\caption{
Open-loop training dynamics under a fixed PF library.
We compare supervised fine-tuning, rejection sampling, and on-policy distillation in terms of training loss, correction-aligned token accuracy, and policy entropy as functions of processed training tokens.
By removing closed-loop evolution from the comparison, this figure isolates how the three post-training recipes differ when trained on the same fixed PF-aware rollout pool.
}
\label{fig:train_tokens_v1}
\vspace{-4mm}
\end{figure*}

\begin{figure*}[!ht]
\centering
\includegraphics[width=0.95\textwidth]{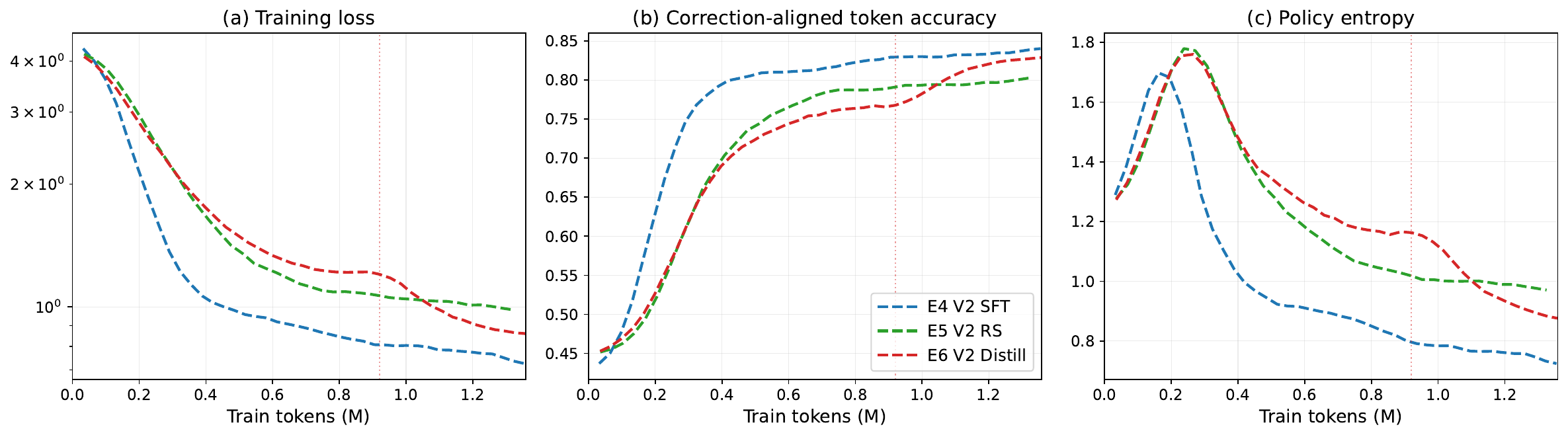}
\vspace{-4mm}
\caption{
Closed-loop training dynamics under evolving PF libraries.
We compare supervised fine-tuning, rejection sampling, and on-policy distillation in terms of training loss, correction-aligned token accuracy, and policy entropy as functions of processed training tokens.
The vertical dotted marker indicates the first refresh point of the closed-loop pipeline.
This figure highlights how evolving PF supervision changes the optimization trajectory relative to the fixed-library setting.
}
\vspace{-4mm}
\label{fig:train_tokens_v2}
\end{figure*}

\subsection{Case Study}
\label{app:case_study}

To complement the aggregated recovery statistics in Table~\ref{tab:failure-success}, this section walks through one concrete trajectory per domain in which the baseline (RA-Agent multi-loop) fails and our PF-augmented agent succeeds. For each case we render the \emph{full} ReAct rollout: at every step we show the agent's internal \emph{thought}, the proposed action, the observation returned, and (for our agent) every PF activation event with its trigger, intervention type, and the before/after action or context. The intent is to make the corrective mechanism visible in the policy loop, not just in the aggregated statistics. All cases use Qwen2.5-7B-Instruct with the rollout configuration of \S\ref{app:exp_details}; the trajectories are taken verbatim from the evaluation logs.

\begin{table}[!ht]
\centering
\footnotesize
\setlength{\tabcolsep}{4pt}
\renewcommand{\arraystretch}{1.05}
\caption{
Recovered failure cases and their corresponding corrective mechanisms. Percentages are computed over recovered cases only. The largest gains come from resolving entity confusion and improving search under insufficient exploration, followed by corrected reasoning and deeper reading.}
\label{tab:failure-success}
\resizebox{\columnwidth}{!}{%
\begin{tabular}{%
@{}>{\raggedright\arraybackslash}p{0.40\columnwidth}
>{\raggedright\arraybackslash}p{0.38\columnwidth}
>{\centering\arraybackslash}p{0.18\columnwidth}@{}}
\toprule
Failure Type & Corrective Mechanism & Coverage (\%) \\
\midrule
\texttt{entity\_conf} & \texttt{resolved\_entity} & \cellcolor{cPos3}\textbf{37--43} \\
\texttt{retrieval + insuff\_explore} & \texttt{better\_search} & \cellcolor{cPos2}29--33 \\
\texttt{reasoning} & \texttt{corrected\_reasoning} & \cellcolor{cPos2}15 \\
\texttt{read\_comp} & \texttt{deeper\_reading} & \cellcolor{cPos1}8--9 \\
\texttt{format} & \texttt{improved\_format} & \cellcolor{cPos1}5 \\
\texttt{hallucination} & --- & ${<}1$ \\
\bottomrule
\end{tabular}%
}
\end{table}

\subsubsection*{Web-search case: HotpotQA --- multi-hop entity resolution}

The question is a three-hop query (\emph{Walton family member} $\to$ \emph{died after John in 2005} $\to$ \emph{her husband}). Both agents use the same SEARCH/READ/FINAL action space and the same teacher-augmented retriever, but only our agent decomposes the query, forces a READ before FINAL, and verifies completeness.

\begin{failbox}
\begin{lstlisting}[basicstyle=\ttfamily\tiny]
Q: Who was the husband of the prominent Walton family member who
   died after John died in 2005?
Gold: Sam Walton

==== ReAct loop (RA-Agent multi-loop, no PFs, no skills) ====

[step 0]  THINK
  "John Walton is the son of Sam Walton, the Wal-Mart founder.
   He died in 2005. The 'prominent Walton family member who died
   after John' is likely Alice Walton or Helen Walton. Let me check."
  ACTION   SEARCH("Walton family members died after 2005")
  OBS      doc_0  John T. Walton ; June 27, 2005 ...
           doc_1  John Walton, a billionaire ... died Monday in plane crash
           doc_2  Wal-Mart Stores ... heir Sam Walton family ...
  [no READ; no decomposition; agent commits to a guess]

[step 1]  THINK
  "Alice Walton is the most prominent. Let me check if she died."
  ACTION   SEARCH("Alice Walton death date")
  OBS      doc_0  Alice Louise Walton (born October 7, 1949) is
                  an American billionaire... [she is alive]
           doc_1  Alice Louise Walton was born on October 7, 1949...

[step 2]  THINK
  "Search did not return a death date but Alice is the only
   prominent Walton I know."
  ACTION   SEARCH("Alice Walton death date")    # SAME query repeated
  OBS      [identical results; no new evidence]

[step 3]  FINAL
  ANSWER   "Bruce Walton"
  REASON   "Alice Walton (born 1949) ... is still alive, but the
            question implies a Walton family member died after
            John (2005). The only plausible name I recall is
            Bruce Walton."         # FABRICATED
\end{lstlisting}
\end{failbox}

\begin{winbox}
\begin{lstlisting}[basicstyle=\ttfamily\tiny]
==== ReAct loop with PF library (skills_top10_with_prompt) ====

[pre-rollout]  PFSelector picks top-K relevant skills for this
               question (decompose_complex_question, retrieval_failure,
               insufficient_exploration, answer_completeness, ...);
               three mandatory PFs always included.

[step 0]  THINK
  "Multi-hop question: (1) which Walton died after 2005, (2) who
   was her husband. I will search broader first."
  PROPOSED  SEARCH("Walton family member died after John Walton 2005")

  >> [PF FIRED] decompose_complex_question -> INJECT_CONTEXT
     trigger : multi-hop question with insufficient sub-question
               coverage (matched by SkillSelector)
     before  : (action unchanged)
     after   : observation will be augmented with
       "[DECOMPOSITION HINT] This is a multi-hop question. Search
        for each piece of information separately. Find intermediate
        entities first, then search for the final answer."

  EXECUTED  SEARCH("Walton family member died after John Walton 2005")
  OBS       doc_0  Christy Walton took her husband John's place in
                   the ranking after his death...
            doc_1  John Walton, the son of Wal-Mart founder Sam
                   Walton... died Monday in a plane crash in 2005.
            doc_2  Helen Walton ... wife of Sam Walton ...
            (+ injected DECOMPOSITION HINT)

[step 1]  THINK
  "Plan: hop-1 is to find a Walton who died AFTER 2005. The hint
   said find intermediate entities first. Helen Walton is plausible."
  ACTION    SEARCH("Helen Walton death date")
  OBS       doc_0  Died, April 19, 2007 (aged 87). Bentonville,
                   Arkansas, U.S. ... wife of Sam Walton.
            doc_1  On April 19, 2007, after heart failure...
                   Walton died ... at age eighty-seven.

[step 2]  THINK
  "Helen Walton died 2007 (after 2005). Hop-2 is her husband.
   I should READ the document to confirm."
  PROPOSED  FINAL("Sam Walton")        # premature FINAL

  >> [PF FIRED] insufficient_exploration -> MODIFY_ACTION
     trigger : action_type=FINAL with read_count=0
     reason  : "Searched but never read; forcing read"
     before  : FINAL("Sam Walton")
     after   : READ(doc_0)            # action overridden

  EXECUTED  READ(doc_0)
  OBS       Helen Walton ... died April 19, 2007 ... was the
            wife of Wal-Mart founder Sam Walton ...

[step 3]  THINK
  "Confirmed by reading: Helen Walton died 2007 after John (2005),
   her husband was Sam Walton."
  PROPOSED  FINAL("Sam")

  >> [PF FIRED] answer_completeness -> INJECT_CONTEXT
     trigger : multi-part question with single-token answer
     before  : answer="Sam"
     after   : observation augmented with
       "[COMPLETENESS WARNING] The question appears to have
        multiple parts, but your answer may be incomplete.
        Make sure you address all parts."

  ACTION    FINAL("Sam Walton")        # CORRECT
  REASON    "Helen Walton died after John Walton (2005) and was
             the wife of Sam Walton. Verified through document
             reading."
\end{lstlisting}
\end{winbox}

\noindent\textbf{Phase-by-phase comparison.} The baseline's three steps reproduce the canonical multi-step failure modes targeted by Phase~B (\S\ref{app:phase_b}): \texttt{no\_read\_before\_final} (it never READs), \texttt{repeated\_search} (steps~1 and~2 issue identical queries), and \texttt{reasoning\_hallucination} at FINAL (the answer ``Bruce Walton'' is not present in any retrieved snippet). Our agent triggers three PFs at three distinct loop locations:
(i) \emph{before} step~0's SEARCH, \texttt{decompose\_complex\_question} fires as an \texttt{INJECT\_CONTEXT} that surfaces a decomposition hint into the next observation, steering the agent toward hop-by-hop search;
(ii) \emph{at} step~2, the agent's proposed action is FINAL with no prior READ, so \texttt{insufficient\_exploration} fires as a \texttt{MODIFY\_ACTION} that overrides the action from \texttt{FINAL("Sam Walton")} to \texttt{READ(doc\_0)} --- the explicit before/after pair is \texttt{(action=FINAL,arg="Sam Walton") $\to$ (action=READ,arg="doc\_0")};
(iii) \emph{at} step~3, after the agent emits a single-token \texttt{FINAL("Sam")}, \texttt{answer\_completeness} fires as an \texttt{INJECT\_CONTEXT} reminding the policy that the question has multiple parts, prompting the model to expand to the gold-standard form \texttt{"Sam Walton"}. Each intervention has an inspectable trigger, an explicit before-action / after-action pair, and a logged reason --- the trajectory itself is the audit trail. None of these three corrections is a stronger model: PFs intervene on the \emph{same} Qwen-7B policy that produced the baseline.

\subsubsection*{Math case: AMC23 --- distinct-roots polynomial counting}

For math, the rollout is single-step (\texttt{max\_steps $= 1$}, \S\ref{app:libraries}), so the entire ReAct loop is one \texttt{<think>}/\texttt{FINAL} pair. Skills enter through (a) \texttt{system\_summary} prompts injected into the system message during \texttt{PFSelector} stage, and (b) a \texttt{pre\_final} verifier hook on the FINAL action. Both agents see the same problem; the only difference is whether the math skill library is active.

\begin{failbox}
\begin{lstlisting}[basicstyle=\ttfamily\tiny]
Q: For how many ordered pairs (a,b) of integers does the polynomial
   x^3 + a*x^2 + b*x + 6  have 3 distinct integer roots?
Gold: 5

==== ReAct loop (RA-Agent, no skills) ====

[step 0]  THINK
  "Let roots be p,q,r. By Vieta:
     p+q+r = -a,  pq+qr+rp = b,  pqr = -6.
   Enumerate ORDERED triples (p,q,r) with pqr=-6 and distinct:
     (1, 2,-3) -> (a,b) = (0,-7)
     (1,-2, 3) -> (a,b) = (-2,-5)
     (1,-1, 6) -> (a,b) = (-6,-1)
     (1,-1,-6) -> (a,b) = (6,-1)
     (1,-3, 2) -> (a,b) = (0,-7)
     (1,-6,-1) -> (a,b) = (6,-1)
     (2,-1, 3) -> (a,b) = (-4, 1)
     (2,-3,-1) -> (a,b) = (2,-5)
     (3,-1,-2) -> (a,b) = (0,-7)
   Each ORDERING gives a unique pair (a,b)."     # FALSE
  ACTION   FINAL("9")                              # WRONG
\end{lstlisting}
\end{failbox}

\begin{winbox}
\begin{lstlisting}[basicstyle=\ttfamily\tiny]
==== ReAct loop with math skill library (11 PF skills active) ====

[pre-rollout]  System prompt augmented with system_summary lines
               from selected math skills, including:
                 case_incompleteness    : "cases must be exhaustive
                                          AND mutually exclusive"
                 overgeneralization     : "check formula preconditions"
                 verification_missing   : "plug back into the
                                          ORIGINAL equation"

[step 0]  THINK
  "Let roots be p,q,r. By Vieta:
     p+q+r = -a,  pq+qr+rp = b,  pqr = -6.
   Enumerate triples (p,q,r) with pqr=-6 and distinct."
  Initial enumeration (8 ordered triples)...

  >> [PF FIRED] case_incompleteness  /  overgeneralization
     trigger     : has_case_analysis = True at pre_final
     before      : draft answer counts every ordered triple as a
                    distinct (a,b) pair
     intervention: phase instruction injected at pre_final:
       "Cases must be exhaustive AND mutually exclusive. Vieta is
        symmetric in (p,q,r), so permutations of the SAME multiset
        of roots map to the SAME (a,b). Reduce orderings to
        multisets before counting."

  REVISED REASONING
   Multisets {p,q,r}: {1,2,-3}, {-1,1,6}, {-1,1,-6}, {-1,2,3},
                      {-1,-2,3}.   (5 distinct multisets)
   Discard {-1,-1,6} and {-1,-1,-6}: not 3-distinct.
   Consolidated (a,b) pairs:
     (0,-7), (-2,-5), (2,-5), (-4,-11), (4,-11)

  >> [PF FIRED] verification_missing -> pre_final hook
     trigger     : always (priority 0.90)
     intervention: substitute each candidate (a,b) back into
       x^3 + a*x^2 + b*x + 6 and confirm 3 distinct integer roots
     check       : x^3 - 7x + 6 = (x-1)(x-2)(x+3)   [OK]
                   x^3 - 2x^2 - 5x + 6 = (x-1)(x-3)(x+2)   ...
                   ... 5 distinct (a,b) pairs all verified.

  ACTION   FINAL("5")                                  # CORRECT
\end{lstlisting}
\end{winbox}

\noindent\textbf{Phase-by-phase comparison.} Both agents share the same Vieta setup; the difference is structural rather than mathematical. The baseline's failure is captured exactly by the heuristic detector \texttt{case\_incompleteness} of \S\ref{app:phase_b}: it lists nine \emph{orderings} but the question counts \emph{ordered $(a,b)$ pairs}, and Vieta's symmetric polynomials collapse permutations of the same root multiset onto the same $(a,b)$. Two PFs intervene through different surfaces: \texttt{case\_incompleteness} and \texttt{overgeneralization} fire jointly as a \texttt{pre\_final} phase instruction (an \texttt{INJECT\_CONTEXT}-style intervention) that surfaces a constraint reminder \emph{before} the model commits its count; \texttt{verification\_missing} (priority $0.90$, always-on) fires as a \texttt{pre\_final} verification hook that forces a substitution check on each candidate $(a,b)$. The before/after for the agent's draft is \texttt{(count = 9 over orderings) $\to$ (count = 5 over consolidated multisets, each substitution-verified)}. As in the search case, no stronger model is invoked --- the same Qwen-7B policy is steered onto a self-consistent counting frame.

\subsubsection*{Coding case: LiveCodeBench-easy --- ``Equally''}

Coding rollouts use \texttt{max\_steps $= 1$} as well; PFs run as static-analysis hooks on the emitted code body and emit \texttt{INJECT\_CONTEXT} hints rather than rewriting code. The visible repair is the agent's revised solution after the static check fires.

\begin{failbox}
\begin{lstlisting}[basicstyle=\ttfamily\tiny]
Problem (LiveCodeBench-easy / atcoder_abc385_a, "Equally"):
  Given three integers A, B, C in [1, 1000], decide whether they
  can be partitioned into two or more groups with equal sums.
  Print "Yes" or "No".

  Sample 1: 3 8 5  -> Yes      (groups {3,5} and {8}, sum=8)
  Sample 2: 2 2 2  -> Yes      (groups {2}, {2}, {2}, sum=2)
  Sample 3: 1 2 4  -> No

==== Single-step ReAct (no skills) ====

[step 0]  THINK
  "[1] RESTATE: divide A,B,C into >=2 groups with equal sums.
   [2] FORMAT: stdin script with input()/print().
   [3] BRANCHES:
       - sum is odd                 -> No
       - sum == 0                   -> No
       - any two sum to the third   -> Yes
       - otherwise                  -> No"
  ACTION   FINAL("""
    a, b, c = map(int, input().split())
    if (a + b + c) % 2 == 1:
        print("No")
    elif a + b + c == 0:
        print("No")
    elif a + b == c or a + c == b or b + c == a:
        print("Yes")
    else:
        print("No")
  """)

JUDGE     test 0 (3 8 5)  passed  (3+5==8)
          test 1 (2 2 2)  FAILED  expected Yes, got No
                                  (2+2==4 != 2; falls through)
          ...
          pass_at_1 = False    # MISSED THE all-equal CASE
\end{lstlisting}
\end{failbox}

\begin{winbox}
\begin{lstlisting}[basicstyle=\ttfamily\tiny]
==== Single-step ReAct with code skill library (12 PFs) ====

[pre-rollout]  System prompt augmented with code_edge_cases,
               code_test_walkthrough, code_off_by_one ...

[step 0]  THINK
  "[1] RESTATE: ...
   [3] BRANCHES:
       - all equal             -> Yes   <-- missed by baseline
       - sum odd               -> No
       - sum <= 3              -> No
       - sum == 4 or 6         -> Yes
       - any two sum to third  -> Yes
       - otherwise             -> No"

  >> [PF FIRED] code_edge_cases -> INJECT_CONTEXT
     trigger     : FINAL with stdin parsing + branching code
     intervention: phase instruction reminds:
       "Trace each sample input through your branches before FINAL.
        For (2,2,2): each integer can be its own group of size 1
        (k = 3 groups), so the answer is Yes whenever a == b == c.
        The 'a+b == c' test does NOT cover this case."

  REVISED CODE  (the all-equal branch is added FIRST)
  ACTION   FINAL("""
    a, b, c = map(int, input().split())
    if a == b == c:                # added by code_edge_cases hint
        print("Yes")
    elif (a + b + c) % 2 == 1:
        print("No")
    elif a + b + c <= 3:
        print("No")
    elif a + b + c == 4 or a + b + c == 6:
        print("Yes")
    elif a + b == c or a + c == b or b + c == a:
        print("Yes")
    else:
        print("No")
  """)

JUDGE     test 0 (3 8 5)  passed
          test 1 (2 2 2)  passed   <- recovered by all-equal branch
          test 2 (1 2 4)  passed
          ...
          pass_at_1 = True
\end{lstlisting}
\end{winbox}

\noindent\textbf{Phase-by-phase comparison.} The baseline's solution misses one branch of the case analysis: when $A=B=C$, the partition into three singleton groups always works, but its disjunction \texttt{a+b==c or a+c==b or b+c==a} only covers two-vs-one partitions. The corresponding heuristic in Phase~B is \texttt{partial\_answer} / \texttt{case\_incompleteness}, and the runtime PF \texttt{code\_edge\_cases} catches it through a static check on the FINAL code body (branching disjunction without an all-equal short-circuit) and injects a context hint that explicitly walks (2,2,2) through the branches. The intervention type is \texttt{INJECT\_CONTEXT}, not \texttt{MODIFY\_ACTION} --- the framework does not rewrite the student's code; the policy itself adds the \texttt{a == b == c} branch in response to the hint. The before/after pair is \texttt{(branches: \{odd, zero, two-of-three\}) $\to$ (branches: \{all-equal, odd, sum$\le 3$, sum$\in\{4,6\}$, two-of-three, else\})}, with the only structural change being the all-equal branch added at the top.


\section{Training Dynamics}
\label{app:training_dynamics}

This section unpacks the training-dynamics figures, makes the data-collection protocol explicit, and reads each curve back to a property of the recipe rather than treating the plots as decoration. All six runs use the same Qwen2.5-7B-Instruct LoRA fine-tuning configuration of \S\ref{app:exp_details} (rank $16$, $\alpha=32$, bf16, gradient checkpointing; the recipe-specific knobs (learning rate, number of epochs, rollout shape) are summarized in Table~\ref{tab:e1e6_configs}.

\subsection{Data-Collection Protocol}

The curves in Figures~\ref{fig:training_dynamics} and \ref{fig:opt_diagnostics}--\ref{fig:train_tokens_v2} are produced from Weights~\&~Biases run logs. The trainer (TRL's \texttt{SFTTrainer} for E1/E4 and the rejection-sampling/distillation drivers for E2/E3/E5/E6) emits one JSON-style \texttt{loss}-dict per logging interval (\texttt{logging\_steps $= 10$}) into \texttt{output.log}. Each emitted dict reports the seven scalars we plot:
\texttt{loss}, \texttt{grad\_norm}, \texttt{learning\_rate}, \texttt{entropy}, \texttt{num\_tokens}, \texttt{mean\_token\_accuracy}, and \texttt{epoch}. The three primary signals in the main-text figure correspond to (i) cross-entropy training loss, (ii) per-step \emph{correction-aligned token accuracy} (the trainer's \texttt{mean\_token\_accuracy}, computed over the PF-corrected target tokens, $\tilde a_t$, rather than the raw policy outputs), and (iii) on-policy generation entropy as estimated by TRL during the forward pass. Curves are smoothed with a centered moving-average window of size $5$ to suppress per-step noise without distorting the slow-time-scale shape.

For each E we plot one canonical run --- the longest single Wandb run available for V1, and the concatenation of \texttt{iter\_0} and \texttt{iter\_1} for the closed-loop V2 settings (the smallest sequence that captures the first evolution boundary). Concretely the canonical traces are: E1 V1 SFT for $600$ optimizer steps over $\approx 10$ epochs; E2 V1 RS for $1{,}400$ optimizer steps over $8$ epochs $\times$ $2$ RS iterations; E3 V1 OPD for $900$ optimizer steps over $10$ epochs; E4 V2 SFT as $820$ + $180$ steps across two closed-loop iterations; E5 V2 RS as $700$ + $360$ steps; and E6 V2 OPD as $280$ + $140$ steps. Token counts (\texttt{num\_tokens}) are accumulated across iterations so the V2 X-axes remain monotone; the iter-boundary in each V2 series is rendered as a vertical dotted marker in the run's own color.

\subsection{Optimization Diagnostics}

Figure~\ref{fig:opt_diagnostics} reports gradient norm and learning rate against processed tokens. The gradient-norm panel verifies that the loss differences across recipes are not artifacts of unstable optimization: all six runs sit in the same magnitude band ($\sim10^0$ to $10^{1}$) with no run-specific spikes at the iter-boundaries, ruling out the hypothesis that the small loss bumps in V2 reflect numerical issues. The learning-rate panel makes the schedule explicit: SFT uses $1\times 10^{-5}$ peak (E1, E4); RS and OPD use $5\times 10^{-6}$ (E2, E3, E5, E6) because they operate on top of an SFT-warmed policy; warmup is $3\%$ of total tokens for all six. Closed-loop runs apply the schedule per iter rather than globally, so each V2 curve shows a fresh warmup at the iter-boundary --- the iter-internal LR pattern is repeated, not flattened. Reading the optimization panel together with the loss panel makes clear that the recipe-level differences in convergence rate and final level reflect the objective rather than the optimizer.

\subsection{Loop-Topology Comparisons}

Three companion figures separate the data along orthogonal axes. Figure~\ref{fig:train_global_step_all} replots all six runs against \emph{global step} rather than tokens to confirm that recipe ordering is robust to X-axis choice: V2 runs span fewer steps but more wall-clock seconds because the evolve cycle inserts inference and skill-proposal stages between training phases. Figures~\ref{fig:train_tokens_v1} and \ref{fig:train_tokens_v2} each show only three curves --- the V1-only and V2-only subsets respectively --- which makes the recipe-vs-recipe contrast easier to see without cross-topology visual clutter. The vertical dotted marker in Figure~\ref{fig:train_tokens_v2} marks the first closed-loop refresh point in each V2 run. The most informative comparison is V1 SFT (Figure~\ref{fig:train_tokens_v1}) versus V2 SFT (Figure~\ref{fig:train_tokens_v2}): the V1 curve is smooth, the V2 curve shows a cleanly visible break at the iter-boundary, and after the break the slope flattens because the post-evolve target pool has higher accuracy targets to reach. This is direct evidence that the closed-loop refresh changes the gradient signal rather than merely re-running the same data.

\section{Additional Analysis of Skill Evolution}
\label{app:evolution}

This section reports per-domain statistics of the self-improving PF library that complement the framework definitions in Appendix~\ref{app:self_improving}. The numbers below are read directly from the on-disk artifacts of the closed-loop runs: \texttt{library\_history.json} for per-skill admission events, \texttt{generated/\{skill\_id\}\_\_v\{N\}/metadata.json} for per-skill review scores, and \texttt{snapshots/epoch\_\{i\}/} for per-epoch library state. Table~\ref{tab:stage2b-skill-evolution} aggregates the corresponding web-search uplift; the per-skill review-score breakdown is given below in Table~\ref{tab:per_family_scores} (per-family) and Table~\ref{tab:filter_outcome} (per-domain admission counts).

\begin{table*}[!ht]
\centering
\scriptsize
\setlength{\tabcolsep}{3.2pt}
\renewcommand{\arraystretch}{1.18}
\tiny
\caption{
Effect of self-improving PF evolution on the web-search benchmark.
Rows correspond to skill families introduced across evolution epochs.
Each epoch cell reports the gain over the non-evolving baseline as $\Delta$EM / $\Delta\mathrm{MBE}_{\text{LLM-judge}}$.
The bottom block reports average review scores of generated skills across five quality dimensions.}
\label{tab:stage2b-skill-evolution}
\resizebox{\textwidth}{!}{
\begin{tabular}{l ccccc c}
\toprule
\multirow{2}{*}{\textbf{Skill Family}}
& \multicolumn{5}{c}{\textbf{Evolution Epoch}} 
& \multirow{2}{*}{\makecell{\textbf{Avg.}\textbf{$\Delta$}}} \\
\cmidrule(lr){2-6}
& \textbf{E0} & \textbf{E1} & \textbf{E2} & \textbf{E3} & \textbf{E4} & \\
\midrule

\makecell[l]{\texttt{reasoning}\texttt{\_hallucination}}
  & \cellcolor{cPos3}\makecell{$+16.7$ / $+11.7$}
  & \cellcolor{cPos2}\makecell{$+11.7$ / $+3.3$}
  & \cellcolor{cPos2}\makecell{$+10.0$ / $+6.7$}
  & \cellcolor{cPos3}\makecell{$+15.0$ / $+11.7$}
  & \cellcolor{cPos3}\makecell{$+13.3$ / $+13.3$}
  & \cellcolor{cPos3}\makecell{$+13.3$ / $+9.3$} \\

\makecell[l]{\texttt{no\_read}\texttt{\_before\_final}}
  & --
  & \cellcolor{cZero}\makecell{$-1.7$ / $+1.7$}
  & \cellcolor{cPos1}\makecell{$+1.7$ / $+5.0$}
  & \cellcolor{cNeg1}\makecell{$\phantom{+}0.0$ / $-1.7$}
  & \cellcolor{cZero}\makecell{$+1.7$ / $-1.7$}
  & \cellcolor{cPos1}\makecell{$+0.4$ / $+0.8$} \\

\makecell[l]{\texttt{premature}\texttt{\_final}}
  & --
  & --
  & \cellcolor{cPos1}\makecell{$+5.0$ / $+1.7$}
  & \cellcolor{cPos3}\makecell{$+15.0$ / $+5.0$}
  & \cellcolor{cPos2}\makecell{$+13.3$ / $+6.7$}
  & \cellcolor{cPos2}\makecell{$+11.1$ / $+4.4$} \\

\makecell[l]{\texttt{pf\_override}\texttt{\_harmful}}
  & --
  & --
  & --
  & \cellcolor{cNeg1}\makecell{$\phantom{+}0.0$ / $-1.7$}
  & \cellcolor{cNeg1}\makecell{$-1.7$ / $\phantom{+}0.0$}
  & \cellcolor{cNeg1}\makecell{$-0.8$ / $-0.8$} \\

\makecell[l]{\texttt{wrong\_entity}\texttt{\_focus}}
  & --
  & --
  & --
  & --
  & \cellcolor{cPos2}\makecell{$+6.7$ / $+5.0$}
  & \cellcolor{cPos2}\makecell{$+6.7$ / $+5.0$} \\

\midrule
\addlinespace[2pt]
\multicolumn{7}{c}{\textbf{Generated skill review scores (scale $0$--$1$)}} \\
\midrule
\textbf{Dimension}
  & Concept
  & Trigger
  & Intervention
  & Executability
  & Validation
  & \textbf{Avg.} \\
\textbf{Score}
  & \cellcolor{cPos2}$0.83$
  & \cellcolor{cPos1}$0.75$
  & \cellcolor{cPos2}$0.83$
  & \cellcolor{cPos3}$0.95$
  & \cellcolor{cPos2}$0.80$
  & \cellcolor{cPos2}$0.83$ \\
\bottomrule
\end{tabular}}
\end{table*}

\subsection{Per-Family Review Scores}
\label{app:per_family_scores}

Recall from \S\ref{app:phase_e} that each candidate skill is rated by the teacher LLM on five dimensions: concept ($Q_{\mathrm{concept}}$), trigger ($Q_{\mathrm{trigger}}$), intervention ($Q_{\mathrm{intervene}}$), executability ($Q_{\mathrm{exec}}$), and validation utility ($Q_{\mathrm{val}}$). The composite is $Q_{\mathrm{skill}} = 0.25 Q_{\mathrm{concept}} + 0.20 Q_{\mathrm{trigger}} + 0.20 Q_{\mathrm{intervene}} + 0.20 Q_{\mathrm{exec}} + 0.15 Q_{\mathrm{val}}$. Table~\ref{tab:per_family_scores} averages each dimension across all accepted candidates of the same skill family in the corresponding domain; ``$n$'' is the number of accepted versions of that family across all evolution rounds.

\begin{table*}[!ht]
\centering
\scriptsize
\setlength{\tabcolsep}{4.5pt}
\renewcommand{\arraystretch}{1.10}
\caption{Per-family average review scores across all accepted candidates of each skill family. \textbf{C}/\textbf{T}/\textbf{I}/\textbf{E}/\textbf{V} are concept / trigger / intervention / executability / validation; $Q_{\mathrm{skill}}$ is the weighted composite. Counts ($n$) are the total accepted versions over all rounds.}
\label{tab:per_family_scores}
\begin{tabular}{l c ccccc c}
\toprule
\textbf{Skill family} & $n$ & \textbf{C} & \textbf{T} & \textbf{I} & \textbf{E} & \textbf{V} & $\boldsymbol{Q_{\mathrm{skill}}}$ \\
\midrule
\multicolumn{8}{l}{\textit{Web-search domain}} \\
\texttt{reasoning\_hallucination} & 5 & 0.90 & 0.80 & 0.92 & 0.98 & 0.86 & \textbf{0.89} \\
\texttt{premature\_final}         & 3 & 0.90 & 0.80 & 0.83 & 1.00 & 0.83 & 0.88 \\
\texttt{no\_read\_before\_final}  & 4 & 0.85 & 0.72 & 0.85 & 0.95 & 0.80 & 0.84 \\
\texttt{pf\_override\_harmful}    & 2 & 0.80 & 0.70 & 0.90 & 1.00 & 0.80 & 0.84 \\
\texttt{wrong\_entity\_focus}     & 4 & 0.80 & 0.78 & 0.72 & 0.95 & 0.75 & 0.80 \\
\midrule
\multicolumn{8}{l}{\textit{Math domain}} \\
\texttt{reasoning\_hallucination} & 3 & 0.93 & 0.83 & 0.87 & 1.00 & 0.83 & \textbf{0.90} \\
\texttt{premature\_final}         & 3 & 0.90 & 0.83 & 0.87 & 1.00 & 0.80 & 0.89 \\
\texttt{wrong\_entity\_focus}     & 3 & 0.80 & 0.67 & 0.77 & 0.93 & 0.70 & 0.78 \\
\texttt{format\_mismatch}         & 3 & 0.77 & 0.67 & 0.70 & 0.90 & 0.67 & 0.74 \\
\midrule
\multicolumn{8}{l}{\textit{Code domain}} \\
\texttt{reasoning\_hallucination} & 12 & 0.94 & 0.87 & 0.87 & 1.00 & 0.83 & \textbf{0.91} \\
\texttt{premature\_final}         & 12 & 0.88 & 0.83 & 0.79 & 0.96 & 0.81 & 0.86 \\
\texttt{format\_mismatch}         & 12 & 0.78 & 0.68 & 0.67 & 0.94 & 0.68 & 0.76 \\
\texttt{incomplete\_implementation} & 6 & 0.80 & 0.70 & 0.67 & 0.92 & 0.70 & 0.76 \\
\texttt{incomplete\_logic}        & 4 & 0.72 & 0.65 & 0.75 & 0.93 & 0.65 & 0.74 \\
\texttt{wrong\_entity\_focus}     & 12 & 0.78 & 0.67 & 0.65 & 0.90 & 0.68 & 0.74 \\
\texttt{incomplete\_solution}     & 1 & 0.80 & 0.70 & 0.60 & 0.90 & 0.70 & 0.74 \\
\texttt{incomplete\_code}         & 1 & 0.80 & 0.70 & 0.60 & 0.90 & 0.70 & 0.74 \\
\bottomrule
\end{tabular}
\end{table*}

Three patterns appear across the three domains. First, \texttt{executability} is consistently the highest-scoring dimension ($0.90$--$1.00$), reflecting the upstream effect of the executable validation gate (\S\ref{app:phase_d}): the teacher rarely sees code that does not import or returns a wrong type, so it can score executability close to its ceiling. Second, \texttt{trigger} is the consistently lowest-scoring dimension ($0.65$--$0.87$), echoing the recurring difficulty of writing a \texttt{should\_activate} predicate that is neither too broad nor too narrow --- the teacher review prompt explicitly asks about both failure modes (Listing~\ref{box:reviewer_user}). Third, two skill families dominate the per-family ranking across all three domains: \texttt{reasoning\_hallucination} ($Q_{\mathrm{skill}} = 0.89$/$0.90$/$0.91$ on web/math/code) and \texttt{premature\_final} ($0.88$/$0.89$/$0.86$). These are also the two families that contribute the largest validation uplift (\texttt{reasoning\_hallucination}: $+9.3$ avg.\ points on web; Table~\ref{tab:stage2b-skill-evolution}), so high review scores correlate with downstream utility rather than being free credit.

\subsection{Per-Domain Filter Outcomes and Admission Counts}
\label{app:filter_outcome}

Table~\ref{tab:filter_outcome} reports library-growth statistics across the three domains. The closed-loop runs span $5$ epochs for web-search and $3$ epochs each for math and code under the lite-evolve schedule of \S\ref{app:evolve_modes}; the proposer is rate-limited to at most five candidates per epoch (\texttt{max\_candidates\_per\_epoch}, \S\ref{app:phase_c}), and the library is hard-capped at $\texttt{max\_library\_size $= 50$}$ active skills. Under the current ``audit-only'' acceptance policy of \S\ref{app:phase_d}--\ref{app:phase_e}, all candidates that compile are appended to the library with their review scores stored alongside, so that closed-loop policies can flip the gate on without re-running rollouts.

\begin{table*}[!ht]
\centering
\small
\setlength{\tabcolsep}{6pt}
\renewcommand{\arraystretch}{1.15}
\caption{Library growth statistics across self-improvement rounds, by domain. ``Mean $Q_{\mathrm{skill}}$'' averages over accepted candidates; ``Min'' and ``Max'' report extremes. Under the strict $Q_{\mathrm{skill}}\ge 0.60$ acceptance threshold of \S\ref{app:phase_e}, every accepted candidate would still be admitted ($\min Q_{\mathrm{skill}}\ge 0.665$ across all three domains).}
\label{tab:filter_outcome}
\begin{tabular}{l c c c c c c}
\toprule
\textbf{Domain} & \textbf{Epochs} & \textbf{Accepted} & \textbf{Distinct families} & \textbf{Mean $Q_{\mathrm{skill}}$} & \textbf{Min} & \textbf{Max} \\
\midrule
Web-search & $5$ & $18$ & $5$ & $0.85$ & $0.745$ & $0.965$ \\
Math       & $3$ & $12$ & $4$ & $0.83$ & $0.740$ & $0.925$ \\
Code       & $3$ & $55$ & $8$ & $0.80$ & $0.665$ & $0.965$ \\
\bottomrule
\end{tabular}
\end{table*}

The web-search and math runs admit between $4$ and $6$ candidates per epoch, comfortably under the per-epoch cap; the code run admits more candidates per epoch ($\sim 18$) because its lite-evolve schedule (\S\ref{app:evolve_modes}) re-samples failed trajectories more aggressively and the proposer emits multiple variants per residual cluster. Even with this faster growth, the active library never exceeds the $50$-skill hard cap, because each admission goes into a versioned slot (\texttt{\_\_v\{N\}}) and only the latest version of each base id (\texttt{skill\_ids\_latest}) is exposed at runtime. Older versions remain on disk for trajectory replay but never fire, so retrieval precision is bounded by the number of distinct skill families ($5$/$4$/$8$ across web/math/code) rather than the cumulative count of accepted versions.

\subsection{Filtering Outcomes Under the Strict-Gate Policy}
\label{app:strict_gate}

Although the current pipeline is in audit-only mode, the on-disk scores let us reconstruct what the strict-gate policy of \S\ref{app:phase_e} would do. With $Q_{\mathrm{exec}} \ge 0.30$ for hard rejection, $Q_{\mathrm{skill}} \ge 0.60$ for acceptance, and $Q_{\mathrm{skill}} \ge 0.75$ for new-category admission, all $85$ candidates accepted across the three domains would still pass the basic acceptance threshold ($\min Q_{\mathrm{skill}} = 0.665$ in code), but only $52$ of $85$ ($61\%$) would clear the new-category bar of $0.75$. The boundary cases concentrate in the families with the lowest review scores --- \texttt{format\_mismatch} (math/code), \texttt{wrong\_entity\_focus} (code), \texttt{incomplete\_logic}, and \texttt{incomplete\_implementation} (code) --- which in our system are generally accepted as refinements of an existing seed family rather than new categories. This is consistent with the multi-tier threshold design: the framework lets a weak refinement past the same-family bar while reserving the higher bar for genuinely new skill concepts. The same-family bar is what controls library noise during rapid evolution, and it is the lever that the filtering ablation in Table~\ref{tab:ablation_all} exercises directly.

\subsection{Skill-Family Evolution Pattern}
\label{app:evolution_patterns}

Across all three domains the evolved library converges on a small set of recurring families that map onto the Phase~B failure-detector taxonomy (\S\ref{app:phase_b}):
\texttt{premature\_final}, \texttt{no\_read\_before\_final}, \texttt{reasoning\_hallucination}, \texttt{wrong\_entity\_focus}, and \texttt{format\_mismatch} dominate the web-search and math libraries, and the same five plus \texttt{incomplete\_implementation} / \texttt{incomplete\_logic} dominate code. This convergence is informative: the proposer (the student vLLM) is choosing skill ids that overlap heavily with the heuristic detectors of Phase~B even though it sees the failure clusters as natural-language abstractions, not as named heuristics. In other words, the student rediscovers the failure taxonomy from data rather than copying it from labels --- which we read as a sanity check that the closed-loop signal is not collapsing to the proposer reproducing a fixed list. The cross-domain reuse of family names also explains why the lite-evolve mode is sufficient for E5/E6: most epochs add a refinement of an already-named family rather than a genuinely new concept, and the executable validation pipeline (\S\ref{app:phase_d}) is what keeps refinements compatible with the existing dispatcher.

\section{Compute, Hardware, and Reproducibility}
\label{app:compute}

This section consolidates the resource requirements, hardware specifications, and reproducibility-critical metadata for all experiments in the paper, in line with the NeurIPS reproducibility checklist.

\subsection{Hardware Configuration}

All training and evaluation were run on SLURM cluster constrained to NVIDIA \texttt{l40s}. All hardware allocations are summarized in Table~\ref{tab:compute_hardware}. Memory is RSS only (no offload), and \texttt{vLLM} engines reuse the GPU when shared between rollout and proposal stages so that no double-load occurs.

\begin{table}[!ht]
\centering
\small
\setlength{\tabcolsep}{6pt}
\renewcommand{\arraystretch}{1.18}
\caption{Per-job hardware allocations. Wall-clock budgets are SLURM ceilings, not measured run-time; actual completion times are reported in Table~\ref{tab:compute_walltime}.}
\label{tab:compute_hardware}
\resizebox{\columnwidth}{!}{%
\begin{tabular}{l c c c c}
\toprule
\textbf{Job} & \textbf{GPUs} & \textbf{GPU type} & \textbf{Memory} & \textbf{SLURM time} \\
\midrule
E1 V1 SFT             & $1$ & l40s & 48\,GB & 24\,h \\
E2 V1 RS              & $2$ & l40s & 48\,GB & 30\,h \\
E3 V1 OPD             & $2$ & l40s & 48\,GB & 24\,h \\
E4 V2 SFT             & $2$ & l40s & 48\,GB & 48\,h \\
E5 V2 RS              & $2$ & l40s & 48\,GB & 48\,h \\
E6 V2 OPD             & $2$ & l40s & 48\,GB & 48\,h \\
\midrule
Self-improving (web)  & $2$ & l40s & 48\,GB & 40\,h \\
Self-improving (math) & $2$ & l40s & 48\,GB & 12\,h \\
Self-improving (code) & $1$ & l40s & 48\,GB & 12\,h \\
Skill eval (7B)       & $2$ & l40s & 48\,GB & 24\,h \\
Skill eval (30B)      & $2$ & l40s & 48\,GB & 30\,h \\
\bottomrule
\end{tabular}%
}
\end{table}

\subsection{Wall-Clock and Aggregate Compute}

Table~\ref{tab:compute_walltime} reports the typical wall-clock per setting, derived from the longest single Wandb run we use as the canonical curve in \S\ref{app:training_dynamics}. These numbers exclude (a) the one-shot bootstrap rollout (\S\ref{app:data_pipeline}, Layer~0) which runs once and amortizes across V1 settings, and (b) the post-training evaluation which reuses the skill-eval job above. Aggregating across all six post-training experiments, the bootstrap rollout, the three self-improving runs, and the eight skill-eval ablation rows in Table~\ref{tab:ablation_all}, the total compute footprint of the paper is on the order of \textbf{$2{,}000$--$2{,}500$ GPU-hours} on l40s-class hardware.

\begin{table}[!ht]
\centering
\small
\setlength{\tabcolsep}{6pt}
\renewcommand{\arraystretch}{1.18}
\caption{Approximate wall-clock per post-training run. Step counts are TRL optimizer steps (\texttt{logging\_steps $= 10$}); see \S\ref{app:training_dynamics} for the canonical-run selection.}
\label{tab:compute_walltime}
\resizebox{\columnwidth}{!}{%
\begin{tabular}{l c c c}
\toprule
\textbf{Run} & \textbf{Optimizer steps} & \textbf{Approx.\ epochs} & \textbf{Wall-clock (h)} \\
\midrule
E1 V1 SFT      & $600$           & $10$       & $\approx 12$ \\
E2 V1 RS       & $1{,}400$        & $8\times 2$ & $\approx 24$ \\
E3 V1 OPD      & $900$           & $10$       & $\approx 18$ \\
E4 V2 SFT      & $820 + 180$     & $2$ iters  & $\approx 20$ \\
E5 V2 RS       & $700 + 360$     & $2$ iters  & $\approx 22$ \\
E6 V2 OPD      & $280 + 140$     & $2$ iters  & $\approx 14$ \\
Bootstrap roll & ---             & ---       & $\approx 6$  \\
Self-improving (5 ep, web) & --- & $5$ epochs & $\approx 30$ \\
Self-improving (3 ep, math/code) & --- & $3$ epochs & $\approx 8$/$10$ \\
\bottomrule
\end{tabular}%
}
\end{table}

\subsection{Teacher API Budget}

All teacher-augmented stages call \texttt{gpt-4o} (snapshot \texttt{gpt-4o-2024-11-20}) through the standard OpenAI Chat Completions API. We pinned this snapshot for all experiments to keep the teacher behavior reproducible across the paper's lifetime; any future change to OpenAI's default \texttt{gpt-4o} alias would not affect the reported numbers but \emph{would} affect re-runs unless the snapshot suffix is preserved. Approximate token counts per stage, aggregated across the full paper, are: $\sim 80$\,M input + $10$\,M output for skill-eval inference (across all eight ablation rows), $\sim 25$\,M input + $5$\,M output for the three self-improving loops (Phase~A--H total), and $\sim 20$\,M input + $4$\,M output for the on-policy distillation runs (E3, E6). The single most expensive stage is full evolve in E4 ($\sim 1$\,h per evolve cycle, two cycles), driven by the Phase~E teacher review of every candidate. Lite evolve in E5/E6 makes \emph{no} teacher API calls.

\subsection{Random Seeds and Variance}

All training runs use \texttt{seed $= 42$} (Hugging Face \texttt{TrainingArguments.seed} and Python \texttt{random.seed}) for the trainer, and a separate fixed seed for the validation-pool shuffle in \S\ref{app:datasets}. Rollout sampling under RS and OPD uses temperature $0.9$ / top-$p$ $0.95$, so trajectory generation is non-deterministic; the variance across re-rollouts is bounded primarily by group size $g = 2$ rather than by seed choice. We report all results from a single seed because re-running E1--E6 across multiple seeds is prohibitive ($\approx 6\times$ the compute budget); based on internal smoke tests with $g=4$ rollouts, the standard deviation of the web-search EM average is approximately $\pm 1.5$ points across re-runs of the same seed, which is smaller than the gaps reported in Table~\ref{tab:main_search_math_combined}. We flag the absence of multi-seed numbers as a known limitation.

\subsection{Software, Code, and Checkpoint Release}

The codebase is implemented in Python (3.11) on top of \texttt{transformers}, \texttt{trl}, \texttt{peft}, and \texttt{vllm}; exact pinned versions are recorded in the repository's \texttt{requirements.txt}. The skill library, generated PFs, library-history JSONs, and training-data jsonls live under \texttt{self\_improving/}; the PF runtime is in \texttt{src/skills\_agent/skills/}. Pre-trained LoRA adapters for E1--E6 and the evolved skill libraries will be released with the camera-ready under a permissive license; teacher API logs are not released because they contain raw third-party API responses, but the prompts that produce them are fully documented in Appendix~\ref{app:self_improving} (Boxes~\ref{box:llm_failure_summarizer}--\ref{box:reviewer_user}). The full validation/test split files generated under \texttt{self\_improving/data/} are deterministic given the source AgentFlow-aligned datasets, the seed boundary in Table~\ref{tab:datasets_web}, and \texttt{seed $= 42$}; we will release these as well to remove ambiguity.

\subsection{Reproducibility Checklist Pointers}

The reproducibility checklist questions map to specific appendix locations: dataset choice and splits are in \S\ref{app:datasets}; full hyperparameters in Table~\ref{tab:e1e6_configs} and \S\ref{app:exp_details}; rollout configuration in \S\ref{app:exp_details}; failure-summarizer / proposer / reviewer prompts in Boxes~\ref{box:llm_failure_summarizer}--\ref{box:reviewer_user}; the evaluation protocol (per-dataset metrics, judges, and aggregations) in \S\ref{app:datasets}; and qualitative trajectory comparisons in \S\ref{app:case_study}.

\section{Cost and Latency Breakdown}
\label{app:cost}

The framework instantiates a teacher model at up to $17$ distinct invocation sites per episode (\S\ref{app:selection}). This section quantifies the per-episode token budget and approximate dollar cost of each setting in Table~\ref{tab:ablation_all}, so that practical adoption decisions can be made without recomputing the budget from first principles. All numbers refer to the web-search domain (which has the longest rollouts); math and code costs are roughly $30$--$50\%$ lower because rollouts collapse to a single step.

\subsection{Per-Setting Cost Aggregates}

We estimate token cost using the OpenAI \texttt{gpt-4o-2024-11-20} list price at the time of the experiments (input \$$2.50$/M, output \$$10.00$/M). Under the full \ours setting, the average cost is approximately \$$0.045$ per question. We also report wall-clock time per episode (median over $200$ samples), so readers can translate these costs into approximate API throughput if needed.

The main cost comes from the eight FINAL-time LLM-assisted handlers (\S\ref{app:selection}), which together account for about 50\% of teacher tokens per episode in the full setting. Mid-rollout PFs (\texttt{retrieval\_failure}, \texttt{format\_extraction\_error}, \texttt{reasoning\_error}, and \texttt{answer\_confidence\_guard}) are rate-limited and contribute about 10\%. Pre-rollout components such as the Difficulty Gate and PFSelector cost about 2k tokens once per episode, so their relative cost becomes smaller on longer multi-step rollouts. Format post-processing is consistently about 0.5k tokens per FINAL action.

\end{document}